\documentclass[11pt]{article}

\usepackage[preprint]{acl}

\usepackage{times}
\usepackage{latexsym}

\usepackage[T1]{fontenc}

\usepackage[utf8]{inputenc}

\usepackage{microtype}

\usepackage{inconsolata}

\usepackage[dvipsnames]{xcolor}
\usepackage[table, x11names]{xcolor}
\usepackage{amsmath}
\usepackage{amsfonts}
\usepackage{hyperref}

\usepackage{multirow} 
\usepackage{makecell}  
\usepackage{booktabs}

\usepackage{algorithm2e}
\usepackage{algpseudocode}
\RestyleAlgo{ruled}
\SetKwComment{Comment}{/* }{ */}

\usepackage{subcaption} 

\usepackage{graphicx}
\usepackage{comment}
\usepackage{amsmath}
\title{APP: Accelerated Path Patching with Task-Specific Pruning}

\author{
    Frauke Andersen$^{\star 1}$, William Rudman$^{\star 2}$, Ruochen Zhang$^{\star 3}$, \\
    {\bf Carsten Eickhoff$^{1}$} \\
    $^{1}$University of Tübingen, $^{2}$The University of Texas at Austin, $^{3}$Brown University \\
    \texttt{william.rudman@utexas.edu} \\
}

\begin{document}
\maketitle
\begin{abstract}
Circuit discovery is a key step in many mechanistic interpretability pipelines. Current methods, such as Path Patching, are computationally expensive and have limited in-depth circuit analysis for smaller models. In this study, we propose \textit{Accelerated Path Patching} (APP), a hybrid approach leveraging our novel \textit{contrastive} attention head pruning method to drastically reduce the search space of circuit discovery methods. Our Contrastive-FLAP pruning algorithm uses techniques from causal mediation analysis to assign higher pruning scores to task-specific attention heads, leading to higher performing sparse models compared to traditional pruning techniques. Although Contrastive-FLAP is successful at preserving task-specific heads that existing pruning algorithms remove at low sparsity ratios, the circuits found by Contrastive-FLAP alone are too large to satisfy the minimality constraint required in circuit analysis. APP first applies Contrastive-FLAP to reduce the search space on required for circuit discovery algorithms by, on average, 56\%.  Next, APP, applies traditional Path Patching on the remaining attention heads, leading to a speed up of 59.63\%-93.27\% compared to Path Patching applied to the dense model. Despite the substantial computational saving that APP provides, circuits obtained from APP exhibit substantial overlap and similar performance to previously established Path Patching circuits. \footnote{Equal contribution. Order determined by coin flip.} 
\footnote{Code: \href{https://github.com/Frau-Ke/AcceleratedPathPatching}{
\textit{https://github.com/Frau-Ke/AcceleratedPathPatching}}}
\end{abstract}

\section{Introduction}
A central focus of mechanistic interpretability research~\citep{elhage2021mathematical, olah2022mechanistic} is the study of minimal subgraphs, or circuits, within Large Language Models (LLMs) to identify the internal mechanisms responsible for particular functions~\citep{elhage2021mathematical, olah2022mechanistic}.  Circuits consist of subsets of model components, such as attention heads or feed-forward layers, that can largely account for a model’s performance on certain tasks~\citep{vig2020causalmediationanalysisinterpreting, wang2022interpretabilitywildcircuitindirect, GreaterThan}. Locating these components within a highly overparameterized LLM is computationally expensive. To verify whether a particular component belongs to a circuit, recent studies employ techniques such as Path Patching~\citep{goldowskydill2023localizingmodelbehaviorpath, wang2022interpretabilitywildcircuitindirect}, which requires multiple model runs to measure the causal effects of individual components. Since this process must be repeated for \textit{every component} to fully outline a circuit, it quickly becomes prohibitively costly as we increase the number of parameters in a model.

The objective of identifying a minimal, well-performing subnetwork aligns with pruning~\citep{Frankle2018TheLT}, which seeks to improve model efficiency by removing uninformative model components. Unlike circuit discovery methods, many pruning algorithms are computationally efficient, requiring only a few forward passes and using weight- or activation-based heuristics~\citep{PruningSurveyZhu2024, WANDA2024} to remove unimportant model components. In this paper, we explore the connection between Path Patching and unstructured pruning~\citep{sparseGPT, WANDA2024, FLAP2024} that are adapted to preserve full-attention heads, rather than focusing on algorithms that target individual model weights ~\citep{PruningSurveyZhu2024}. Specifically, we employ FLAP ~\citep{FLAP2024} to investigate whether pruning alone can recover minimal circuits. We demonstrate that, under the same performance budget, circuits obtained through pruning are significantly larger than those identified via path patching. Namely, pruning cannot truly recover \textit{minimal} circuits.

By comparing FLAP-generated circuits to those produced by Path Patching, we observe that FLAP often removes task-specific heads that are critical for subnetwork performance. These task-specific heads only activate when exposed to a particular input and often implement specialized, interpretable functions.  Although pruning alone cannot reduce an LLM to its minimal subnetwork, it can serve as an effective preprocessing step that substantially \textit{reduces} the search space for circuit discovery. Building on this insight, we propose Accelerated Path Patching (APP), which incorporates pruning our novel Contrastive-FLAP pruning algorithm as a preliminary stage of the patching process. Contrastive-FLAP utilizes the contrastive minimal-pair setup commonly used in causal mediation analysis to target task-specific heads found by Path Patching. Using APP for circuit discovery recovers minimal circuits with comparable performance to Path Patching while reduces computational costs by up to 93\%.

Our main contributions are as follows:
\begin{itemize}
\item We provide a detailed analysis of the differences between path patching and the most closely related attention-head pruning algorithm, FLAP, showing that pruning alone fails to recover minimal circuits.
\item We propose Contrastive-FLAP to better preserve task-specific attention heads that are otherwise pruned at low sparsity ratios.
\item We replace the original Path Patching search space with the union set identified by Contrastive-FLAP and FLAP. By incorporating pruning as a prior step, we introduce APP, which significantly reduces search cost while faithfully recovering minimal circuits.
\end{itemize}

\begin{figure*}
    \centering
    \includegraphics[width=0.85\textwidth]{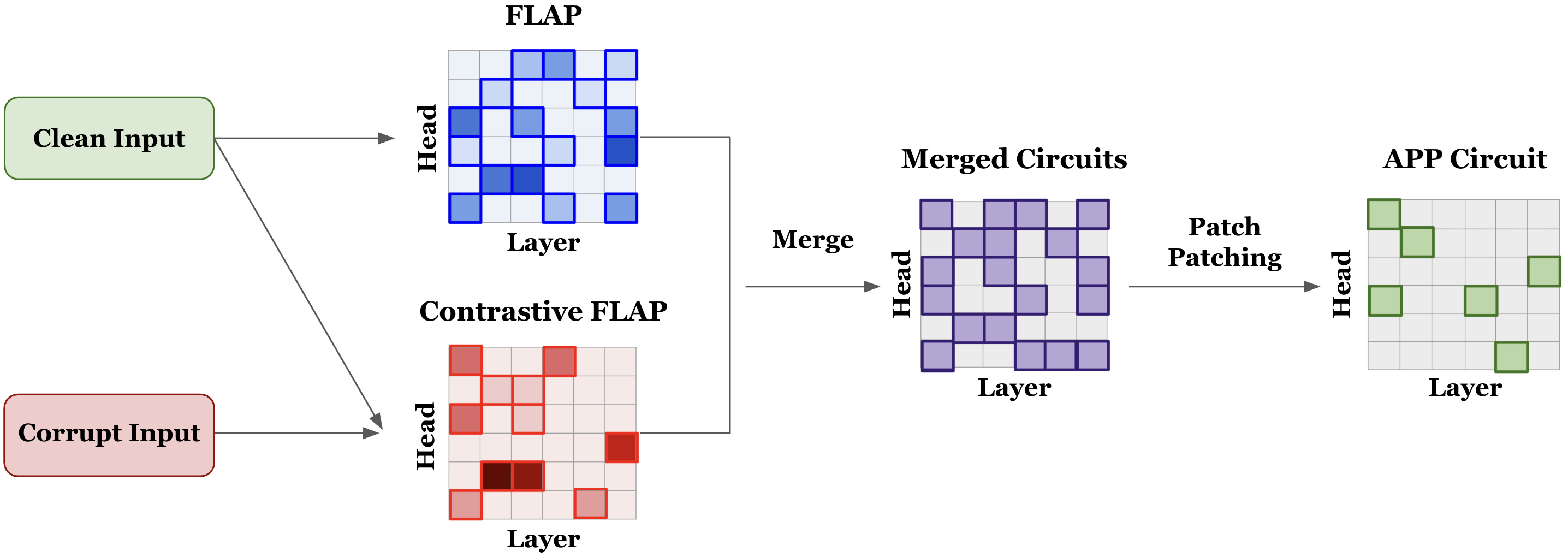}    \caption{Depiction of the Accelerated Path Patching (APP) Algorithm. APP reduces the search space of circuit discovery methods by successfully pruning task-irrelevant heads while preserving task-critical attention heads. APP then runs Path Patching on the remaining sparse model.}
    \label{fig:app-diagram}
\end{figure*}

\section{Related Work}
\label{sec:relatedWork}
\subsection{Circuits}
In transformer models, circuits refer to computational subgraphs that implement specific, often human-interpretable, behaviors~\citep{zoomIn2020}. Depending on the level of granularity, nodes in these graphs may correspond to attention heads, MLPs, or even individual query, key, and value activations~\citep{acdc2023}. Edges typically represent residual connections, linear projections on the residual stream, or interactions within attention and MLP blocks~\citep{elhage2021mathematical}.

Path patching is one of the most widely used techniques for circuit discovery~\citep{vig2020causalmediationanalysisinterpreting, Geiger2021CausalAO, wang2022interpretabilitywildcircuitindirect, GreaterThan}. The method localizes model components that causally influence the output by contrasting clean and corrupted input pairs. It involves caching activations for both inputs and selectively replacing the activations of individual components (e.g., heads) in the clean run with those from the corrupted run. This process identifies components whose intervention most significantly alters the final logits. When combined with logit attribution techniques~\citep{logit_lens, Yu2023CharacterizingMF, Zhang2024TheSB, notice}, path patching enables fine-grained functional interpretations of model behavior. For instance, in the IOI (Indirect Object Identification) task, \citet{wang2022interpretabilitywildcircuitindirect} identify interpretable head types such as previous token heads and name mover heads.

However, circuit discovery still remains computationally expensive. Recent methods like ACDC~\citep{acdc2023} aim to automate this process by iteratively applying activation patching while pruning components with sub-threshold effects. However, circuits discovered via such automated approaches can be noisy and may omit components whose contributions to the logits negatively impact model performance. Other works~\citep{EAP2023, eap-ig2024} explore related automated techniques focusing on edge-level localization rather than component-level discovery, which is a complementary but distinct research focus from this paper.

\begin{table*}[t]
\centering
\resizebox{\textwidth}{!}{%
\begin{tabular}{llll}
\toprule
     \textbf{Task} & \textbf{Prompt} & \textbf{Corrupted Prompt} & \textbf{LD} = L(\textcolor{Green}{correct}) - L (\textcolor{Red}{wrong}) \\
     \midrule
     
     \textbf{IOI} 
     & \makecell[l]{When \textcolor{Red}{John} and \textcolor{Green}{Mary} went to the store, \\ \textcolor{Red}{John} bought a drink for $\ldots$} 
     &  \makecell[l]{When \textcolor{Red}{John} and \textcolor{Green}{Mary} went to the store, \\ \textcolor{DarkOrange1}{Alex} bought a drink for $\ldots$} 
     & L(\textcolor{Green}{Mary}) - L(\textcolor{Red}{John})   \\
    \midrule
     
     \textbf{Greater Than} 
     & The war lasted from 18\textcolor{DeepSkyBlue3}{73} to 18$\ldots$ 
     & The war lasted from 18\textcolor{DarkOrange1}{01} to 18$\ldots$
     & L($\{x|(\textcolor{Green}{ x > 73}) \land (\textcolor{Green}{x \leq 98})\}$) - L($\{x|\textcolor{Red}{x \leq 73}\}$) \\
    \midrule
     
     \textbf{Gendered Pronouns}
     & So \textcolor{DeepSkyBlue3}{Emily} is such a good friend, isn't $\ldots$ 
     & That \textcolor{DarkOrange1}{Person} is such a good friend, isn't $\ldots$ 
     & L(\textcolor{Green}{she}) - L(\textcolor{Red}{he}) \\
    \midrule
     
     \textbf{Induction} 
     & Today,  \textcolor{DeepSkyBlue3}{Cl}\textcolor{Green}{aire} visited the library. There \textcolor{DeepSkyBlue3}{Cl}$\ldots$
     & Today,  \textcolor{DeepSkyBlue3}{Cl}\textcolor{Green}{aire} visited the library. There \textcolor{DarkOrange1}{Tr}$\ldots$
     
     & L(\textcolor{Green}{aire}) -  L(\textcolor{Red}{istan}) \\
    \midrule
     
     \textbf{Docstring}
     & \makecell[l]{def old(self, \textcolor{Red}{first}, \textcolor{Red}{page}, \textcolor{Red}{names}, \textcolor{Green}{size}, \textcolor{Red}{files}, \textcolor{Red}{read}): \\
     """sector gap population \\
       :param \textcolor{Red}{page}: message tree \\
       :param \textcolor{Red}{names}: detail mine \\
    :param $\ldots$
     } 
     & \makecell[l]{def old(self, \textcolor{Red}{first}, \textcolor{Red}{project}, \textcolor{Red}{target}, \textcolor{Red}{new}, \textcolor{Red}{files}, \textcolor{Red}{read}): \\
    """sector gap population \\
    :param \textcolor{Red}{image}: message tree \\
    :param \textcolor{Red}{update}: detail mine \\
    :param
    }
     & \makecell{\textcolor{Red}{Wrong Variables} = \\ $\{\text{first, page, names, files, read, project, target, new}\}$ \\
     $L(\text{\textcolor{Green}{size}}) - \mathop{\arg \max}\limits_{x \in \text{\textcolor{Red}{Wrong Variables}}} L(x)$} \\
\bottomrule
\end{tabular}
}
\caption{Example prompts and the logit differences for all five tasks used in this work. Logit Difference (LD) is defined by the difference between the logits of the correct and wrong answer.}
\label{tab:tasks}
\end{table*}
\subsection{Pruning}
Pruning refers to the process of reducing a model’s size or computational complexity while preserving overall performance. It can be broadly categorized as unstructured, semi-structured, or structured~\citep{PruningSurveyZhu2024}. Unstructured pruning~\citep{WANDA2024, sparseGPT} removes individual parameters, resulting in irregular sparse connectivity that often requires retraining or specialized sparse kernels when applied at high sparsity. Structured pruning, in contrast, removes entire model components, such as attention heads~\citep{FLAP2024}, channels~\citep{LLMPruner2023}, or even full layers~\citep{LayerDrop2019}, yielding more hardware-efficient architectures. Semi-structured pruning combines aspects of both approaches.

Since our goal is to compare pruning methods that operate at the attention-head level, similar to circuit discovery via path patching, we focus on structured, one-shot pruning techniques, specifically Fluctuation-based Adaptive Structured Pruning (FLAP)~\citep{FLAP2024}. FLAP eliminates the need for retraining by computing an importance score for each attention head, which integrates both weight magnitude and activation statistics~\citep{WANDA2024}. Heads with the lowest scores are pruned iteratively until a target sparsity ratio is reached. Finally, FLAP introduces a bias correction term to each layer to mitigate pruning-induced errors. In this work, we omit this final correction step, as our objective is to use the pruning decisions themselves to identify functionally important components for circuit discovery.




\begin{figure*}[t]
    \centering
    \begin{subfigure}{0.25\textwidth}
        \centering
        \includegraphics[width=\linewidth]{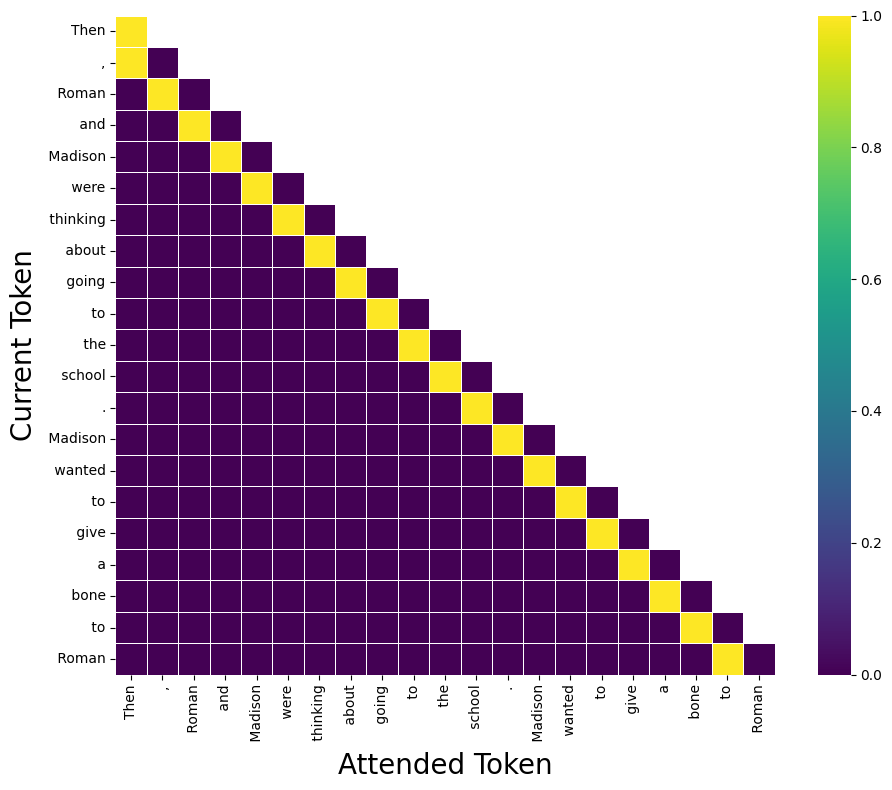}
        \caption{\textbf{clean}: Previous Token Head}
    \end{subfigure}
    \hfill
    \begin{subfigure}{0.25\textwidth}
        \centering
        \includegraphics[width=\linewidth]{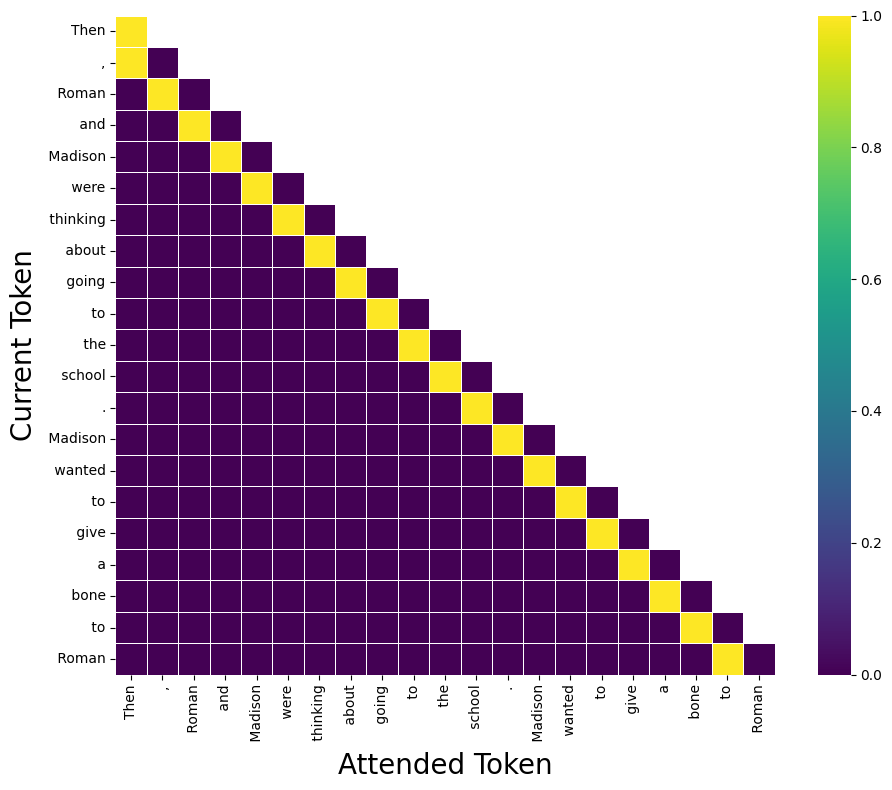}
        \caption{\textbf{corrupted}: Previous Token Head}
    \end{subfigure}
    \hfill
    \begin{subfigure}{0.25\textwidth}
        \centering
        \includegraphics[width=\linewidth]{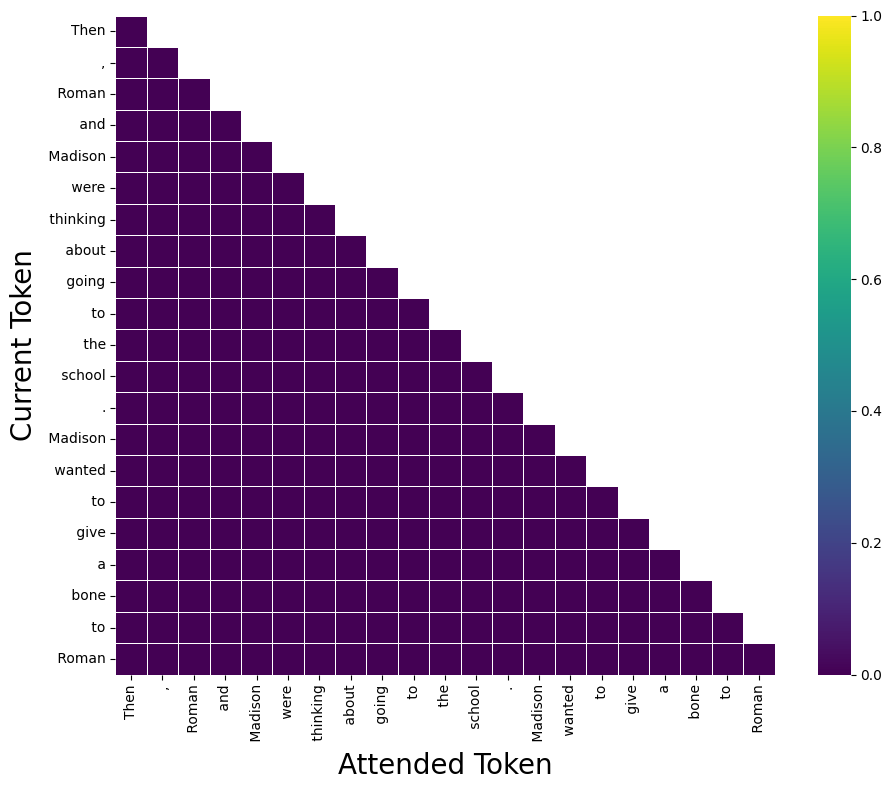}
        \caption{\textbf{contrastive}: Previous Token Head}
    \end{subfigure} \\ 
    \par\medskip
    \begin{subfigure}{0.25\textwidth}
        \centering
        \includegraphics[width=\linewidth]{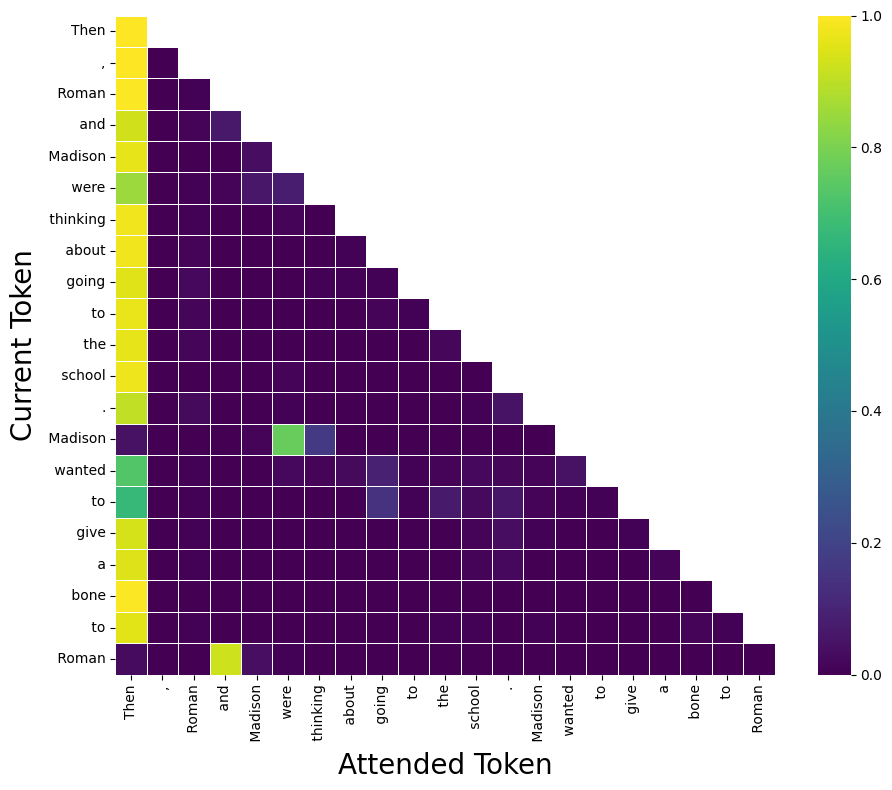}
        \caption{\textbf{clean}: Induction Head}
    \end{subfigure}
    \hfill
    \begin{subfigure}{0.25\textwidth}
        \centering
        \includegraphics[width=\linewidth]{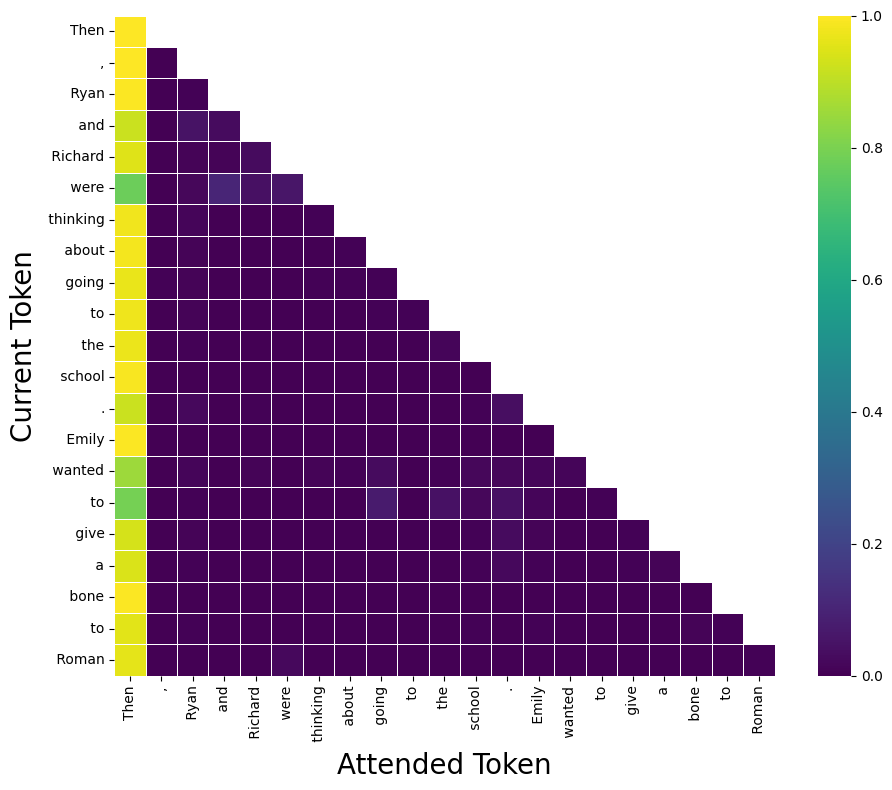}
        \caption{\textbf{corrupted}: Induction Head}
    \end{subfigure}
    \hfill
    \begin{subfigure}{0.25\textwidth}
        \centering
        \includegraphics[width=\linewidth]{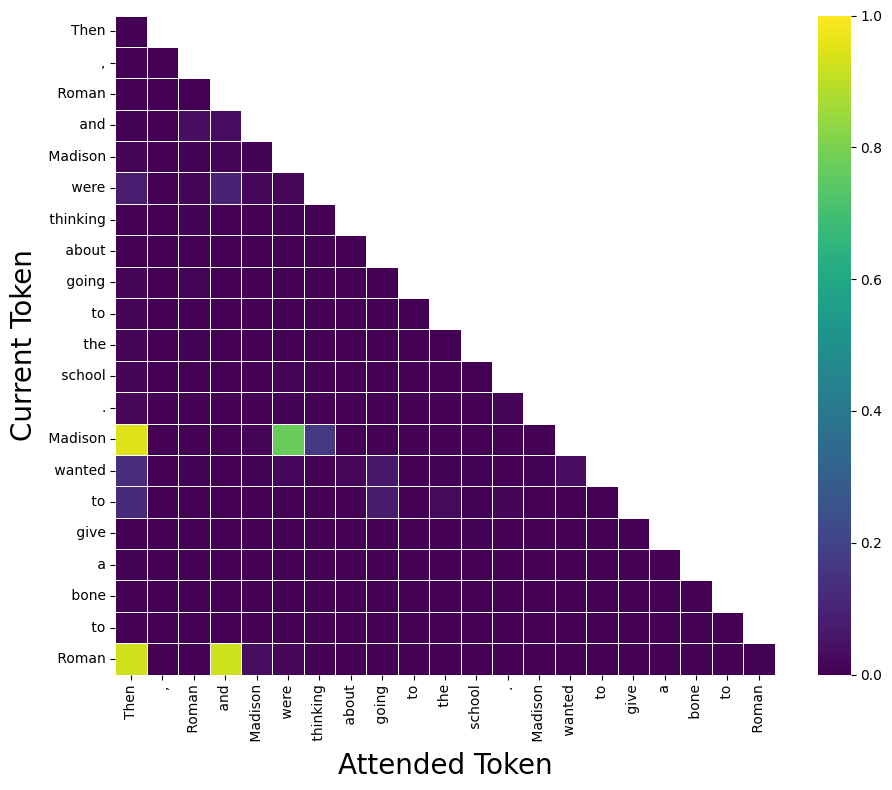}
        \caption{\textbf{contrastive}: Induction Head}
    \end{subfigure}
    \caption{Activation Patterns of context-insensitive (\textbf{top}) and context-senstive (\textbf{bottom}) heads.
    \textbf{Left}: clean activations, \textbf{middle}: corrupted activations, \textbf{right}: contrastive activations.}
    \label{fig:hybridFLAP_activations}
\end{figure*}

\begin{figure}
    \centering
    \includegraphics[width=\linewidth]{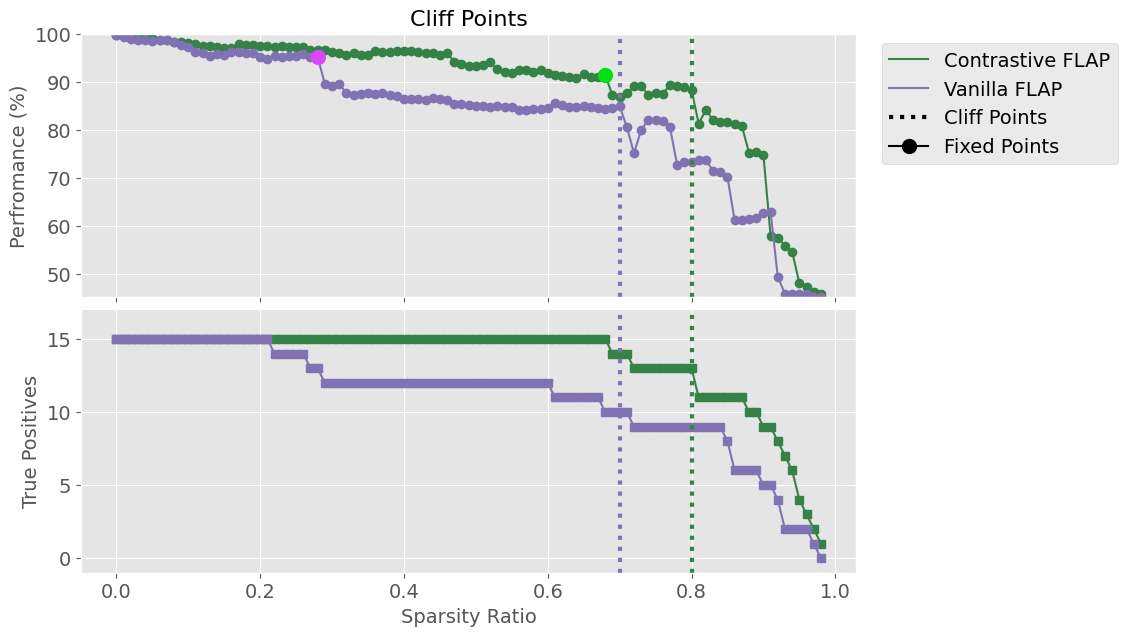}
    \caption{Performance and TP over a sparsity of 0 to 1 for vanilla and contrastive FLAP. Points are the fixed sparsity used in Table \ref{tab:exp1}, dotted lines are the sparsity ratios found via cliff points.}
    \label{fig:performance_curve}
\end{figure}

\section{Methodology}

\subsection{Experimental Setup}
\paragraph{Models \& Tasks}
In this paper, we conduct all patching experiments using four models: GPT-2 Small, GPT-2 Large, Qwen2.5-0.5B, and Qwen2.5-7B. We include the GPT-2 family~\citep{Radford2019LanguageMA} since it has been extensively used in prior mechanistic interpretability research, and well-characterized circuits have been identified in GPT-2 Small. The Qwen2.5 models~\citep{Yang2024Qwen25TR} are selected for their strong performance across a range of downstream tasks. We evaluate these models on five benchmark tasks commonly used in mechanistic interpretability: Indirect Object Identification (IOI) \cite{wang2022interpretabilitywildcircuitindirect}, Greater Than \cite{GreaterThan}, Gendered Pronouns \cite{GenderedPronouns2023}, Induction \cite{induction}, and Docstring \cite{Docstring2023}. Each of these tasks contains a paired clean set $\mathcal{D}^{\text{clean}}$ and a corrupted set $\mathcal{D}^{\text{corr}}$, which is identical except that the task-relevant information is removed. See Table~\ref{tab:tasks} for a brief description and \ref{sec:tasks} for a long description of the individual tasks.

\paragraph{Metrics}
We quantify patching effects using the average logit difference (LD) over all samples for a given task. Let $X^{\text{clean}}$ denote the correct answer for the clean input and let $X^{\text{corr}}$ be the correct answer for the corrupted input. Let $L(X^{\text{clean}}, X^{\text{corr}}) = \text{Logit}(X^{\text{clean}}) \ \text{-} \ \text{Logit}(X^{\text{corr}})$. Logit difference is then defined as $L^{\star}(X^{\text{clean}}, X^{\text{corr}})\text{-}L^{'}(X^{\text{clean}}, X^{\text{corr}})$ where $L^{\star}$ and $L^{'}$ denote logits from the patched and the corrupted runs. A positive score means that the model predicts the correct answer with a higher probability than the incorrect answer. Different LD implementations for various tasks are discussed in the Appendix \ref{sec:tasks}. In addition to using LD as our measure for circuit inclusion, we use average LD to measure the patching effect during PP and circuit performance. To evaluate circuit performance, all heads in the model, except for the ones included in the circuit are set to the corrupted activations. Performance is measured by the percentage of the original, average LD restored by the patched model. 

\subsection{Circuit Discovery with Path Patching}
Activation patching and causal mediation analysis using contrastive minimal pairs form the basis for many circuit discovery methods. Activation patching measures the causal influence of model components by replacing their activations with corrupted ones and comparing the resulting output. Path Patching (PP) \citep{wang2022interpretabilitywildcircuitindirect, goldowskydill2023localizingmodelbehaviorpath} extends this idea by tracing information flow along specific computational paths. A path is defined by a sender node $p$ and a receiver node $r$. During a forward pass on the clean dataset, $\mathcal{D}^{clean}$, activations at $p$ are replaced with those from the corrupt dataset, $\mathcal{D}^{corr}$. A large logit difference between the patched and unpatched runs indicates that $p$ causally affects $r$. PP is applied iteratively, starting from the final logits as receivers and all attention heads as senders. Heads with a causal effect become the new receivers, and earlier heads become senders. In this study, we apply PP only to attention heads and exclude MLPs. To reduce the manual effort required for circuit discovery, we use an iterative version of PP proposed by \citet{AutomaticPP2024}, which determines inclusion criteria based on the standard deviation of the average patching effect, referred to as Automatic PP. We further adapt this procedure using two thresholding criteria to determine circuit inclusion: the importance threshold and the maximum value threshold. First, the \textbf{importance threshold} is evaluated over each logit difference score $x_{i, j}$ 
$$|x_{i, j}| - |\bar{X}| > K * \mathbf{SD}(X)$$
where $\bar{X}$ and $\mathbf{SD}(X)$ denote the mean and standard deviation calculated over all logit differences respectively, and $K$ is a predefined constant.
At early layers in the model, the constant is adjusted to $$K' = K + \frac{2}{\sqrt{l_s * H}}$$ where $K'$ increases as $l_s$ decreases. The adjusted constant $K'$ is evaluated for each sender layer $l_s$. This adjustment ensures that path patching focuses on sequential influence: components that are more distant from the receiver must meet stricter significance criteria to be included. For evaluation, K is set to $[1, 1.5, 2, 2.5]$. The second criterion, \textbf{maximum value thresholds}, ensures that at least one score $x_{i,j}$ is above a chosen value $\epsilon$. Otherwise, no head for the current receiver will be included in the circuit. This prevents the inclusion of spurious or weakly contributing components in the constructed circuit. Values chosen for $\epsilon$ are $[0.01, 0.001, 0.02, 0.002]$

Heads that exceed both threshold are considered significant contributors to the model's behavior and are selected as receivers in subsequent iterations of the algorithm. Further details on automated path patching and comparisons between manual and automatic PP are provided in Appendix~\ref{sec:automatic_PP}. Circuits discovered by Automatic PP serve as the ground truth for pruning comparisons.

\subsection{Contrastive FLAP}


Preserving context-sensitive, task-critical heads is necessary for discovering high-performing, minimal circuits. As discussed in Section~\ref{sec:context_sensitive}, context-sensitive heads produce different activations across clean and corrupted data, while context-insensitive heads remain constant. Figure~\ref{fig:hybridFLAP_activations} illustrates the activation patterns of an induction head in GPT-2 under clean and corrupted inputs, as well as the corresponding contrastive activations (clean - corrupted). Since FLAP is based purely on the weight and activation patterns, it is not guaranteed to preserve task-specific heads. In Appendix~\ref{sec:SubtractedFLAP}, we show empirically that vanilla FLAP prunes task-specific, context-sensitive heads at higher sparsity ratios, leading to poor task performance. In order to preserve task-specific, context-sensitive heads, we propose Contrastive-FLAP. Contrastive-FLAP computes activation scores for both clean and corrupted inputs and derives final scores on the contrastive activations (clean$-$corrupted). This procedure effectively isolates task-specific heads, as context-insensitive heads exhibit near-identical activation patterns across clean and corrupted conditions, yielding negligible importance scores.

Formally, let $X_{clean} \in \mathbb{R}^{(B\times S) \times C_{in}}$, be the input activation under the clean dataset, where $B$ is the batch size, $S$ is the sequence length and $C_{in}$ is the dimension of the input channel. Likewise, the input activations obtained by the corrupted dataset is $X_{corr} \in \mathbb{R}^{(B\times S) \times C_{in}}$ and $W \in \mathbb{R}^{(C_{out} \times C_{in})}$ is the weight matrix.
Then, the importance scores for Contrastive FLAP $\tilde{S}$ are calculated by
$$\tilde{S} = |W_{i, j}| * ||X_{clean} - X_{corr}||_2$$.
Figure \ref{fig:hybridFLAP_activations} (c) and (f) visualize that contrastive activations allow us to retain context-sensitive heads while excluding context-insensitive ones. In Appendix~\ref{sec:SubtractedFLAP}, we find that Contrastive FLAP indeed excludes task-critical heads only at higher sparsity compared to vanilla FLAP. 

\paragraph{Determining Sparsity Ratios with Cliff Points}
In typical pruning scenarios, a sparsity ratio is predefined by the user. To determine the appropriate level of sparsity for contrastive pruning, we automatically select the sparsity ratio based on identified \textbf{cliff points}. Cliff points indicate the removal of task-critical heads that are crucial for maintaining high-performance. Selecting a sparsity level just before the drop ensures these components remain in the circuit. Additionally, a minimal sparsity level is set to 0.6 for both circuits. Using 0.6 as the minimum, we then consider three different cliff points are considered for evaluation: the first drop after 0.6, the biggest drop in performance after 0.6, and a fixed maximal sparsity value of 0.75. In Figure \ref{fig:performance_curve}, the first drop is visualized for vanilla and Contrastive-FLAP.

\subsection{Can Pruning find Circuit Components?}
To investigate whether pruning can find circuit components, we first need to discover ground truth circuits. In this section, we focus on only GPT-2 small since circuits are well studied with this model. Automated PP is evaluated over 100 random samples per task. Two hyperparameters are tested and from all resulting circuit, the smallest circuit with a performance of at least 75\% accuracy is selected. If no circuit achieves 75\% accuracy, we select the circuit that maximizes performance, while minimizing circuit size. Vanilla and contrastive FLAP are computed over 200 samples per task. Instead of fixing a specific sparsity value, sparsity is varied between 0 and 1 in increments of 0.1. The smallest pruning circuit matching the performance of the path patching circuit is selected. Table~\ref{tab:exp1} reports the performance, size, and the sparsity ratio of the vanilla and Contrastive FLAP. Circuit size is defined as the total amount of included attention heads, while sparsity ratio is defined as the percentage of the model excluded from the circuit. If the circuit matches in size and a high TPR value is reported, FLAP circuits are similar to PP circuits.

Table~\ref{tab:exp1} shows that PP circuits are smaller than pruning circuits when performance is matched. The PP circuits are at least 47.62\% smaller than the vanilla FLAP circuits and at least 4.17\% smaller than constrative FLAP. On average, vanilla FLAP retains 86.61\% of the ground truth heads discovered with Path Patching while maintaining an average sparsity of 0.45, while contrastive FLAP retains 86.17\% of the ground truth heads with an average sparsity of 0.68, indicating that Contrastive FLAP is better able to identify task-critical heads and remove task-irrelevant heads when compare to vanilla FLAP. For the Gendered Pronouns task, vanilla FLAP produces a much larger circuit (87 heads) than the ground truth, whereas Contrastive FLAP preserves only 38 heads while achieving higher accuracy (74.78\% vs. 72.98\% for vanilla FLAP). In this case, Contrastive FLAP maintains 100\% of the heads contained in the ground truth, PP circuit. Interestingly, for the Docstring task, Contrastive FLAP yields a circuit that outperforms the Path Patching (PP) circuit in task performance despite having a lower TPR. This suggests that TPR alone is an insufficient metric for evaluating circuit overlap quality. Moreover, it indicates that the PP circuit may not represent a unique or minimal ``gold-standard'' circuit since alternative subnetworks can achieve comparable or even superior task performance.

\textbf{Pruning cannot be used as an alternative for circuit discovery} since the resulting circuits are too large and fail to satisfy the minimality constraint. More importantly, vanilla pruning identifies statistically important rather than causally relevant components, often removing context-dependent heads. While Table~\ref{tab:exp1} demonstrates that Contrastive FLAP is better suited for identifying smaller circuits compared to FLAP, it still produces circuits that are too large compared to Path Patching. Further, the failure of pruning to replace path patching can be linked to how pruning overlooks the compositional structure of circuits and does not consider how multiple heads interact, whereas Path Patching explicitly isolates these causal pathways. Pruning can approximate task performance but cannot recover the causal and functionally minimal circuits revealed by mechanistic discovery methods. Instead of using pruning to replace Path Patching, we use both vanilla and Contrastive FLAP in tandem with circuit discovery methods to improve the efficiency of finding circuits. 

\begin{figure}
    \centering
    \includegraphics[width=\linewidth]{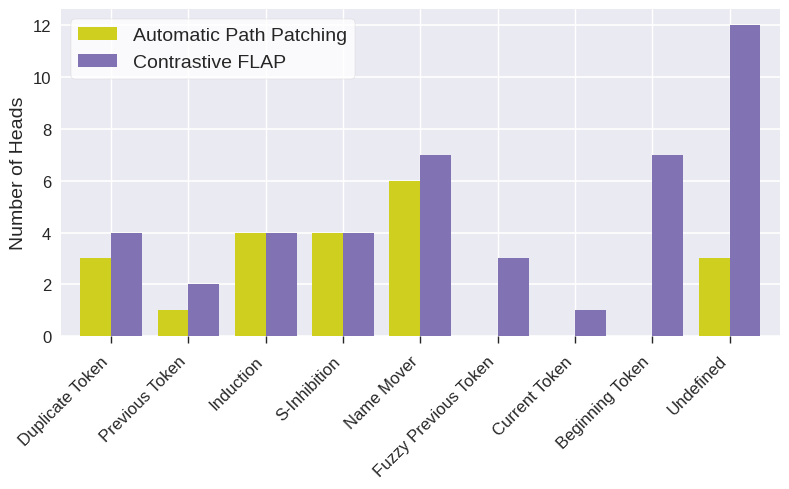}
    \caption{Comparison of attention head types identified by Contrastive FLAP and Automatic Path Patching for the IOI task in GPT-2 small.}
    \label{fig:circuit_analysis_IOI}
\end{figure}

\begin{table*}[h!]
\centering
\resizebox{\textwidth}{!}{%
\begin{tabular}{ll|ccc|ccc|ccc|ccc|ccc}
\toprule
\textbf{Model} & \textbf{Method} 
& \multicolumn{3}{c|}{\textbf{IOI}} 
& \multicolumn{3}{c|}{\textbf{GreaterThan}} 
& \multicolumn{3}{c|}{\textbf{GenderedPronouns}} 
& \multicolumn{3}{c|}{\textbf{Induction}} 
& \multicolumn{3}{c}{\textbf{Docstring}} \\
\cmidrule(lr){3-17}
& & Perf. (\%) & Size (Sparsity) & TPR
& Perf. (\%) & Size (Sparsity) & TPR
& Perf. (\%) & Size (Sparsity) & TPR
& Perf. (\%) & Size (Sparsity) & TPR
& Perf. (\%) & Size (Sparsity) & TPR \\
\midrule

\rowcolor{gray!20} GPT-2 Small & Path Patching & 96.95 & 21 (0.85) & -- & 79.65 & 8 (0.94) & -- & 74.74 & 5 (0.97) & -- & 93.35 & 22 (0.83) & -- & 64.61 & 46 (0.68) & -- \\
GPT-2 Small & FLAP & \textbf{93.51} & 49 (0.66) & 80.95 & 7.63 & \textbf{16 (0.89)} & 37.50 & 33.41 & 35 (0.76) & 60.00 & 48.95 & \textbf{9 (0.94)} & 13.64 & 36.33 & \textbf{23 (0.84)} & 36.33 \\
GPT-2 Small & Contrastive FLAP & 67.34 & \textbf{22 (0.84)} & 76.19 & 69.38 & 31 (0.78) & \textbf{87.50} & 32.44 & \textbf{29 (0.80)} & 80.00 & 89.39 & 45 (0.69) & \textbf{90.91} & 69.39 & 52 (0.64) & 71.73 \\
GPT-2 Small & Merged & 92.60 & 54 (0.63) & \textbf{90.48} & \textbf{73.08} & 37 (0.74) & \textbf{87.50} & \textbf{69.36} & 48 (0.67) & \textbf{100.00} & \textbf{88.69} & 47 (0.67) & \textbf{90.91} & \textbf{76.36} & 60 (0.58) & \textbf{76.09} \\
\midrule

\rowcolor{gray!20} GPT-2 Large & Path Patching & 92.60 & 193 (0.73) & -- & 75.71 & 27 (0.96) & -- & 73.94 & 73 (0.89) & -- & 90.90 & 67 (0.91) & -- & 73.19 & 122 (0.83) & -- \\
GPT-2 Large & FLAP & 69.97 & 186 (0.74) & 46.11 & \textbf{106.90} & 186 (0.74) & 74.07 & 68.15 & 244 (0.66) & 67.12 & 41.57 & \textbf{107 (0.85)} & 35.82 & 75.27 & 201 (0.72) & 56.56 \\
GPT-2 Large & Contrastive FLAP & 51.40 & \textbf{107 (0.85)} & 35.75 & 98.51 & \textbf{150 (0.79)} & 77.78 & \textbf{98.17} & \textbf{179 (0.75)} & 67.12 & 96.77 & 121 (0.83) & 64.18 & 45.18 & \textbf{71 (0.90)} & 27.87 \\
GPT-2 Large & Merged & \textbf{86.85} & 220 (0.69) & \textbf{53.37} & 105.28 & 237 (0.67) & \textbf{85.19} & 96.20 & 310 (0.57) & \textbf{82.19} & \textbf{99.25} & 186 (0.74) & \textbf{71.16} & \textbf{84.05} & 221 (0.69) & \textbf{61.48} \\
\midrule

\rowcolor{gray!20} Qwen2.5-0.5B & Path Patching & 76.33 & 62 (0.82) & -- & 75.14 & 20 (0.94) & -- & 86.84 & 25 (0.93) & -- & 76.08 & 18 (0.95) & -- & 36.04 & 34 (0.90) & -- \\
Qwen2.5-0.5B & FLAP & 17.72 & \textbf{19 (0.94)} & 20.97 & 34.45 & \textbf{53 (0.84)} & 60.00 & 62.24 & 70 (0.79) & 56.00 & 58.58 & \textbf{83 (0.75)} & 55.56 & 35.09 & \textbf{19 (0.94)} & 19.05 \\
Qwen2.5-0.5B & Contrastive FLAP & \textbf{82.84} & 113 (0.66) & \textbf{67.74} & 87.01 & 130 (0.61) & \textbf{95.00} & 55.27 & \textbf{59 (0.82)} & 40.00 & 93.12 & 127 (0.62) & \textbf{100.00} & \textbf{72.13} & 133 (0.60) & \textbf{85.71} \\
Qwen2.5-0.5B & Merged & 82.54 & 114 (0.66) & \textbf{67.74} & \textbf{87.53} & 138 (0.59) & \textbf{95.00} & \textbf{69.90} & 89 (0.74) & \textbf{69.90} & \textbf{94.86} & 149 (0.56) & \textbf{100.00} & \textbf{72.13} & 133 (0.60) & \textbf{85.71} \\
\midrule

\rowcolor{gray!20} Qwen2.5-7B & Path Patching & 73.25 & 221 (0.72) & -- & 79.33 & 34 (0.95) & -- & 68.69 & 125 (0.84) & -- & 67.90 & 10 (0.99) & -- & 64.79 & 150 (0.81) & -- \\
Qwen2.5-7B & FLAP & 9.38 & \textbf{195 (0.75)} & 41.17 & 30.29 & 195 (0.75) & 52.94 & 2.02 & \textbf{195 (0.75)} & 41.60 & 77.46 & 148 (0.81) & 50.00 & 36.45 & \textbf{195 (0.75)} & 47.33 \\
Qwen2.5-7B & Contrastive FLAP & 71.23 & 219 (0.72) & 55.20 & 84.37 & \textbf{132 (0.83)} & 47.06 & 21.27 & 250 (0.68) & 56.00 & 61.61 & \textbf{101 (0.87)} & 50.00 & 70.58 & 258 (0.67) & 70.66 \\
Qwen2.5-7B & Merged & \textbf{82.09} & 303 (0.61) & \textbf{74.38} & \textbf{88.03} & 232 (0.70) & \textbf{67.65} & \textbf{70.92} & 317 (0.59) & \textbf{61.60} & \textbf{93.03} & 206 (0.74) & \textbf{60.00} & \textbf{73.57} & 336 (0.57) & \textbf{79.33} \\
\bottomrule
\end{tabular}%
}
\caption{Comparison of performance, size, and true positive rate (TPR) between Path Patching, FLAP, Contrastive FLAP, and the merged circuits (FLAP + Contrastive FLAP).}
\label{tab:exp1}
\end{table*}


\section{Accelerated Path Patching (APP)}
\label{sec:results}
 Despite the improvements of Contrastive FLAP, Figure \ref{fig:performance_curve} shows that while successfully preserving context-sensitive heads, some circuits depend on task-critical, yet context-insensitive heads that vanilla FLAP preserves. Further, in order to maintain similar performance to Path Patching, both vanilla FLAP and Contrastive-FLAP produce circuit sizes significantly larger than Path Patching, indicating that pruning does not satisfy the minimality constraint required for circuit analysis. Although pruning cannot be substituted for Path Patching since the resulting circuits, it can be used to \textit{Accelerate} Path Patching by reducing the search space of circuit discovery algorithms. Specifically, Accelerated Path Patching (APP) follows a four-step process (See Figure~\ref{fig:app-diagram}): 

\begin{enumerate}
    \item Obtain a vanilla FLAP circuit using cliff points. 
    \item Obtain a Contrastive FLAP circuit using cliff points. 
    \item Merge the circuits found by the vanilla and Contrastive FLAP.
    \item Apply Automated Path Patching on heads included in the merged FLAP circuit.
\end{enumerate}

By merging circuits from vanilla FLAP and Contrastive FLAP, we are able to find task-critical heads that are both context-sensitive and context-insensitive heads in order to preserve highly faithful circuits. Merging the circuits from Step 3 of APP still yields an average 56\% reduction in attention heads and significantly reduces the overall runtime of circuit discovery algorithms. Note that while we apply automated Path Patching as the circuit discovery algorithm in Step 4 of APP, alternative circuit discovery methods that target attention heads as circuit components can be utilized with APP. 

\subsection{Efficient Circuit Discovery with APP}

\begin{figure*}[t]
    \centering
    \begin{subfigure}{0.45\textwidth}
        \centering
        \includegraphics[width=\linewidth]{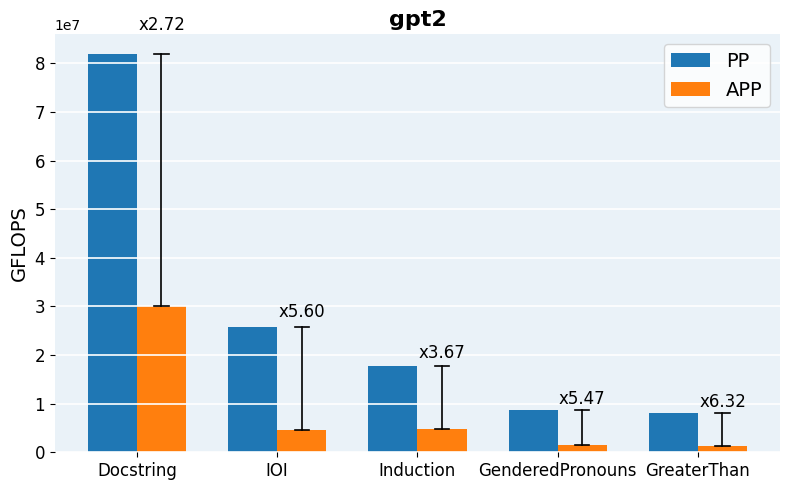}
    \end{subfigure}
    \hfill
    \begin{subfigure}{0.45\textwidth}
        \centering
        \includegraphics[width=\linewidth]{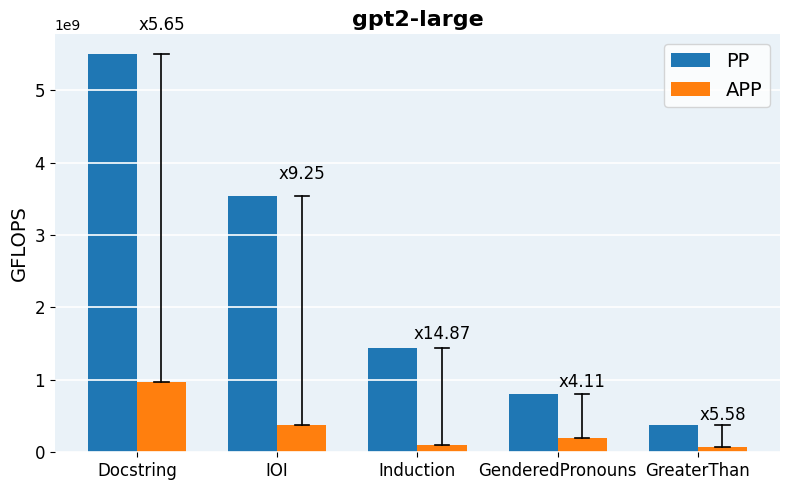}
    \end{subfigure}
    \par\medskip
    
        \begin{subfigure}{0.45\textwidth}
        \centering
        \includegraphics[width=\linewidth]{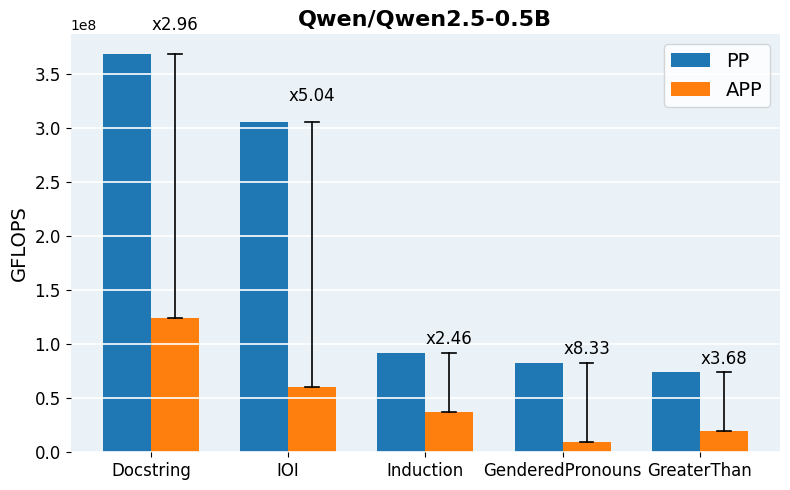}
    \end{subfigure}
    \hfill
    \begin{subfigure}{0.45\textwidth}
        \centering
        \includegraphics[width=\linewidth]{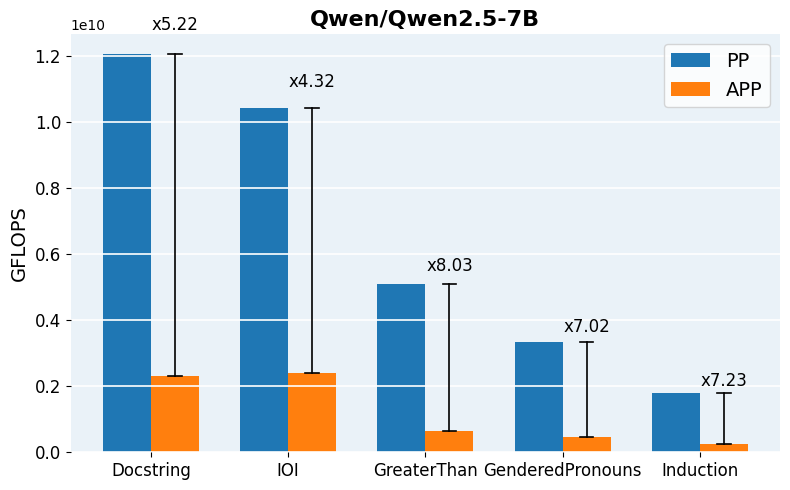}
    \end{subfigure}
    \caption{\centering Difference of required GFLOPs across all models and tasks}
    \label{fig:GFLOPS_efficency}
\end{figure*}

To demonstrate that APP scales to larger models, PP and APP are compared across the GPT-2 small, GPT-2 large, Qwen2.5-0.5B and Qwen2.5-7B model. If available, the smallest circuits with a performance of at least 75\% is chosen in both cases. Both algorithms are executed on 100 random samples from all five tasks. In a preliminary experiment across all five tasks and four models (see Table~\ref{tab:exp4}), circuits for vanilla and Contrastive FLAP are identified via cliff points and merged to form Hybrid FLAP circuits. The resulting Hybrid FLAP circuit maintains a relatively high amount of ground truth heads across all models and tasks. The hybrid circuit reduces the search space by at least 56\%. APP and automated PP are evaluated on two dimensions: accuracy and efficiency. For accuracy, we report performance, circuit size, sparsity, TPR, and precision (P). High TPR and precision indicate close alignment with the PP circuit. Efficiency is measured by required GFLOPs and computation time.

We first analyze the computational efficiency gains of APP over PP. Figure~\ref{fig:GFLOPS_efficency} visualizes the efficiency gains of APP (orange) compared to PP (blue) in terms of GFLOPs. For the large models (GPT-2 large and Qwen2.5-7B) APP requires  \textbf{4.11-14.87x less GFLOPs} than PP and GPT2-small and Qwen2.5-0.5B require \textbf{2.45-8.33 times less GFLOPs}. A similar pattern emerges for the required computation time (see Figure~\ref{fig:time_efficency}). The docstring task consistently requires the most FLOPs for both APP and PP, due to its larger discovered hybrid FLAP circuits, resulting in more PP iterations, and its longer input sequence length (52 tokens, compared to 21 for IOI, the second longest).

\begin{table*}[t]
\centering
\resizebox{\textwidth}{!}{%
\begin{tabular}{ll|cccc|cccc|cccc|cccc|cccc}
\toprule
\textbf{Model} & \textbf{Method} 
& \multicolumn{4}{c}{\textbf{IOI}} 
& \multicolumn{4}{c}{\textbf{GreaterThan}} 
& \multicolumn{4}{c}{\textbf{GenderedPronouns}} 
& \multicolumn{4}{c}{\textbf{Induction}} 
& \multicolumn{4}{c}{\textbf{Docstring}} \\
& & Perf. (\%) & Size & TPR & P 
& Perf. (\%) & Size & TPR & P 
& Perf. (\%) & Size & TPR & P 
& Perf. (\%) & Size & TPR & P 
& Perf. (\%) & Size & TPR & P \\
\midrule

\rowcolor{gray!20} GPT-2 small & PP    & \textbf{96.95} & 21 (0.85) & - & - 
        & \textbf{79.65} & \textbf{8 (0.94)} & - & -
        & \textbf{74.74} & \textbf{5 (0.97)} & - & - 
        & \textbf{93.35} & 22 (0.83) & - & - 
        & \textbf{60.00}  & 36 (0.75) & - & - \\
GPT-2 small & APP   & 82.47 & \textbf{19 (0.87)} &  85.71 & 94.73 
        & 76.71 & 18 (0.88) & 87.5 & 38.89 
        & 74.61  & \textbf{5 (0.97)} & 80.0 & 80.0 
        & 90.35  & \textbf{16 (0.88)} & 68.18 & 93.35 
        & \textbf{56.21} & \textbf{26 (0.82)} & 66.67 & 92.31 \\
\midrule

\rowcolor{gray!20} GPT-2 large & PP    & \textbf{92.60} & 193 (0.73)&  - & - 
        & 75.71 & 27 (0.96) & - & - 
        & 73.94  & 73 (0.89) & - & - 
        & \textbf{90.90} & \textbf{67 (0.91)} &  - & - 
        & \textbf{68.04} & 95 (0.87) & - & - \\
GPT-2 large & APP   & 68.33  & \textbf{92 (0.87)} & 47.66 & 100.0 
        & \textbf{76.89 } & \textbf{23 (0.97)} & 77.78 & 91.30 
        & \textbf{79.65} & \textbf{23 (0.97)} & 34.24 & 100.0 
        & 84.14  & 87 (0.88) & 58.21 & 44.82 
        & 54.36 & \textbf{50 (0.93)} & 51.04 & 98.0 \\
\midrule

\rowcolor{gray!20} Qwen2.5-0.5B & PP    & \textbf{76.33 } & 62 (0.82) & - & - 
        & \textbf{75.14 } & \textbf{20 (0.94)} & - & - 
        & \textbf{86.84 } & \textbf{25 (0.93)} & - & - 
        &  76.08  & \textbf{18 (0.95)} & - & - 
        & 36.04  & 34 (0.90) & - & - \\
Qwen2.5-0.5B & APP   & 76.26  & \textbf{36 (0.89)} & 56.45 & 97.22 
        & 75.05  & 28 (0.92) & 80.0 & 57.14 
        & 60.83  & 33 (0.90) & 44.0 & 33.33 
        & \textbf{84.20 } & 21 (0.94) & 83.33 & 71.43 
        & \textbf{36.63}  & \textbf{32 (0.91)} & 64.71 & 68.75 \\
\midrule

\rowcolor{gray!20} Qwen2.5-7B & PP    & 61.06 & \textbf{121 (0.85)} & - & - 
        & 79.33  & \textbf{34 (0.95)} & - & - 
        & 68.69  & \textbf{125 (0.84)}& - & - 
        & \textbf{67.90 } & 10 (0.99) & - & - 
        & \textbf{58.77 } & 101 (0.87) & - & - \\
Qwen2.5-7B & APP   & \textbf{61.87}  & 137 (0.83) & 74.38 & 53.57 
        & \textbf{79.43 } & 62 (0.92) & 67.65 & 37.10 & 
        \textbf{76.88} & 126 (0.84) & 48.0 & 47.62 
        & 64.98  & \textbf{6 (0.99)} & 40.0 & 66.67 
        & 48.68  & \textbf{79 (0.89)} & 60.40 & 72.49 \\
\midrule \midrule

\rowcolor{gray!20} Average & PP 
    &  \textbf{84.78} & 0.78  & - & - 
    & \textbf{77.46} & \textbf{0.95}  & - & - 
    & \textbf{76.05} & 0.91  & - & -  
    & \textbf{82.06} & \textbf{0.92}  & - & -
    & \textbf{59.66} & 0.81  & - & -  \\
Average & APP 
    & 72.23  & \textbf{0.87} & 66.06 & 86.38 
    & 77.02 & 0.92 & 78.23 & 56.11 
    & 72.99 & \textbf{0.92} & 51.56 & 65.24 
    & 80.92 & \textbf{0.92} & 62.43 & 69.07 
    & 48.96 & \textbf{0.89} & 60.71 & 82.89\\
\bottomrule
\end{tabular}%
}
\caption{Compare the APP and PP method based on their performance (Perf), size (and sparsity), recall (TPR) and precision (P).}
\label{tab:pp-app-task-metric}
\end{table*}

Despite the improved efficiency, Table~\ref{tab:pp-app-task-metric} shows that APP circuits achieve an average performance of 70.42 with a sparsity ratio of 0.9, whereas PP produces circuits with higher performance (76.00) and a slightly lower sparsity ratio of 0.87. When comparing models, the highest average TPR (77.53\%) is obtained with GPT-2 small and the highest average precision (86.22\%) with the GPT-2 large model. 
The chosen set of hyperparameters is not optimal for GPT-2 large. For the IOI and induction task, APP misses some heads, therefor PP results in non-minimal circuits for the other tasks. 

\section{Conclusion and Discussion}
Circuit discovery and pruning share the common goal of identifying subnetworks that replicate the full model’s behavior on downstream tasks. In this study, we first examine whether highly efficient pruning can provide a faster, more computationally efficient alternative to Path Patching. We find that circuits identified through pruning are substantially larger than those obtained via Path Patching, suggesting that pruning alone is insufficient for circuit discovery, as it fails to satisfy the minimality constraints required by circuit analysis. The inability to find minimal circuits is reflected in the limitations of current pruning algorithms at high sparsity levels, with models often experiencing catastrophic performance degradation when pruned beyond 70\% \cite{trim}. Further, we find that even at lower sparsity ratios, methods such as FLAP often prune critical context-sensitive attention heads, such as induction heads that are required for tasks such as IOI. 

We posit that this is largely due to the pruning decision in FLAP being based solely on weights and activations. A high importance score for an attention head does not necessarily imply that the head is critical for solving the task. For instance, \citet{highLayerAttention2025} claims that the position of an attention head is an important factor to identify critical attention heads, not only the attention score. In order to better preserve context-sensitive heads, we propose Contrastive FLAP, which calculates importance scores on the activation differences of contrastive pairs. Contrastive FLAP is able to achieve a significantly sparser model for the same performance as vanilla FLAP, indicating that Contrastive FLAP is a promising standalone pruning method. Figure~\ref{fig:circuit_analysis_IOI} demonstrates that Contrastive FLAP identifies all head types but also includes undefined, task-agnostic heads, indicating that, despite its improvements, Contrastive FLAP alone cannot serve as a replacement for circuit discovery methods.

Although pruning alone cannot replace Path Patching due to the large circuits it produces, Table~\ref{tab:exp1} shows that it achieves a high recall of Path Patching circuits. Building on this, we introduce Accelerated Path Patching (APP), which uses Contrastive-FLAP to reduce the search space before applying any circuit discovery algorithm. APP maintains high task performance while substantially improving efficiency, enabling circuit discovery methods to scale to larger models. Table~\ref{tab:pp-app-task-metric} illustrates that APP can achieve high precision even when TPR is low, indicating that it successfully removes redundant heads and identifies more compact circuits. In other cases, high TPR with lower precision suggests that minimal circuits remain elusive. Overall, despite the significant gains in computational efficiency, APP often produces circuits of similar size and performance compared to Path Patching. By serving as a general preprocessing step, APP makes any circuit discovery method faster and more practical, offering a scalable path toward efficient mechanistic interpretability.

\section{Limitations}
Table~\ref{tab:pp-app-task-metric} shows that neither automated PP nor APP consistently yields optimal circuits in terms of minimality and performance. One reason is that the hyperparamter search is limited by the computational cost of PP on large models like Qwen2.5-7B. Exploring a wider range of hyperparameters solely on APP could find better circuits across all models and tasks. Another challenge is deciding the best trade-off between circuit size and performance. While curve fitting or Pareto analysis can help, the final decision remains the responsibility of the researcher. Furthermore, the ability of APP to retrieve the same circuits as PP is capped by the TPR of hybrid FLAP. If the hybrid FLAP circuit is too small, APP will not recover all heads. Our work focuses on circuit discovery and not circuit analysis. To the best of our knowledge, except for GPT-2 small, path patching has not been applied to the here tested large models, and thus comparison to other work is not possible. Studying whether circuits discovered in GPT-2 small translate to larger models, or if new behavior arises, would be very interesting. APP, as proposed here, focuses on attention heads, since various related works in mechanistic interpretability make the same choice. Nonetheless, some natural language tasks, such as factual association \cite{ROME2022}, rely heavily on the MLP components. Since FLAP supports MLP patching, these components could be incorporated into the automated path patching method if desired. The heuristic of reducing the search space via pruning is not limited to PP. Other algorithms, such as ACDC could be enhanced in the same way, but this is left for future work. 

\section{Acknowledgments}
This work was supported by the National Science Foundation under Cooperative
Agreement 2421782 and the Simons Foundation award MPS-AI-00010515
(NSF-Simons AI Institute for Cosmic Origins - CosmicAI, https://www.cosmicai.org/

\bibliography{custom}

@misc{AutomaticPP2024,
      title={Towards Best Practices of Activation Patching in Language Models: Metrics and Methods}, 
      author={Fred Zhang and Neel Nanda},
      year={2024},
      eprint={2309.16042},
      archivePrefix={arXiv},
      primaryClass={cs.LG},
      url={https://arxiv.org/abs/2309.16042}, 
}

@misc{wang2022interpretabilitywildcircuitindirect,
      title={Interpretability in the Wild: a Circuit for Indirect Object Identification in GPT-2 small}, 
      author={Kevin Wang and Alexandre Variengien and Arthur Conmy and Buck Shlegeris and Jacob Steinhardt},
      year={2022},
      eprint={2211.00593},
      archivePrefix={arXiv},
      primaryClass={cs.LG},
      url={https://arxiv.org/abs/2211.00593}, 
}

@article{FLAP2024, 
title={Fluctuation-Based Adaptive Structured Pruning for Large Language Models}, 
volume={38}, 
url={https://ojs.aaai.org/index.php/AAAI/article/view/28960},
DOI={10.1609/aaai.v38i10.28960},
abstractNote={Network Pruning is a promising way to address the huge computing resource demands of the deployment and inference of Large Language Models (LLMs). Retraining-free is important for LLMs’ pruning methods. However, almost all of the existing retraining-free pruning approaches for LLMs focus on unstructured pruning, which requires specific hardware support for acceleration. In this paper, we propose a novel retraining-free structured pruning framework for LLMs, named FLAP (FLuctuation-based Adaptive
Structured Pruning). It is hardware-friendly by effectively reducing storage and enhancing inference speed. For effective structured pruning of LLMs, we highlight three critical elements that demand the utmost attention: formulating structured importance metrics, adaptively searching the global compressed model, and implementing compensation mechanisms to mitigate performance loss. First, FLAP determines whether the output feature map is easily recoverable when a column of weight is removed, based on the fluctuation pruning metric. Then it standardizes the importance scores to adaptively determine the global compressed model structure. At last, FLAP adds additional bias terms to recover the output feature maps using the baseline values. We thoroughly evaluate our approach on a variety of language benchmarks. Without any retraining, our method significantly outperforms the state-of-the-art methods, including LLM-Pruner and the extension of Wanda in structured pruning. The code is released at https://github.com/CASIA-IVA-Lab/FLAP.}, 
number={10}, 
journal={Proceedings of the AAAI Conference on Artificial Intelligence},
author={An, Yongqi and Zhao, Xu and Yu, Tao and Tang, Ming and Wang, Jinqiao},
year={2024}, 
month={Mar.},
pages={10865-10873} }

@misc{induction,
      title={Localizing Model Behavior with Path Patching}, 
      author={Nicholas Goldowsky-Dill and Chris MacLeod and Lucas Sato and Aryaman Arora},
      year={2023},
      eprint={2304.05969},
      archivePrefix={arXiv},
      primaryClass={cs.LG},
      url={https://arxiv.org/abs/2304.05969}, 
}

@InProceedings{sparseGPT,
  title = 	 {{S}parse{GPT}: Massive Language Models Can be Accurately Pruned in One-Shot},
  author =       {Frantar, Elias and Alistarh, Dan},
  booktitle = 	 {Proceedings of the 40th International Conference on Machine Learning},
  pages = 	 {10323--10337},
  year = 	 {2023},
  editor = 	 {Krause, Andreas and Brunskill, Emma and Cho, Kyunghyun and Engelhardt, Barbara and Sabato, Sivan and Scarlett, Jonathan},
  volume = 	 {202},
  series = 	 {Proceedings of Machine Learning Research},
  month = 	 {23--29 Jul},
  publisher =    {PMLR},
  url = 	 {https://proceedings.mlr.press/v202/frantar23a.html}
}

@inproceedings{LLMPruner2023,
 author = {Ma, Xinyin and Fang, Gongfan and Wang, Xinchao},
 booktitle = {Advances in Neural Information Processing Systems},
 editor = {A. Oh and T. Naumann and A. Globerson and K. Saenko and M. Hardt and S. Levine},
 pages = {21702--21720},
 publisher = {Curran Associates, Inc.},
 title = {LLM-Pruner: On the Structural Pruning of Large Language Models},
 url = {https://proceedings.neurips.cc/paper_files/paper/2023/file/44956951349095f74492a5471128a7e0-Paper-Conference.pdf},
 volume = {36},
 year = {2023}
}

@article{GenderedPronouns2023,
  title={Identifying a preliminary circuit for predicting gendered pronouns in gpt-2 small},
  author={Mathwin, Chris and Corlouer, Guillaume and Kran, Esben and Barez, Fazl and Nanda, Neel},
  journal={URL: https://itch. io/jam/mechint/rate/1889871},
  year={2023}
}

@article{zoomIn2020,
  author = {Olah, Chris and Cammarata, Nick and Schubert, Ludwig and Goh, Gabriel and Petrov, Michael and Carter, Shan},
  title = {Zoom In: An Introduction to Circuits},
  journal = {Distill},
  year = {2020},
  note = {https://distill.pub/2020/circuits/zoom-in},
  doi = {10.23915/distill.00024.001}
}

@article{elhage2021mathematical,
  title={A mathematical framework for transformer circuits},
  author={Elhage, Nelson and Nanda, Neel and Olsson, Catherine and Henighan, Tom and Joseph, Nicholas and Mann, Ben and Askell, Amanda and Bai, Yuntao and Chen, Anna and Conerly, Tom and others},
  journal={Transformer Circuits Thread},
  volume={1},
  number={1},
  pages={12},
  year={2021}
}

@misc{Docstring2023,
    title={A circuit for Python docstrings in a 4-layer attention-only}, 
    author={Heimersheim, Stefan and Janiak, Jett}, 
    url={https://www.alignmentforum.org/posts/u6KXXmKFbXfWzoAXn/acircuit-for-python-docstrings-in-a-4-layer-attention-only.}
}

@misc{acdc2023,
      title={Towards Automated Circuit Discovery for Mechanistic Interpretability}, 
      author={Arthur Conmy and Augustine N. Mavor-Parker and Aengus Lynch and Stefan Heimersheim and Adrià Garriga-Alonso},
      year={2023},
      eprint={2304.14997},
      archivePrefix={arXiv},
      primaryClass={cs.LG},
      url={https://arxiv.org/abs/2304.14997}, 
}

@article{meng2022locating,
  title={Locating and Editing Factual Associations in {GPT}},
  author={Kevin Meng and David Bau and Alex Andonian and Yonatan Belinkov},
  journal={Advances in Neural Information Processing Systems},
  volume={36},
  year={2022},
  note={arXiv:2202.05262}
}

@inproceedings{
eap-ig2024,
title={Have Faith in Faithfulness: Going Beyond Circuit Overlap When Finding Model Mechanisms},
author={Michael Hanna and Sandro Pezzelle and Yonatan Belinkov},
booktitle={ICML 2024 Workshop on Mechanistic Interpretability},
year={2024},
url={https://openreview.net/forum?id=grXgesr5dT}
}

@misc{vig2020causalmediationanalysisinterpreting,
      title={Causal Mediation Analysis for Interpreting Neural NLP: The Case of Gender Bias}, 
      author={Jesse Vig and Sebastian Gehrmann and Yonatan Belinkov and Sharon Qian and Daniel Nevo and Simas Sakenis and Jason Huang and Yaron Singer and Stuart Shieber},
      year={2020},
      eprint={2004.12265},
      archivePrefix={arXiv},
      primaryClass={cs.CL},
      url={https://arxiv.org/abs/2004.12265}, 
}

@misc{goldowskydill2023localizingmodelbehaviorpath,
      title={Localizing Model Behavior with Path Patching}, 
      author={Nicholas Goldowsky-Dill and Chris MacLeod and Lucas Sato and Aryaman Arora},
      year={2023},
      eprint={2304.05969},
      archivePrefix={arXiv},
      primaryClass={cs.LG},
      url={https://arxiv.org/abs/2304.05969}, 
}

@misc{notice,
      title={What Do VLMs NOTICE? A Mechanistic Interpretability Pipeline for Gaussian-Noise-free Text-Image Corruption and Evaluation}, 
      author={Michal Golovanevsky and William Rudman and Vedant Palit and Ritambhara Singh and Carsten Eickhoff},
      year={2025},
      eprint={2406.16320},
      archivePrefix={arXiv},
      primaryClass={cs.CL},
      url={https://arxiv.org/abs/2406.16320}, 
}

@misc{trim,
      title={TRIM: Achieving Extreme Sparsity with Targeted Row-wise Iterative Metric-driven Pruning}, 
      author={Florentin Beck and William Rudman and Carsten Eickhoff},
      year={2025},
      eprint={2505.16743},
      archivePrefix={arXiv},
      primaryClass={cs.CL},
      url={https://arxiv.org/abs/2505.16743}, 
}

@misc{highLayerAttention2025,
      title={High-Layer Attention Pruning with Rescaling}, 
      author={Songtao Liu and Peng Liu},
      year={2025},
      eprint={2507.01900},
      archivePrefix={arXiv},
      primaryClass={cs.CL},
      url={https://arxiv.org/abs/2507.01900}, 
}

@article{PruningSurveyZhu2024,
    author = {Zhu, Xunyu and Li, Jian and Liu, Yong and Ma, Can and Wang, Weiping},
    title = {A Survey on Model Compression for Large Language
                    Models},
    journal = {Transactions of the Association for Computational Linguistics},
    volume = {12},
    pages = {1556-1577},
    year = {2024},
    month = {11},
    issn = {2307-387X},
    doi = {10.1162/tacl_a_00704},
    url = {https://doi.org/10.1162/tacl_a_00704},
    eprint = {https://direct.mit.edu/tacl/article-pdf/doi/10.1162/tacl_a_00704/2482209/tacl_a_00704.pdf},
}

@inproceedings{GreaterThan,
 author = {Hanna, Michael and Liu, Ollie and Variengien, Alexandre},
 booktitle = {Advances in Neural Information Processing Systems},
 editor = {A. Oh and T. Naumann and A. Globerson and K. Saenko and M. Hardt and S. Levine},
 pages = {76033--76060},
 publisher = {Curran Associates, Inc.},
 title = {How does GPT-2 compute greater-than?: Interpreting mathematical abilities in a pre-trained language model},
 url = {https://proceedings.neurips.cc/paper_files/paper/2023/file/efbba7719cc5172d175240f24be11280-Paper-Conference.pdf},
 volume = {36},
 year = {2023}
}

@misc{WANDA2024,
      title={A Simple and Effective Pruning Approach for Large Language Models}, 
      author={Mingjie Sun and Zhuang Liu and Anna Bair and J. Zico Kolter},
      year={2024},
      eprint={2306.11695},
      archivePrefix={arXiv},
      url={https://arxiv.org/abs/2306.11695}, 
}

@misc{LayerDrop2019,
      title={Reducing Transformer Depth on Demand with Structured Dropout}, 
      author={Angela Fan and Edouard Grave and Armand Joulin},
      year={2019},
      eprint={1909.11556},
      archivePrefix={arXiv},
      primaryClass={cs.LG},
      url={https://arxiv.org/abs/1909.11556}, 
}

@inproceedings{ROME2022,
 author = {Meng, Kevin and Bau, David and Andonian, Alex and Belinkov, Yonatan},
 booktitle = {Advances in Neural Information Processing Systems},
 editor = {S. Koyejo and S. Mohamed and A. Agarwal and D. Belgrave and K. Cho and A. Oh},
 pages = {17359--17372},
 publisher = {Curran Associates, Inc.},
 title = {Locating and Editing Factual Associations in GPT},
 url = {https://proceedings.neurips.cc/paper_files/paper/2022/file/6f1d43d5a82a37e89b0665b33bf3a182-Paper-Conference.pdf},
 volume = {35},
 year = {2022}
}

@article{rutherford1905xxxvii,
  title={XXXVII. Slow transformation products of radium},
  author={Rutherford, E},
  journal={The London, Edinburgh, and Dublin Philosophical Magazine and Journal of Science},
  volume={10},
  number={57},
  pages={290--306},
  year={1905},
  publisher={Taylor \& Francis}
}

@misc{EAP2023,
      title={Attribution Patching Outperforms Automated Circuit Discovery}, 
      author={Aaquib Syed and Can Rager and Arthur Conmy},
      year={2023},
      eprint={2310.10348},
      archivePrefix={arXiv},
      primaryClass={cs.LG},
      url={https://arxiv.org/abs/2310.10348}, 
}

@inproceedings{Radford2019LanguageMA,
  title={Language Models are Unsupervised Multitask Learners},
  author={Alec Radford and Jeff Wu and Rewon Child and David Luan and Dario Amodei and Ilya Sutskever},
  year={2019},
  url={https://api.semanticscholar.org/CorpusID:160025533}
}

@misc{olah2022mechanistic,
  author       = {Chris Olah},
  title        = {Mechanistic Interpretability, Variables, and the Importance of Interpretable Bases},
  year         = {2022},
  howpublished = {\url{https://www.transformer-circuits.pub/2022/mech-interp-essay}},
  note         = {Published June 27, 2022; Accessed: 2025-10-02}
}

@article{Frankle2018TheLT,
  title={The Lottery Ticket Hypothesis: Finding Sparse, Trainable Neural Networks},
  author={Jonathan Frankle and Michael Carbin},
  journal={arXiv: Learning},
  year={2018},
  url={https://api.semanticscholar.org/CorpusID:53388625}
}

@inproceedings{Geiger2021CausalAO,
  title={Causal Abstractions of Neural Networks},
  author={Atticus Geiger and Hanson Lu and Thomas F. Icard and Christopher Potts},
  booktitle={Neural Information Processing Systems},
  year={2021},
  url={https://api.semanticscholar.org/CorpusID:235358214}
}

@misc{logit_lens,
  author = {Nostalgebraist},
  title = {Interpreting GPT: the logit lens},
  url = {https://www.lesswrong.com/posts/AcKRB8wDpdaN6v6ru/interpreting-gpt-the-logit-lens},
  year = {2020},
}

@article{Zhang2024TheSB,
  title={The Same But Different: Structural Similarities and Differences in Multilingual Language Modeling},
  author={Ruochen Zhang and Qinan Yu and Matianyu Zang and Carsten Eickhoff and Ellie Pavlick},
  journal={ArXiv},
  year={2024},
  volume={abs/2410.09223},
  url={https://api.semanticscholar.org/CorpusID:273346857}
}

@article{Yu2023CharacterizingMF,
  title={Characterizing Mechanisms for Factual Recall in Language Models},
  author={Qinan Yu and Jack Merullo and Ellie Pavlick},
  journal={ArXiv},
  year={2023},
  volume={abs/2310.15910},
  url={https://api.semanticscholar.org/CorpusID:264439114}
}

@article{Yang2024Qwen25TR,
  title={Qwen2.5 Technical Report},
  author={Qwen An Yang and Baosong Yang and Beichen Zhang and Binyuan Hui and Bo Zheng and Bowen Yu and Chengyuan Li and Dayiheng Liu and Fei Huang and Guanting Dong and Haoran Wei and Huan Lin and Jian Yang and Jianhong Tu and Jianwei Zhang and Jianxin Yang and Jiaxin Yang and Jingren Zhou and Junyang Lin and Kai Dang and Keming Lu and Keqin Bao and Kexin Yang and Le Yu and Mei Li and Mingfeng Xue and Pei Zhang and Qin Zhu and Rui Men and Runji Lin and Tianhao Li and Tingyu Xia and Xingzhang Ren and Xuancheng Ren and Yang Fan and Yang Su and Yi-Chao Zhang and Yunyang Wan and Yuqi Liu and Zeyu Cui and Zhenru Zhang and Zihan Qiu and Shanghaoran Quan and Zekun Wang},
  journal={ArXiv},
  year={2024},
  volume={abs/2412.15115},
  url={https://api.semanticscholar.org/CorpusID:274859421}
}
\appendix
\section{Tasks}
\label{sec:tasks}

\subsection{IOI Task}
The IOI task \cite{wang2022interpretabilitywildcircuitindirect} evaluates whether a model can correctly predict the indirect object in a sentence. For example, a clean prompt is \textit{``When John and Mary went to the store, John bought a drink for...''}. John is the subject of the sentence and the correct continuation; \textit{``Mary''} is the indirect object. A corrupted prompt removes the semantic cue from the indirect object, for example, \textit{``When John and Mary went to the store, Alex bought a drink for...''}. 

Task performance is measured using the logit difference, defined as the logit of the indirect object minus the logit of the subject. In the example above: $$\text{LD = Logit(Mary) - Logit(John)}$$

We use the implementation of the IOI dataset from the ARENA tutorial repository\footnote{\url{https://github.com/callummcdougall/ARENA_3.0/blob/main/chapter1_transformer_interp/exercises/part41_indirect_object_identification/ioi_dataset.py}}. The corrupted dataset is generated using the ''ABB$\rightarrow$XYZ, BAB$\rightarrow$XYZ'' command. 

\subsection{GreaterThan Task}
The objective of the GreaterThan task \cite{GreaterThan} is to evaluate the model's ability to predict a number greater than another within a natural language context.
The template is: \textit{``The <\textsc{NOUN}> lasted from XXYY to the year XX..''}, where XX $\in$ \{11,..., 17\} and YY $\in$ \{02,..., 98\}. The model is supposed to predict a number $\geq$ YY. The corrupted prompt is of the form: \textit{``The <\textsc{NOUN}> lasted from XX01 to the year XX..''}. 

The tokenizer of the GPT family encodes YY as [YY], and it predicts answer $v$ as one token. Thus, the logit difference is as follows.
\begin{align*}
&\text{Correct} =\{v|( v > YY) \land (v \leq 98)\} \\
&\text{Wrong} = \{v|v \leq YY\} \\
&\text{LD} = \text{Logits(Correct) - Logits(Wrong)}
\end{align*}

The Qwen2.5 tokenizer encodes YY in two tokens [Y$_1$][Y$_2$]. The predicted answer is tokenized in the same manner [$v_1$][$v_2$]. With $Y_1, Y_2, v1, v2 = 0,\ldots 9$ Hence the average logit difference is 
\begin{align*}
    &\text{Correct} =\\
    &\{v1v2|(v1 > Y1) \lor (v1 = Y1 \land v2 > Y2)\} &&\\
    &\text{Wrong} =\\ 
    &\{v1v2|(v1 < Y1) \lor (v1 = Y1 \land v2 \leq Y2)\} &&\\
    &\text{LD} = \text{Logits(Correct) - Logits(Wrong)}&&
\end{align*}

The GreaterThan Dataset is downloaded from the original HuggingFace data repo\footnote{\url{https://huggingface.co/datasets/mwhanna/greater-than}}.

\subsection{GenderedPronouns Task}
The GenderedPronouns Task~\citep{GenderedPronouns2023} evaluates a model’s ability to predict the pronoun conventionally associated with a gendered name. While exceptions to these associations exist, we adopt this simplification for clarity.

A clean sample from the dataset is \textit{``So Emily is such a good friend, isn't...''} with the correct answer \textit{she}. The corresponding corrupted prompt replaces the gendered name with a non-gendered description:  \textit{``That person is such a good friend, isn't...''}. The clean prompt begins with the filler word \textit{``So''} to preserve grammatical cohesion in the corrupted version, while maintaining structural equivalence between the two prompts. Without \textit{``So''}, the corrupted prompt would be one token longer, requiring the removal of \textit{``That''} in the corrupted prompt.

The average logit difference is calculated by the difference between the correct pronoun and the wrong pronoun for the name. In the given example, that is $\text{LD = Logits(she) - Logits(he)}$.

\subsection{Induction Task}
The induction task uses token sequences of the form [A][B]...[A], where the model's objective is to predict [B] as the next token. Our implementation differs from~\citet{induction}, by focusing on name induction. For example, a clean prompt has the following form: \textit{``Today, Claire visited the library. There Cl...''}. The correct next prediction is \textit{``aire''}. The corrupted prompt replaces the repeated name with another, e.g. \textit{``Today, Claire visited the library. There Tr...''}. It is important that only names that the tokenizer splits into exactly two tokens [Name$_1$][Name$_2$] are used. 

The average logit difference is defined as the difference between the logits of the second token of the correct name and the second token of the corrupted name: $\text{LD = Logits(aire) - Logits(istan)}$.

\subsection{Docstring Task}
The docstring task~\citep{Docstring2023} evaluates a model's ability to complete a Python Docstring. A clean prompt has the following form: 

\begin{minipage}{.45\textwidth}
\parbox{\linewidth}{
\textsc{def old(self, first, page, names, size, files, read):}\\
\texttt{"""sector gap population}\\
\texttt{:param page: message tree}\\
\texttt{:param names: detail mine}\\
\texttt{:param ..."""}\\
}
\end{minipage}

The correct next token is \textsc{size}. The corresponding corrupted prompt is the following.

\begin{minipage}{.45\textwidth}
\parbox{\linewidth}{
\textsc{def old(self, first, project, target, new, files, read):}\\
\texttt{"""sector gap population}\\
\texttt{:param image: message tree}\\
\texttt{:param update: detail mine}\\
\texttt{:param ..."""}\\
}
\end{minipage}

The logit difference is the difference between the logits of the correct answer and max logit of all wrong variables of the clean and corrupted prompt.
{\small
\begin{align*}
    &\text{Wrong} = \\ 
    & \{\text{first, page, names, files, read, project, target, new}\} \\
 & \text{LD} = \text{Logits({size}})-\mathop{\arg \max}\limits_{x \in \text{Wrong}} \text{Logits}(x)
\end{align*}
}%

The code for the dataset is taken from this repository\footnote{\url{https://github.com/jettjaniak/mi_utils_public/blob/main/prompts.py}}.

\section{Context Sensitive vs. Task-Critical Heads}
\paragraph{Context-Sensitive Attention Heads}
\label{sec:context_sensitive}
\citet{meng2022locating} identify various types of attention heads for the IOI task in the GPT-2 small model. For example, the \textit{Previous Token Heads} always attend to the token immediately preceding the current one. These heads exhibit the same activation pattern regardless of the provided input and are therefore called \textbf{context-insensitive}. Figure \ref{fig:hybridFLAP_activations} visualizes the activation pattern of a previous token head on the clean (a) and corrupted (b) dataset. Although the token input of the two datasets differs, the observed activation pattern is the same. 

Conversely, heads that display a distinct activation pattern only when a specific token pattern is present, are \textbf{context-sensitive}. For example, an \textit{Induction Head} strongly attends to the token \textsc{[B]} when it encounters the token sequence \textsc{[A][B] ... [A]}. By definition, context-sensitive heads should remain inactive on samples from the corrupted dataset, because the token pattern is not detected (see Figure \ref{fig:hybridFLAP_activations}, (d) + (e)). Note that context-sensitive heads are activated only by specific token context, whereas task-critical heads are defined as essential for solving a task and are ideally included in the corresponding circuit. Many, but not all, task-critical heads are context-sensitive (e.g. previous token heads).

\section{Automatic Path Patching}
\label{sec:automatic_PP}
\subsection{Implementation}
The procedure to automate path patching is summarized in Algorithm \ref{alg:AutomaticPP} and is heavily inspired by~\citet{AutomaticPP2024}. \textsc{PathPatching} encodes the standard path patching mechanism introduced by~\citet{wang2022interpretabilitywildcircuitindirect}. In the \textsc{EvaluateThreshold} function, two thresholds are decided based on a head-wise influence score to determine whether an attention head is included in the circuit. 

To discuss the threshold, we will first provide some notations. Assume that the total number of layers of a model is $L$ and that each layer contains $H$ attention heads. A receiver head $r$ can be defined by the tuple $r = (l_r, h_r)$, where $l_r \leq L$ and $h_r \leq H$. Sender heads are denoted as $s = (l_s, h_s)$ and correspond to all heads from earlier layers ($l_s < l_r$) that patch to $r$. The influence of patching a head from earlier layers is measured by an influence score, typically the \textit{average logit difference}. The influence scores of all previous heads are collected in the matrix $X \in \mathbb{R}^{(l_r - 1) \times H}$. 

\begin{algorithm}[t]
\DontPrintSemicolon
\LinesNumbered
\caption{Automatic Path Patching}
\label{alg:AutomaticPP}
\SetKwFunction{PathPatching}{PathPatching}
\SetKwFunction{EvaluateThresholds}{EvaluateThresholds}
\SetKwData{influenceScores}{influenceScores}
\SetKwData{circuit}{circuit}
\SetKwData{receiverList}{receiverList}
\SetKwData{receiver}{receiver}
\SetKwData{senderList}{senderList}
\KwResult{\circuit}

\BlankLine
\circuit $\gets \{\}$\;
\influenceScores $\gets$ \PathPatching(logits)  \tcp*[r]{\influenceScores$\in \mathbb{R}^{(L \times H)}$}
\receiverList $\gets$ \EvaluateThresholds(\influenceScores) \;
\BlankLine
\While{\receiverList not empty}{
    \receiver $\gets$ pop(\receiverList) \tcp*[r]{layer l, head h}
    \circuit.add(\receiver)\;
    \influenceScores $\gets $ \PathPatching(\receiver) \tcp*[r]{\influenceScores $\in \mathbb{R}^{((l-1) \times H)}$} 
    \senderList $\gets $  \EvaluateThresholds(\influenceScores)\;
    \BlankLine
    \For{s in \senderList}{
        \If{s not in \receiverList $\And$ s not in \circuit}{
            receiverList.add(s) \;
        }
    }
}
\end{algorithm}
First, the \textbf{importance threshold} is evaluated over each importance score $x_{i, j}$ 
$$|x_{i, j}| - |\bar{X}| > K * \mathbf{SD}(X)$$,
where $\bar{X}$ and $\mathbf{SD}(X)$ denote the mean and standard deviation calculated over all influence scores respectively, and $K$ is a predefined constant.
At lower-level layers, the scores become noisier. To account for this, the constant is adjusted to $$K' = K + \frac{2}{\sqrt{l_s * H}}$$, where $K'$ increases as $l_s$ decreases. The adjusted constant $K'$ is evaluated for each sender layer $l_s$. This adjustment is desirable, since path patching focuses on sequential influence: components that are more distant from the receiver must meet stricter significance criteria to be included. For evaluation, K is set to $[1, 1.5, 2, 2.5]$.

Second, the \textbf{maximum value thresholds} ensure that at least one score $x_{i,j}$ is above a chosen value $\epsilon$. Otherwise, no head for the current receiver will be included in the circuit. This prevents the inclusion of spurious or weakly contributing components in the constructed circuit. Values chosen for $\epsilon$ are $[0.01, 0.001, 0.02, 0.002]$

Heads that exceed both thresholds are considered significant contributors to the model's behavior and are selected as receivers in subsequent iterations of the algorithm.

\label{appendix:B}
\begin{table*}[t]
\centering
\resizebox{\textwidth}{!}{%
\begin{tabular}{l|cc|cc|cc|cc|cc}
\toprule
& \multicolumn{2}{c|}{\textbf{IOI}} 
& \multicolumn{2}{c|}{\textbf{GreaterThan}} 
& \multicolumn{2}{c|}{\textbf{GenderedPronouns}} 
& \multicolumn{2}{c|}{\textbf{Induction}} 
& \multicolumn{2}{c}{\textbf{Docstring}} \\
& Manual & Automatic & Manual & Automatic & Manual & Automatic & Manual & Automatic & Manual & Automatic \\
\hline \hline
Perf. (\%)  & 75.58 & \textbf{96.95} & 71.68 & \textbf{79.65} & \textbf{80.16} & 74.74 & 86.34 & \textbf{93.35} & 41.06 & \textbf{60.36} \\

Size         & \textbf{15}    & 21    & \textbf{8}     & 8    & \textbf{7}     & 5     & \textbf{14}    & 22    & \textbf{23}    & 36    \\
Intersection & -     & 14    & -     & 6     & -     & 4     & -     & 13    & -     & 22    \\
Difference   & -     & 7     & -     & 2     & -     & 1     & -     & 9     & -     & 14     \\
\bottomrule
\end{tabular}
}
\caption{Circuits discovered by manual and automatic path patching on GPT-2 small}
\end{table*}

\section{Pruning}
\label{sec:vanilla FLAP}
\subsection{Implementation of Vanilla FLAP}
Since FLAP \citep{FLAP2024} is a zero-shot pruning algorithm that utilizes head-wise importance scores and does not require retraining it is possible to have a direct comparison with path patching. The influence scores are calculated by passing samples of the dataset $\mathcal{D}_{clean}$ to the FLAP algorithm. Based on a sparsity level $p$, the $1 - p$\% heads with the highest importance score are chosen to be in the circuit. By design, heads that yield in higher importance scores have higher weights and/or input activation values during the forward pass. Theoretically, these higher values translate to a higher influence on the output logits. Thus, these heads are likely to have a stronger influence on the next prediction and are considered to be in the circuit. 

Since FLAP is computationally very efficient, it can be run on multiple times to evaluate the performance of the resulting circuits under different sparsity levels. 

\subsubsection{Implementation of Contrastive FLAP}
Contrastive FLAP is summarized in Algorithm \ref{algo:contrastiveFLAP}. The method computes FLAP scores for both clean and corrupted inputs, then evaluates the absolute difference between the two. Attention heads are subsequently ranked globally according to their contrastive FLAP scores, and the top s\% with the highest ratios are selected and incorporated into the circuit.

\begin{algorithm}[t]
\DontPrintSemicolon
\LinesNumbered
\caption{Contrastive FLAP}
\SetKwFunction{calculateImportanceScore}{calculateImportanceScore}
\SetKwFunction{pruneHeads}{pruneHeads}
\SetKwData{FLAPScores}{FLAPScores}
\SetKwData{corruptedFLAPScores}{corruptedFLAPScores}
\SetKwData{contrastiveFLAPScores}{contrastiveFLAPScores}
\SetKwData{circuit}{circuit}

\KwResult{\circuit}

\BlankLine
\FLAPScores $\gets$ \calculateImportanceScore(cleanDataset) \\
\corruptedFLAPScores $\gets$ \calculateImportanceScore(corruptedDataset) \\
\contrastiveFLAPScores $\gets  \mid \mid$\FLAPScores - \corruptedFLAPScores$\mid \mid$ \\
\circuit $\gets$ pruneHeads(\contrastiveFLAPScores, sparsityRatio);
\label{algo:contrastiveFLAP}
\end{algorithm}

\subsubsection{Does Contrastive FLAP find smaller circuits than vanilla FLAP?}
\label{sec:SubtractedFLAP}

\noindent\textbf{Methods}
Contrastive and Vanilla FLAP are evaluated on the GPT-2 small model for all five tasks presented in \ref{sec:relatedWork}. Each task had a clean and corrupted dataset with 200 randomly picked samples. 

\noindent\textbf{Metric}
The \textbf{half-life} metric ($t_\frac{1}{2}$), originally used to quantify the exponential nature of radioactive decay  \cite{rutherford1905xxxvii}, measures the time required for a quantity to reach half of its initial value. Here, the metric is applied to the total number of true positive heads, with sparsity levels ranging from 0 to 1 in increments of 0.01 serving as the analog of time. A higher half-life value translates to a later exclusion of critical heads and thus an earlier exclusion of task-irrelevant heads.

\begin{figure*}[t]
    \centering
    \begin{minipage}{1\textwidth}
        \centering
        \begin{subfigure}{0.45\textwidth}
            \centering
            \includegraphics[width=\linewidth]{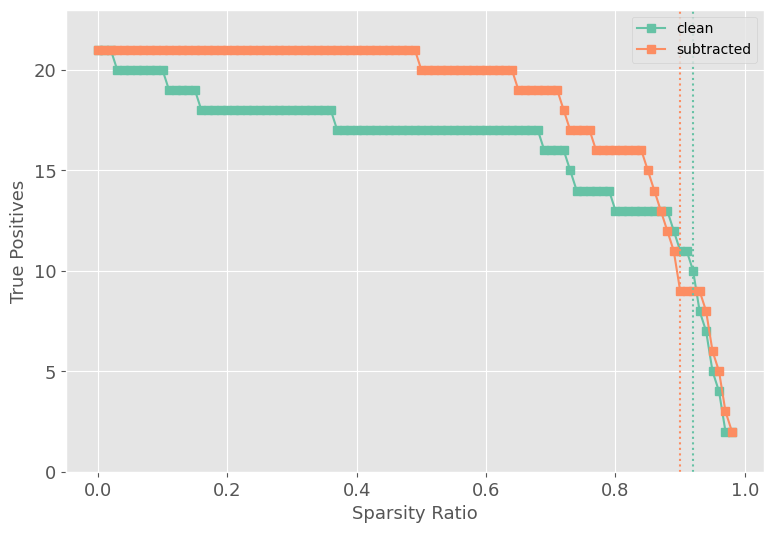}
            \caption{IOI task}
        \end{subfigure} \qquad
        \begin{subfigure}{0.45\textwidth}
            \centering
            \includegraphics[width=\linewidth]{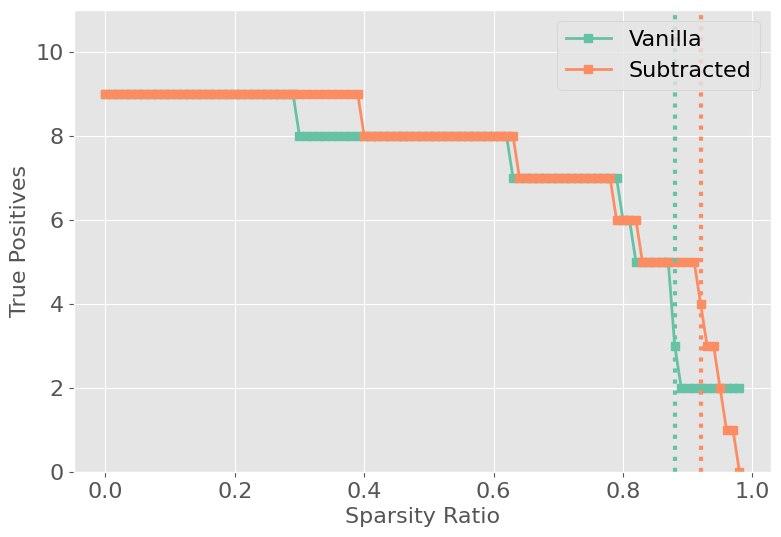}
            \caption{GreaterThan task}
        \end{subfigure} \\
        \par\medskip
        \begin{subfigure}{0.45\textwidth}
            \centering
            \includegraphics[width=\linewidth]{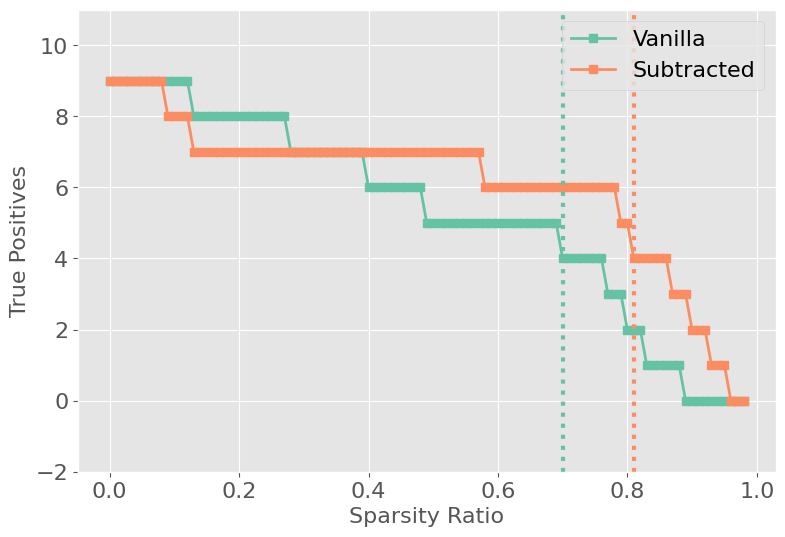}
            \caption{GenderedPronouns task}
        \end{subfigure} \qquad
        \begin{subfigure}{0.45\textwidth}
            \centering
            \includegraphics[width=\linewidth]{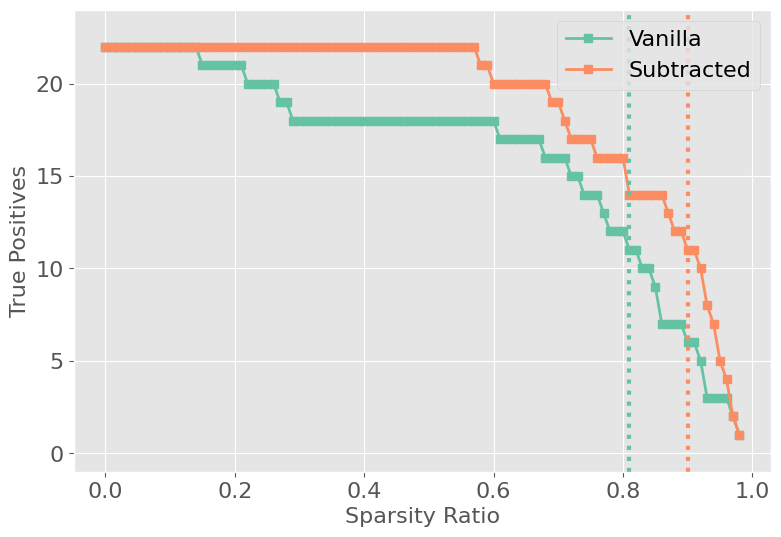}
            \caption{Induction task}
        \end{subfigure} \\
        \par\medskip
        \begin{subfigure}{0.45\textwidth}
            \centering
            \includegraphics[width=\linewidth]{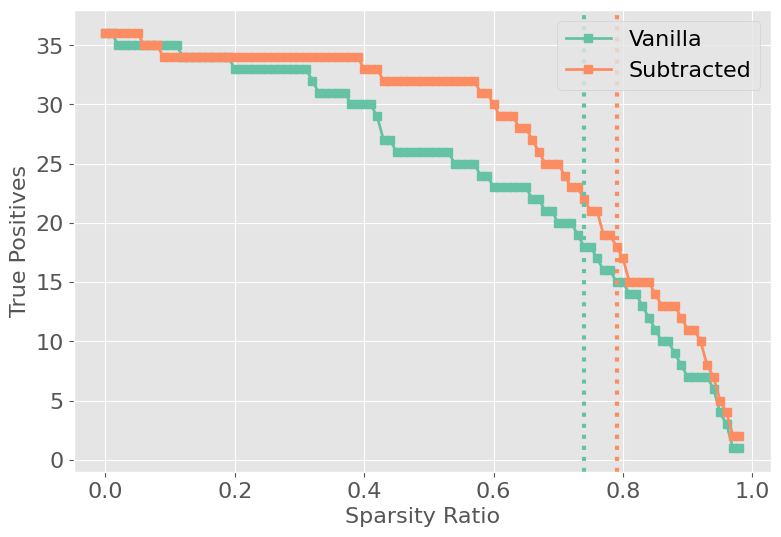}
            \caption{Docstring task}
        \end{subfigure} \qquad
        \caption{Amount of ground truth heads included by vanilla FLAP (\textbf{green}) and contrastive FLAP (\textbf{orange}). The dotted line shows the sparsity level at which the true positive reaches its \textbf{half-life value}.}
        \label{fig:half_life}
    \end{minipage}
    \par \medskip \medskip
    \begin{minipage}{\textwidth}
        \centering
        \resizebox{0.8\textwidth}{!}{%
            \begin{tabular}{l|ccccc}
                \toprule
                 & \textbf{IOI} & \textbf{GreaterThan} & \textbf{GenderedPronouns} & \textbf{Induction} & \textbf{Docstring} \\
                \midrule \midrule
                Vanilla FLAP    & \textbf{0.92} & 0.88 & 0.70 & 0.81 & 0.74 \\
                Contrastive FLAP& 0.90 & \textbf{0.92} & \textbf{0.81} & \textbf{0.87} & \textbf{0.79} \\
                \bottomrule
            \end{tabular}
        }
        \caption{Half-life values for Vanilla and Contrastive FLAP across all tasks}
        \label{tab:half_life}
    \end{minipage}
\end{figure*}
\noindent\textbf{Results} Except for the IOI task, the half-life value is reached only at higher sparsity levels for circuits identified by contrastive FLAP (see Table \ref{tab:half_life}). This indicates that 50\% of the ground truth heads are removed at higher sparsity levels. Figure \ref{fig:half_life} visualizes the exclusion of ground truth heads across all tested sparsity levels. contrastive FLAP maintains a plateau longer at lower sparsity compared to Vanilla FLAP. This effect is particularly pronounced for the Induction task, where the first exclusion of a ground truth head occurs at a sparsity level of 0.57, compared to 0.13 for Vanilla FLAP. A similar trend is observed for the IOI task. On the other hand, the two curves for the GreaterThan task are nearly identical, suggesting that some tasks benefit more

\subsubsection{Hybrid FLAP}
\paragraph{Methods}
Circuits of all three methods are evaluated over 200 samples for all five tasks and all three cliff points. See Appendix~\ref{sec:app_efficiency} to see which circuit was chosen for each task. 
Since the goal is to scale path patching to larger models, GPT-2 small, GPT-2 large, Qwen2.5-0.5B and Qwen2.5-7B are evaluated. See Appendix~\ref{sec:automatic_PP} to see results for the automatic path patching algorithms for all models. 
Circuits for Hybrid FLAP are chosen by evaluating all possible combinations of cliff points from contrastive and vanilla FLAP. The one with the best trade-off between performance and size is chosen. 
\paragraph{Results}
\begin{table*}[h!]
\centering
\resizebox{\textwidth}{!}{%
\begin{tabular}{ll|ccc|ccc|ccc|ccc|ccc}
\toprule
\textbf{Model} &  \textbf{FLAP Method}
& \multicolumn{3}{c|}{\textbf{IOI}} 
& \multicolumn{3}{c|}{\textbf{GreaterThan}} 
& \multicolumn{3}{c|}{\textbf{GenderedPronouns}} 
& \multicolumn{3}{c|}{\textbf{Induction}} 
& \multicolumn{3}{c}{\textbf{Docstring}} \\
& & Perf. (\%) & Size (Sparsity) & TPR & Perf. (\%) & Size (Sparsity) & TPR  & Perf. (\%) & Size (Sparsity) & TPR  & Perf. (\%) & Size (Sparsity) & TPR & Perf. (\%) & Size (Sparsity) & TPR \\
\midrule \midrule

\rowcolor{gray!20} GPT-2 small & Path Patching &  96.95 & 21 (0.85) & - & 79.65 & 8 (0.94) & - & 74.74 & 5 (0.97) & - & 93.35 & 22 (0.83) & - & 64.61 & 46 (0.68) & - \\
GPT-2 small & Clean      &  \textbf{93.51} & 49 (0.66) & 80.95 & 7.63 &  \textbf{16 (0.89)} & 37.50 & 33.41 & 35 (0.76) & 60.00 & 48.95 &  \textbf{9 (0.94)} & 13.64 & 36.33 &  \textbf{23 (0.84)} & 36.33 \\
GPT-2 small & Contrastive & 67.34  &  \textbf{22 (0.84)} & 76.19 & 69.38 & 31 (0.78) &  \textbf{87.50} & 32.44 &  \textbf{29 (0.80)} & 80.00 & 89.39 & 45 (0.69) &  \textbf{90.91} & 69.39 & 52 (0.64) & 71.73 \\
GPT-2 small & Merged    & 92.60 & 54 (0.63) & \textbf{90.48} &  \textbf{73.08} & 37 (0.74) &  \textbf{87.50} &  \textbf{69.36} & 48 (0.67) &  \textbf{100.00} & \textbf{88.69} & 47 (0.67) &  \textbf{90.91} &  \textbf{76.36} & 60 (0.58) &  \textbf{76.09} \\
\hline 

\rowcolor{gray!20} GPT-2 large & Path Patching &  92.60 & 193 (0.73) & - & 75.71 & 27 (0.96) & - & 73.94 & 73 (0.89) & - & 90.90 & 67 (0.91) & - & 73.19 & 122 (0.83) & - \\
GPT-2 large & Clean      &  69.97 & 186 (0.74) & 46.11 & \textbf{106.90} &  186 (0.74) & 74.07 & 68.15 & 244 (0.66) & 67.12 & 41.57 &  \textbf{107 (0.85)} & 35.82 & 75.27 &  201 (0.72) & 56.56 \\
GPT-2 large & Contrastive & 51.40 & \textbf{107 (0.85)} & 35.75 & 98.51 & \textbf{150 (0.79)} & 77.78 & \textbf{98.17} & \textbf{179 (0.75)} & 67.12 & 96.77 & 121 (0.83) & 64.18 & 45.18 & \textbf{71 (0.90)} & 27.87 \\
GPT-2 large & Merged     & \textbf{86.85} & 220 (0.69) & \textbf{53.37} &  105.28 & 237 (0.67) &  \textbf{85.19} &  96.20 & 310 (0.57) & \textbf{82.19} & \textbf{99.25} & 186 (0.74) & \textbf{71.16} & \textbf{84.05} & 221 (0.69) & \textbf{61.48} \\
\hline 

\rowcolor{gray!20} Qwen2.5-0.5B & Path Patching &  76.33 & 62 (0.82) & - & 75.14 & 20 (0.94) & - & 86.84 & 25 (0.93) & - & 76.08 & 18 (0.95) & - & 36.04 & 34 (0.90) & - \\
Qwen2.5-0.5B & Clean      &  17.72 & \textbf{19 (0.94)} & 20.97 & 34.45 &  \textbf{53 (0.84)} & 60.00 & 62.24 & 70 (0.79) & 56.00 & 58.58 &  \textbf{83 (0.75)} & 55.56 & 35.09 &  \textbf{19 (0.94)} & 19.05 \\
Qwen2.5-0.5B & Contrastive & \textbf{82.84}  &  113 (0.66) & \textbf{67.74} & 87.01 & 130 (0.61) & \textbf{95.00} & 55.27 & \textbf{59 (0.82)} & 40.00 & 93.12 & 127 (0.62) & \textbf{100.00} & \textbf{72.13} & 133 (0.60) & \textbf{85.71} \\
Qwen2.5-0.5B & Merged    &  82.54 & 114 (0.66) & \textbf{67.74} &  \textbf{87.53} & 138 (0.59) &  \textbf{95.00} &  \textbf{69.90} & 89 (0.74) & \textbf{69.90} & \textbf{94.86} & 149 (0.56) & \textbf{100.00} & \textbf{72.13} & 133 (0.60) & \textbf{85.71} \\
\hline 

\rowcolor{gray!20} Qwen2.5-7B & Path Patching &  73.25 & 221 (0.72) & - & 79.33 & 34 (0.95) & - & 68.69 & 125 (0.84) & - & 67.90 & 10 (0.99) & - & 64.79 & 150 (0.81) & - \\
Qwen2.5-7B & Clean      & 9.38  & \textbf{195 (0.75)} & 41.17 & 30.29 & 195 (0.75) & 52.94 & 2.02 & \textbf{195 (0.75)} & 41.60 & 77.46 & 148 (0.81) & 50.00 & 36.45 & \textbf{195 (0.75)} & 47.33 \\
Qwen2.5-7B & Contrastive & 71.23  & 219 (0.72) & 55.20 & 84.37 & \textbf{132 (0.83)} & 47.06 & 21.27 & 250 (0.68) & 56.00 & 61.61 & \textbf{101 (0.87)} & 50.00  & 70.58 & 258 (0.67) & 70.66 \\
Qwen2.5-7B & Merged     & \textbf{82.09} & 303 (0.61) &\textbf{74.38} & \textbf{88.03} & 232 (0.70) & \textbf{67.65} & \textbf{70.92} & 317 (0.59) & \textbf{61.60} & \textbf{93.03} & 206 (0.74) & \textbf{60.00} & \textbf{73.57} & 336 (0.57) & \textbf{79.33} \\
\bottomrule
\end{tabular}%
}
\caption{Comparing the performance, size and true positive ratio of Path Patching to pruning with vanilla FLAP, Contrastive FLAP and the resulting circuit of merging vanilla and Contrastive FLAP.}
\label{tab:exp4}
\end{table*}

Hybrid FLAP yields in circuits with an average TPR of 82.39\% for GPT-2 small, 68.12\% for GPT-2 large, 80.07\% for Qwen2.5-0.5B and 73.73\% for Qwen2.5-7B (see Table \ref{tab:exp4}). With a minimum sparsity ratio of 0.58 for the Docstring task on GPT-2 small, 0.64 for the GreaterThan task for GPT-2 large, 0.56 for the Induction task on Qwen2.5-0.5B and 0.57 for the Docstring task on Qwen2.5-7B.
Altough Hybrid FLAP circuits are larger than the ones discovered by Vanilla or contrastive FLAP, the TPR is higher and performance are generally higher.

\section{Accelerated Path Patching: Efficiency}
\label{sec:app_efficiency}
\begin{figure*}[t]
    \centering
    \begin{subfigure}{0.45\textwidth}
        \centering
        \includegraphics[width=\linewidth]{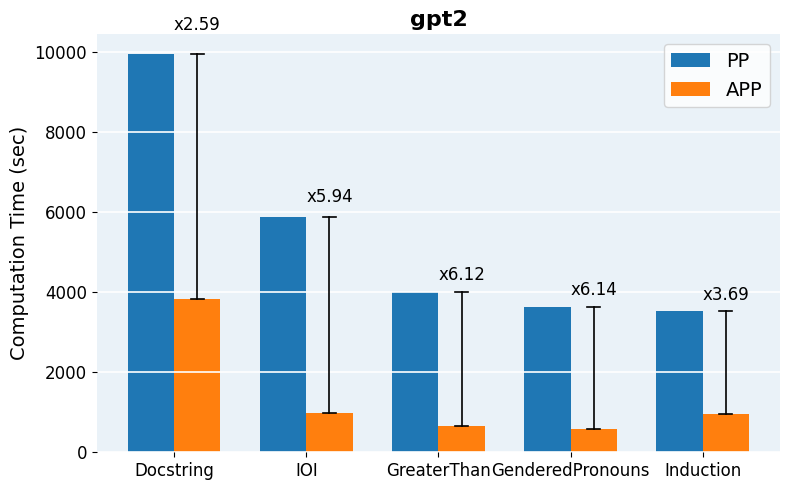}
    \end{subfigure}
    \hfill
    \begin{subfigure}{0.45\textwidth}
        \centering
        \includegraphics[width=\linewidth]{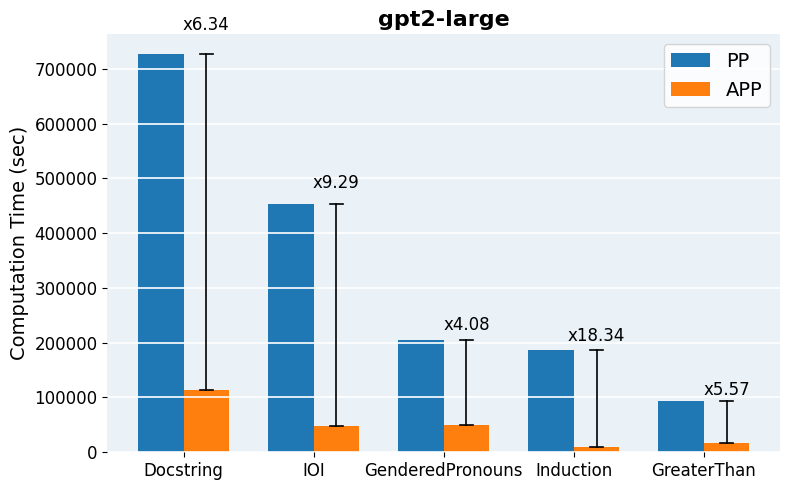}
    \end{subfigure} \\
    \par\medskip
    \begin{subfigure}{0.45\textwidth}
        \centering
        \includegraphics[width=\linewidth]{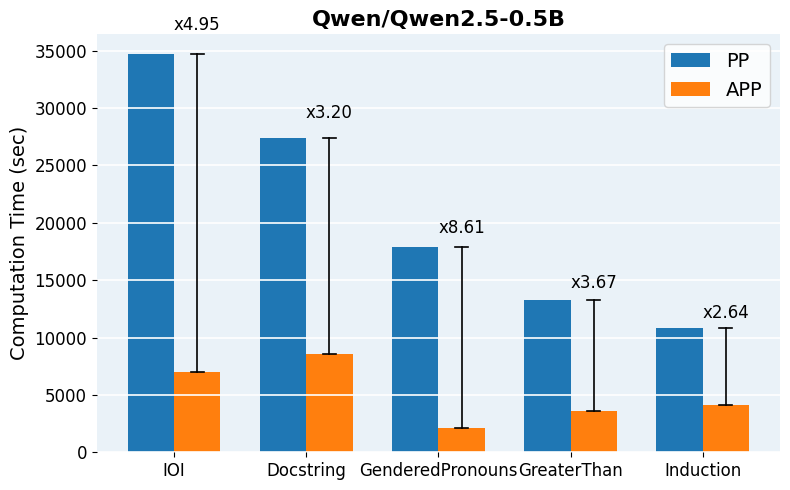}
    \end{subfigure} 
    \hfill
    \begin{subfigure}{0.45\textwidth}
        \centering
        \includegraphics[width=\linewidth]{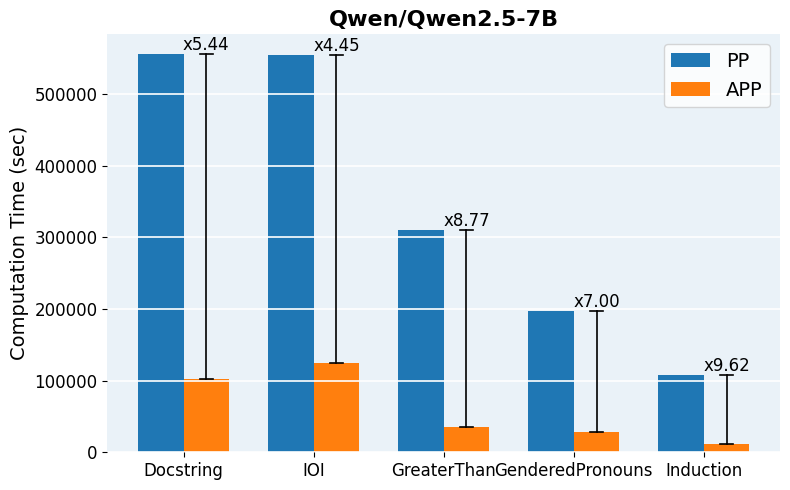}
    \end{subfigure} \\
    \caption{\centering Difference of required computation time across all models and tasks}
    \label{fig:time_efficency}
\end{figure*}

The efficiency gains in terms of computation time paint a similar pattern to the GFLOPs. APP yields a significant speed-up compared to PP (see Figure \ref{fig:time_efficency}).

\section{Hyperparameters Tuning}

\subsection{Automatic Path Patching}

\begin{figure*}[t]
\scalebox{0.85}{
    \centering
    \begin{minipage}{1\textwidth}
        \centering
        \textbf{Automatic Path Patching: Hyperparameter testing for GPT-2 small}
    
        \begin{subfigure}{0.45\textwidth}
                \centering
                \includegraphics[width=\linewidth]{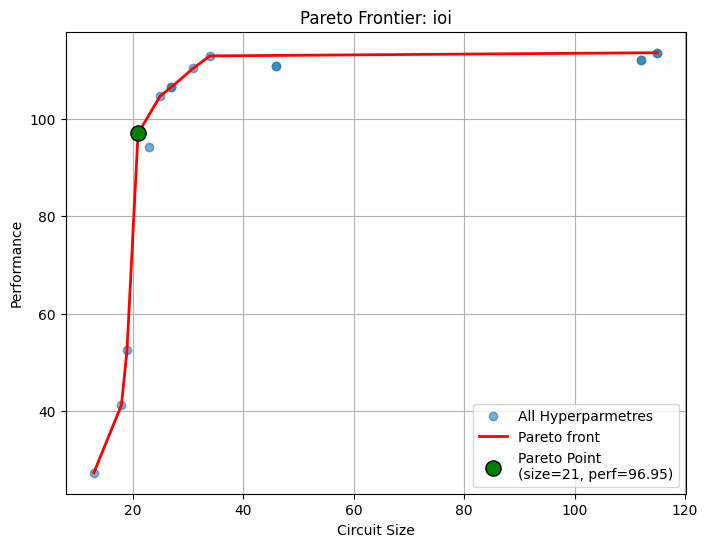}
                \caption{IOI task}
        \end{subfigure} \qquad
        \begin{subfigure}{0.45\textwidth}
                \centering
                \includegraphics[width=\linewidth]{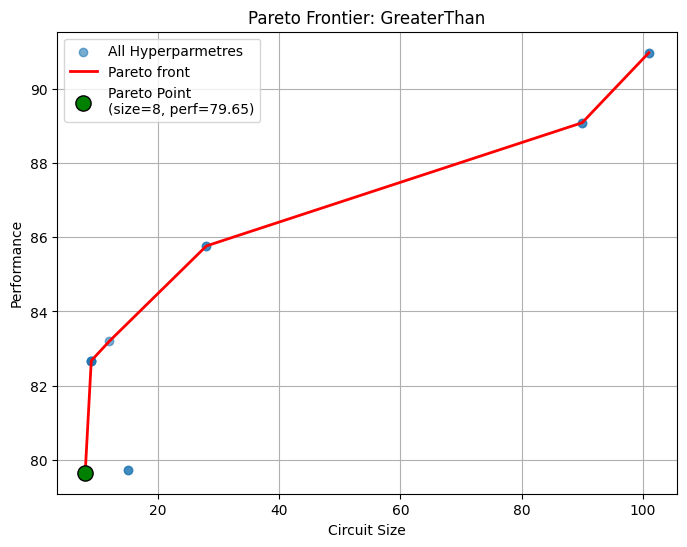}
                \caption{GreaterThan task}
        \end{subfigure}   \\
            \par \medskip
        \begin{subfigure}{0.45\textwidth}
                \centering
                \includegraphics[width=\linewidth]{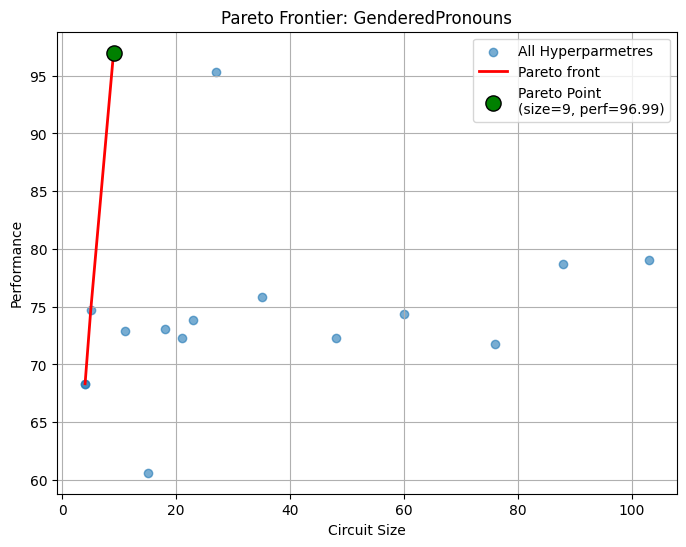}
                \caption{GenderedPronouns task}
        \end{subfigure} \qquad
        \begin{subfigure}{0.45\textwidth}
                \centering
                \includegraphics[width=\linewidth]{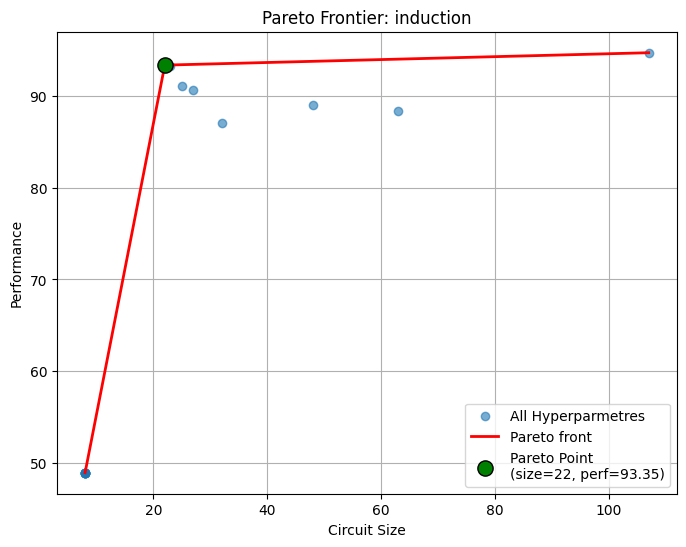}
                \caption{Induction task}
        \end{subfigure}  \\
            \par \medskip
        \begin{subfigure}{0.45\textwidth}
            \centering
            \includegraphics[width=\linewidth]{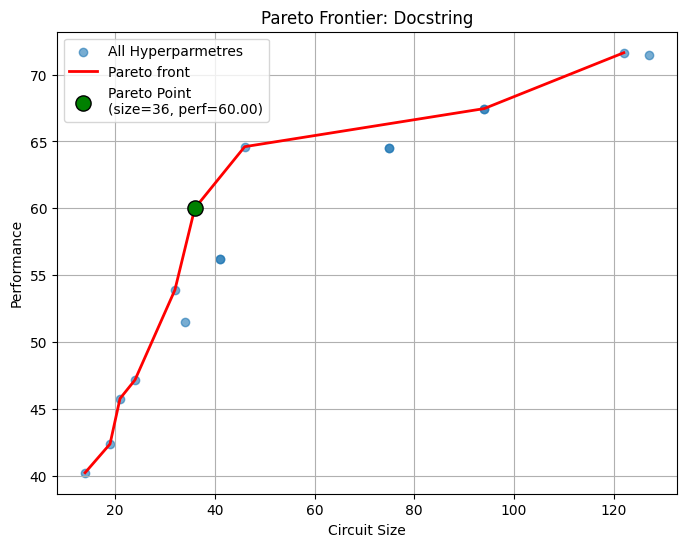}
                \caption{Docstring task}
        \end{subfigure} 
                \\
        \par \medskip
        \par \medskip
        \small
        \centering
        \resizebox{\textwidth}{!}{%
        \begin{tabular}{l|l|cc|cc|cc|cc|cc}
        \toprule
        \multirow{2}{*}{Maximum Value} & \multirow{2}{*}{Importance}
        & \multicolumn{2}{c|}{\textbf{IOI}}
        & \multicolumn{2}{c|}{\textbf{GreaterThan}}
        & \multicolumn{2}{c|}{\textbf{GenderedPronouns}}
        & \multicolumn{2}{c|}{\textbf{Induction}}
        & \multicolumn{2}{c}{\textbf{Docstring}} \\
        
        & & \textbf{P} & size
        & \textbf{P} & size
        & \textbf{P} & size
        & \textbf{P} & size
        & \textbf{P} & size \\
        \midrule \midrule
        \multirow{4}{*}{0.01}
        & 1   & 112.86\% & 34  & 83.20\%  & 12  & 95.35\%  & 27  & 48.95\%  & 8  & 64.61\%  & 46  \\
        & 1.5 & 110.32\%  & 31  & 82.67\%  & 9   & 73.11\%  & 18  & 48.95\%  & 8  & \textbf{60.00\%}  & \textbf{36}  \\
        & 2   & 104.61\%  & 25  & 82.67\%  & 9   & 60.63\%  & 15  & 48.95\%  & 8  & 53.88\%  & 32  \\
        & 2.5 & \textbf{96.95\%}   & \textbf{21}  & \textbf{79.65\%}  & \textbf{8}   & 72.89\%  & 11  & 48.95\%  & 8  & 45.76\%  & 21  \\
        \hline
        \multirow{4}{*}{0.001}
        & 1   & 113.51\%  & 115 & 90.98\%  & 101 & 79.04\%  & 103 & 94.70\%  & 107 & 71.45\%  & 127 \\
        & 1.5 & 112.01\%  & 112 & 89.09\%  & 90  & 71.77\%  & 76  & 88.39\%  & 63  & 67.45\%  & 94  \\
        & 2   & 110.84\%  & 46  & 85.77\%  & 28  & 72.29\%  & 48  & 90.60\%  & 27  & 64.48\%  & 75  \\
        & 2.5 & 106.4\% 3 & 27  & 79.73\%  & 15  & 73.88\%  & 23  & 93.25\%  & 23  & 56.21\%  & 41  \\
        \hline
        \multirow{4}{*}{0.02}
        & 1   & 94.16\%   & 23  & 82.67\%  & 9   & 96.99\%  & 9  & 48.95\%  & 8  & 51.45\%  & 34  \\
        & 1.5 & 52.50\%   & 19  & 82.67\%  & 9   & \textbf{74.74\%}  & \textbf{5}   & 48.95\%  & 8  & 47.15\%  & 24  \\
        & 2   & 41.23\%   & 18  & \textbf{79.65\%}  & \textbf{8}   & 68.33\%  & 4   & 48.95\%  & 8  & 42.36\%  & 19  \\
        & 2.5 & 27.37\%   & 13  & \textbf{79.65\%}  & \textbf{8}   & 68.33\%  & 4   & 48.95\%  & 8  & 40.21\%  & 14  \\
        \hline
        \multirow{4}{*}{0.002}
        & 1   & 113.51\%  & 115 & 90.98\%  & 101 & 78.70\%  & 88  & 88.99\%  & 48  & 71.64\%  & 122 \\
        & 1.5 & 112.01\%  & 112 & 89.09\%  & 90  & 74.35\%  & 60  & 87.09\%  & 32  & 67.45\%  & 94  \\
        & 2   & 110.84\%  & 46  & 85.77\%  & 28  & 75.86\%  & 35  & 91.10\%  & 25  & 64.48\%  & 75  \\
        & 2.5 & 106.43\%  & 27  & 79.73\%  & 15  & 72.29\%  & 21  & \textbf{93.35\%}  & \textbf{22}  & 56.21\%  & 41  \\
        \bottomrule
        \end{tabular}
        }
        \caption{PP on GPT-2 small: Pareto points for each task are marked in bold.}
        
        \label{tab:Pareto_PP_GPT2-small}
    \end{minipage}
    }
\end{figure*}

\begin{figure*}[t]
\scalebox{0.85}{
    \centering
    \begin{minipage}{1\textwidth}
        \centering
        \textbf{Automatic Path Patching: Hyperparameter testing for GPT-2 large}
    
        \begin{subfigure}{0.45\textwidth}
                \centering
                \includegraphics[width=\linewidth]{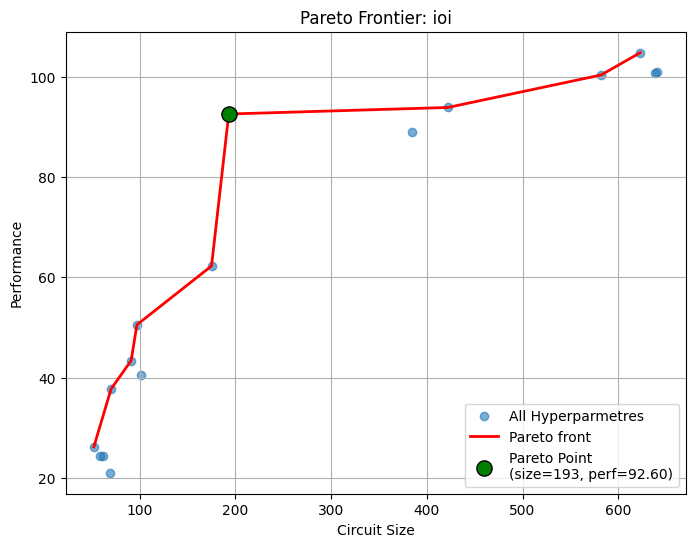}
                \caption{IOI task}
        \end{subfigure} \qquad
        \begin{subfigure}{0.45\textwidth}
                \centering
                \includegraphics[width=\linewidth]{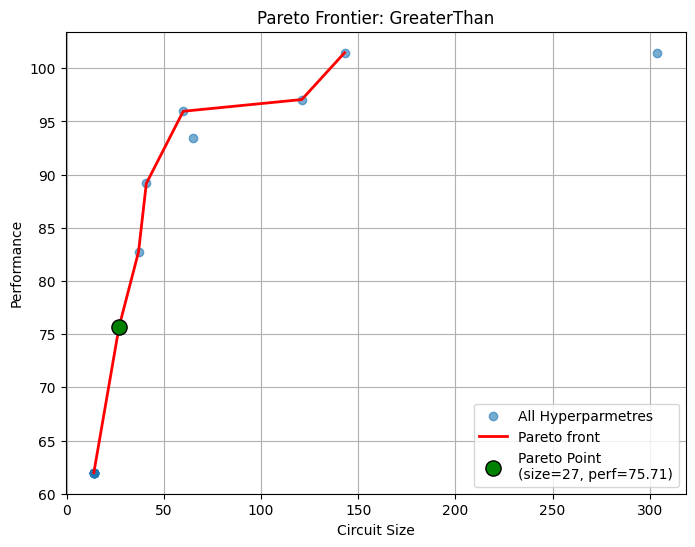}
                \caption{GreaterThan task}
        \end{subfigure}   \\
            \par \medskip
        \begin{subfigure}{0.45\textwidth}
                \centering
                \includegraphics[width=\linewidth]{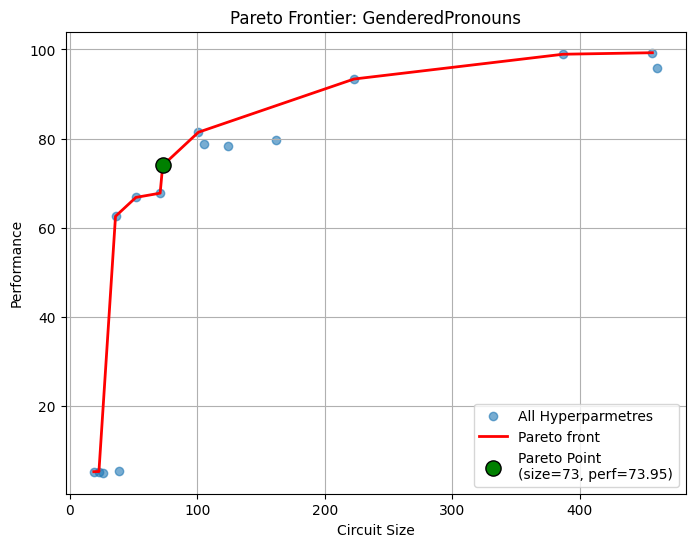}
                \caption{GenderedPronouns task}
        \end{subfigure} \qquad
        \begin{subfigure}{0.45\textwidth}
                \centering
                \includegraphics[width=\linewidth]{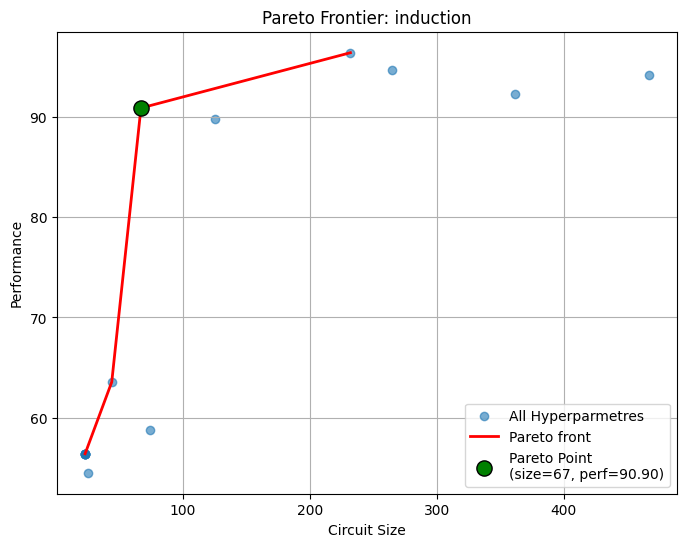}
                \caption{Induction task}
        \end{subfigure}  \\
            \par \medskip
        \begin{subfigure}{0.45\textwidth}
            \centering
            \includegraphics[width=\linewidth]{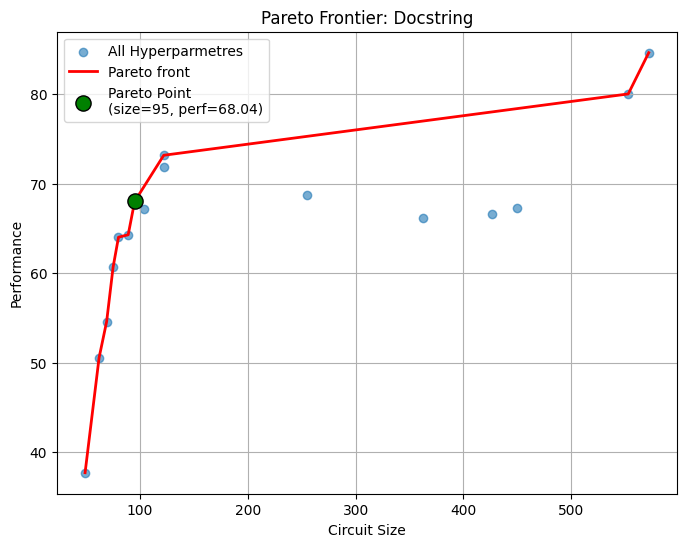}
                \caption{Docstring task}
        \end{subfigure}  
        \\
        \par \medskip
        \par \medskip

        \small
        \resizebox{\textwidth}{!}{%
        \begin{tabular}{l|l|cc|cc|cc|cc|cc}
        \toprule
        \multirow{2}{*}{Maximum Value} & \multirow{2}{*}{Importance}
        & \multicolumn{2}{c|}{\textbf{IOI}}
        & \multicolumn{2}{c|}{\textbf{GreaterThan}}
        & \multicolumn{2}{c|}{\textbf{GenderedPronouns}}
        & \multicolumn{2}{c|}{\textbf{Induction}}
        & \multicolumn{2}{c}{\textbf{Docstring}} \\
        
        & & \textbf{P} & size
        & \textbf{P} & size
        & \textbf{P} & size
        & \textbf{P} & size
        & \textbf{P} & size \\
        \midrule \midrule
        \multirow{4}{*}{0.01}
        & 1   & 62.28\% & 175 & 61.97\%  & 14  & 81.41\%  & 101 & 56.39\%  & 23  & 73.19\%  & 122 \\
        & 1.5 & 43.40\%  & 91  & 61.97\%  & 14  & 67.74\%  & 71  & 56.39\%  & 23  & \textbf{68.04\%}  & \textbf{95}  \\
        & 2   & 21.01\%  & 69  & 61.97\%  & 14  & 66.79\%  & 52  & 56.39\%  & 23  & 64.04\%  & 80  \\
        & 2.5 & 24.43\%  & 62  & 61.97\%  & 14  & 62.55\%  & 36  & 56.39\%  & 23  & 60.69\%  & 75  \\
        \hline
        \multirow{4}{*}{0.001}
        & 1   & 101.06\%  & 641 & 101.45\%  & 304 & 95.79\%  & 461 & 94.18\%  & 467 & 67.32\%  & 450 \\
        & 1.5 & 104.84\%  & 623 & 101.47\%  & 143 & 79.58\%  & 162 & 92.34\%  & 362 & 66.61\%  & 427 \\
        & 2   & 100.39\%  & 582 & 95.96\%   & 60  & 78.29\%  & 124 & 94.68\%  & 265 & 68.75\%  & 255 \\
        & 2.5 & 93.91\%   & 422 & 89.17\%   & 41  & \textbf{73.95\%}  & \textbf{73}  & 58.78\%  & 74  & 67.21\%  & 104 \\
        \hline
        \multirow{4}{*}{0.02}
        & 1   & 50.53\%  & 97  & 61.97\%  & 14  & 5.33\%   & 39  & 56.39\%  & 23  & 64.31\%  & 89  \\
        & 1.5 & 37.72\%  & 70  & 61.97\%  & 14  & 5.01\%   & 26  & 56.39\%  & 23  & 54.61\%  & 69  \\
        & 2   & 24.36\%  & 58  & 61.97\%  & 14  & 5.28\%   & 23  & 56.39\%  & 23  & 50.58\%  & 62  \\
        & 2.5 & 26.16\%  & 52  & 61.97\%  & 14  & 5.24\%   & 19  & 54.50\%  & 25  & 37.72\%  & 49  \\
        \hline
        \multirow{4}{*}{0.002}
        & 1   & 100.87\%  & 638 & 97.06\%  & 121 & 99.25\%  & 457 & 96.42\%  & 232 & 80.04\%  & 553 \\
        & 1.5 & \textbf{92.60\%}   & \textbf{193} & 93.45\%  & 65  & 98.91\%  & 387 & 89.81\%  & 125 & 66.23\%  & 363 \\
        & 2   & 89.02\%   & 385 & 82.74\%  & 37  & 93.35\%  & 223 & \textbf{90.90\%}  & \textbf{67}  & 71.93\%  & 122 \\
        & 2.5 & 40.59\%   & 101 & \textbf{75.71\%}  & \textbf{27}  & 78.70\%  & 105 & 63.55\%  & 44  & 84.65\%  & 572 \\
        \bottomrule
        \end{tabular}
        }
        \caption{PP om GPT-2 large: Pareto points for each task are marked in bold.}
        \label{tab:Pareto_PP_GPT2-large}
    \end{minipage}
    }
\end{figure*}

\begin{figure*}[t]
\scalebox{0.85}{
    \centering
    \begin{minipage}{1\textwidth}
        \centering
        \textbf{Automatic Path Patching: Hyperparameter testing for Qwen2.5-0.5B}
    
        \begin{subfigure}{0.45\textwidth}
                \centering
                \includegraphics[width=\linewidth]{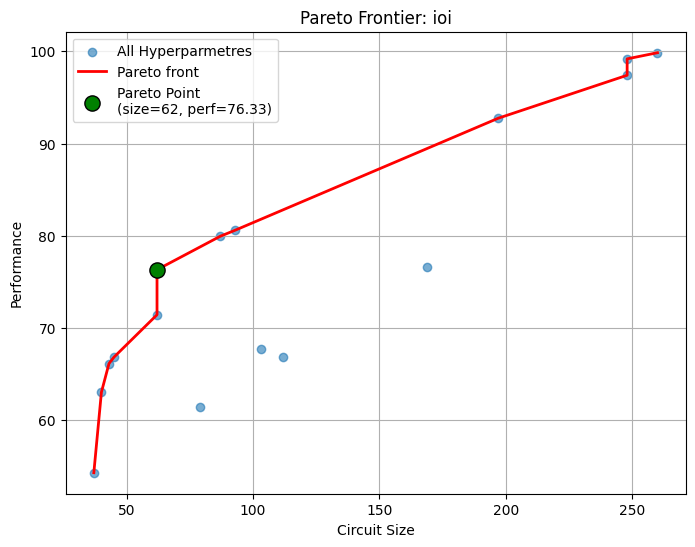}
                \caption{IOI task}
        \end{subfigure} \qquad
        \begin{subfigure}{0.45\textwidth}
                \centering
                \includegraphics[width=\linewidth]{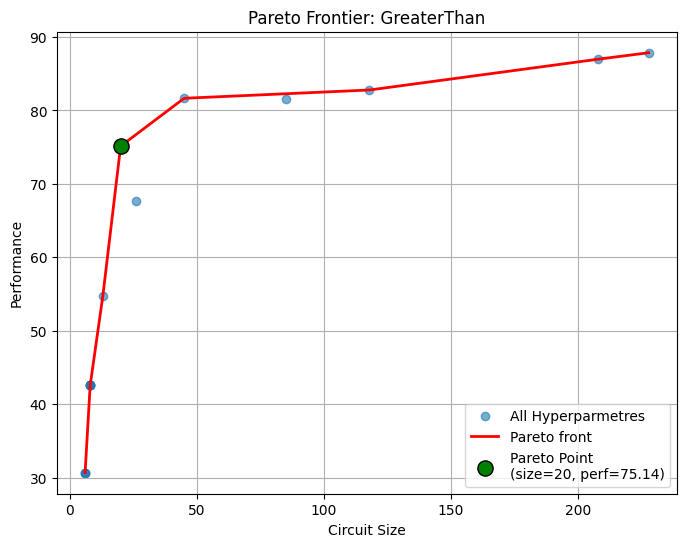}
                \caption{GreaterThan task}
        \end{subfigure}   \\
            \par \medskip
        \begin{subfigure}{0.45\textwidth}
                \centering
                \includegraphics[width=\linewidth]{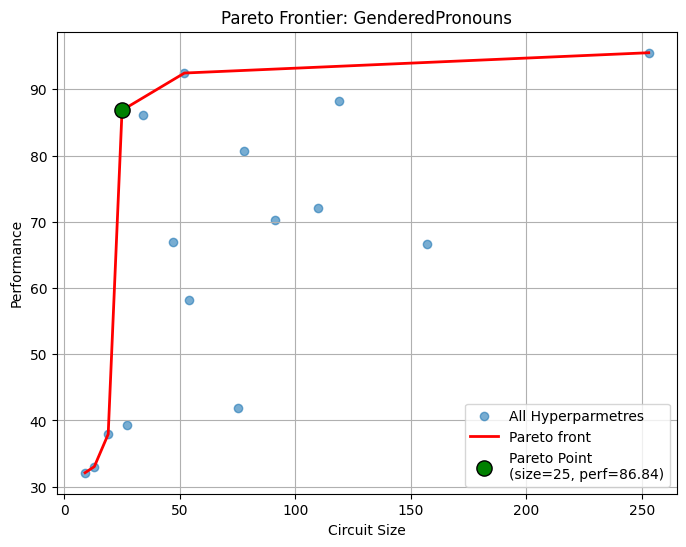}
                \caption{GenderedPronouns task}
        \end{subfigure} \qquad
        \begin{subfigure}{0.45\textwidth}
                \centering
                \includegraphics[width=\linewidth]{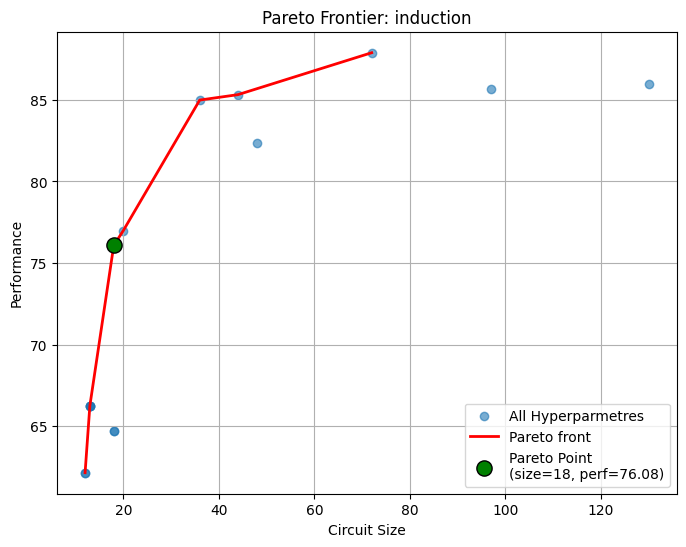}
                \caption{Induction task}
        \end{subfigure}  \\
            \par \medskip
        \begin{subfigure}{0.45\textwidth}
            \centering
            \includegraphics[width=\linewidth]{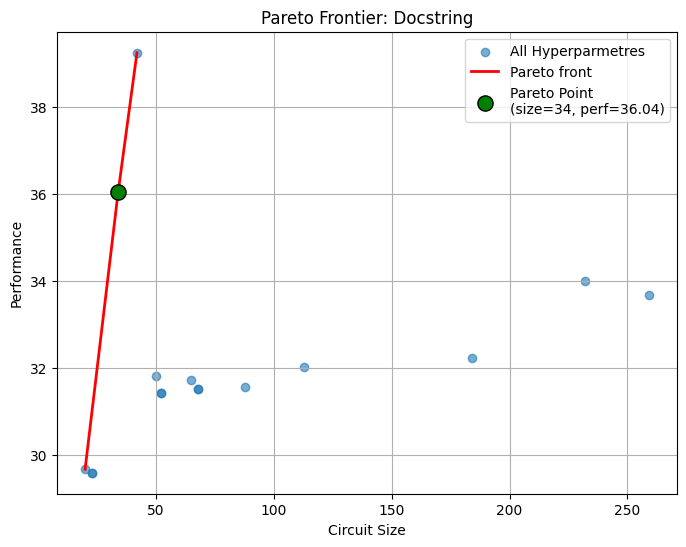}
                \caption{Docstring task}
        \end{subfigure} 
        \\
        \par \medskip
        \par \medskip
        \small
        \resizebox{\textwidth}{!}{%
        \begin{tabular}{l|l|cc|cc|cc|cc|cc}
        \toprule
        \multirow{2}{*}{Maximum Value} & \multirow{2}{*}{Importance}
        & \multicolumn{2}{c|}{\textbf{IOI}}
        & \multicolumn{2}{c|}{\textbf{GreaterThan}}
        & \multicolumn{2}{c|}{\textbf{GenderedPronouns}}
        & \multicolumn{2}{c|}{\textbf{Induction}}
        & \multicolumn{2}{c}{\textbf{Docstring}} \\
        
        & & \textbf{P} & size
        & \textbf{P} & size
        & \textbf{P} & size
        & \textbf{P} & size
        & \textbf{P} & size \\
        \midrule \midrule
        \multirow{4}{*}{0.01}
        & 1   & 80.62\% & 93  & 42.55\% & 8   & 70.29\% & 91  & 66.25\% & 13  & 31.57\% & 88  \\
        & 1.5 & \textbf{76.33\%} & \textbf{62}  & 42.55\% & 8   & 92.46\% & 52  & 66.25\% & 13  & 31.72\% & 65  \\
        & 2   & 66.86\% & 45  & 42.55\% & 8   & 86.20\% & 34  & 66.25\% & 13  & 31.83\% & 50  \\
        & 2.5 & 63.05\% & 40  & 42.55\% & 8   & \textbf{86.84\%} & \textbf{25}  & 66.25\% & 13  & 39.25\% & 42  \\
        \hline
        \multirow{4}{*}{0.001}
        & 1   & 99.85\% & 260 & 87.86\% & 228 & 95.53\% & 253 & 85.99\% & 130 & 33.68\% & 259 \\
        & 1.5 & 97.41\% & 248 & 86.98\% & 208 & 88.18\% & 119 & 87.89\% & 72  & 32.24\% & 184 \\
        & 2   & 76.63\% & 169 & 81.60\% & 85  & 80.64\% & 78  & 82.38\% & 48  & 31.52\% & 68  \\
        & 2.5 & 66.86\% & 112 & 67.66\% & 26  & 66.96\% & 47  & 84.99\% & 36  & 31.42\% & 52  \\
        \hline
        \multirow{4}{*}{0.02}
        & 1   & 79.96\% & 87  &  42.55\% & 8   & 39.27\% & 27  & 62.15\% & 12  & \textbf{36.04\%} & \textbf{34}  \\
        & 1.5 & \textbf{76.33\%} & \textbf{62}  & 42.55\% & 8   & 37.92\% & 19  & 62.15\% & 12  & 29.59\% & 23  \\
        & 2   & 66.16\% & 43  &  42.55\% & 8   & 33.04\% & 13  &  62.15\% & 12  & 29.59\% & 23  \\
        & 2.5 & 54.33\% & 37  & 42.55\% & 8   & 32.11\% & 9   & 62.15\% & 12  & 29.67\% & 20  \\
        \hline
        \multirow{4}{*}{0.002}
        & 1   & 99.19\% & 248 & 82.79\% & 118 & 66.71\% & 157 & 85.66\% & 97  & 34.01\% & 232 \\
        & 1.5 & 92.75\% & 197 & 81.66\% & 45  & 72.08\% & 110 & 85.32\% & 44  & 32.03\% & 113 \\
        & 2   & 67.79\% & 103 & \textbf{75.14\%} & \textbf{20}  & 41.85\% & 75  & \textbf{76.08\%} & \textbf{18}  & 31.52\% & 68  \\
        & 2.5 & 61.43\% & 79  & 54.76\% & 13  & 58.21\% & 54  & 76.99\% & 20  & 31.42\% & 52  \\
        \bottomrule
        \end{tabular}
        }
        
        \caption{PP on Qwen2.5-0.5B Pareto points for each task are marked in bold.}
        \label{tab:Pareto_PP_Qwen2.5-0.5}
    \end{minipage}
    }
\end{figure*}

\begin{figure*}[t]
\scalebox{0.85}{
    \centering
    \begin{minipage}{1\textwidth}
        \centering
        \textbf{Automatic Path Patching: Hyperparameter testing for Qwen2.5-7B}

        \begin{subfigure}{0.45\textwidth}
            \centering
            \includegraphics[width=\linewidth]{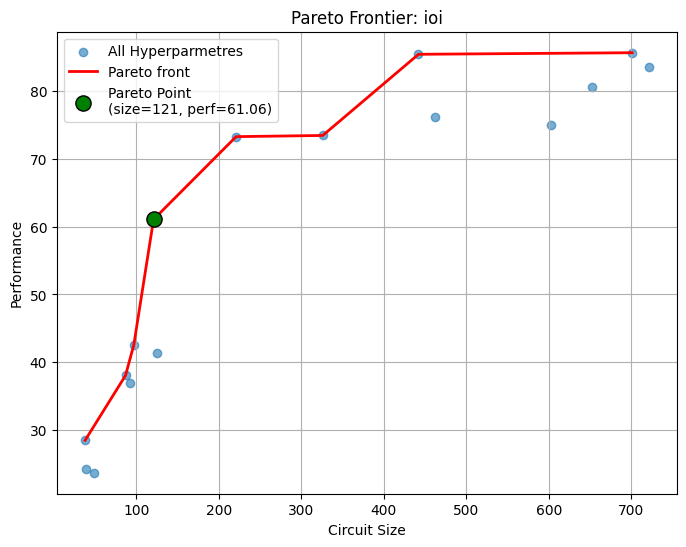}
            \caption{IOI task}
        \end{subfigure} \qquad
        \begin{subfigure}{0.45\textwidth}
            \centering
            \includegraphics[width=\linewidth]{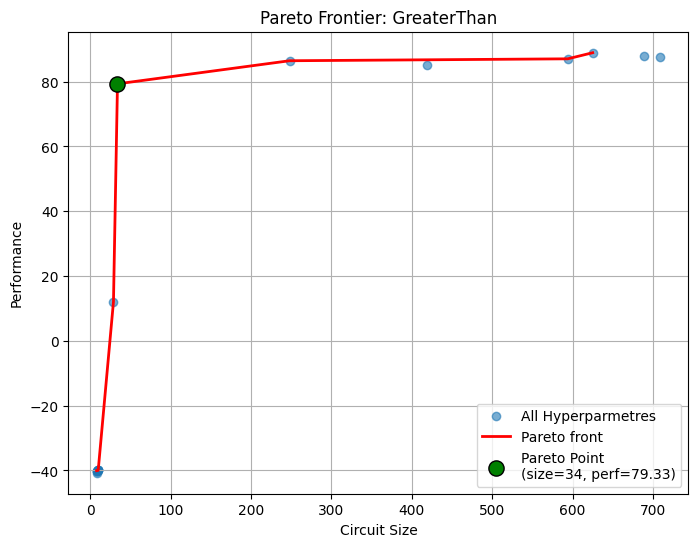}
            \caption{GreaterThan task}
        \end{subfigure} \\
        \par \medskip
        \begin{subfigure}{0.45\textwidth}
            \centering
            \includegraphics[width=\linewidth]{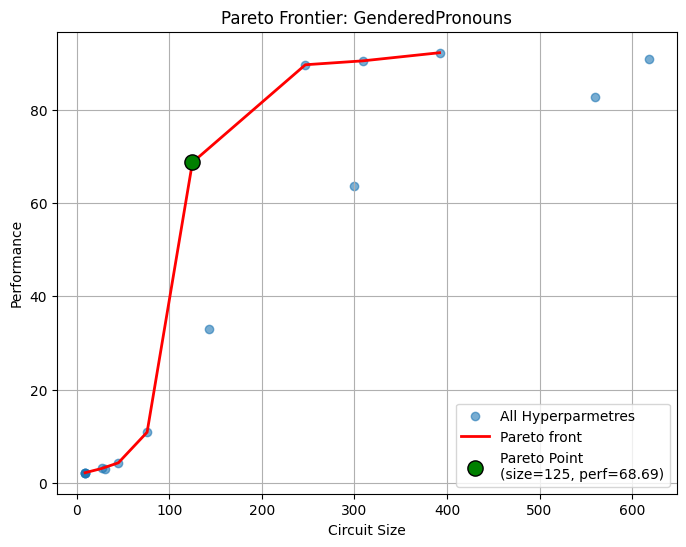}
            \caption{GenderedPronouns task}
        \end{subfigure} \qquad
        \begin{subfigure}{0.45\textwidth}
            \centering
            \includegraphics[width=\linewidth]{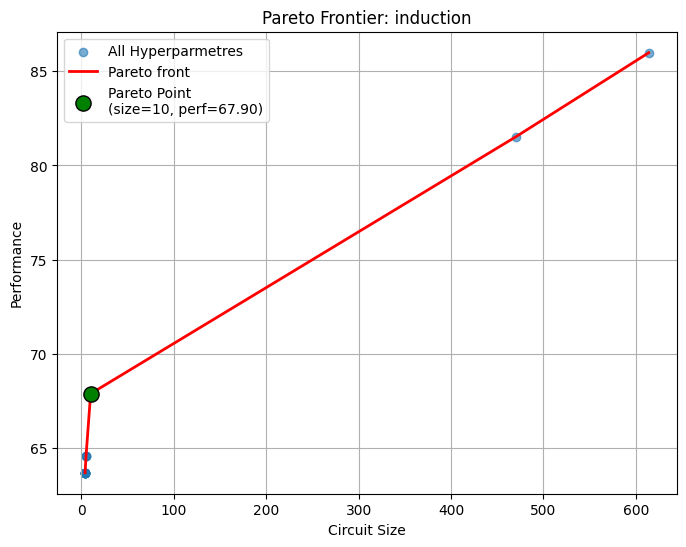}
            \caption{Induction task}
        \end{subfigure} \\
        \par \medskip
        \begin{subfigure}{0.45\textwidth}
            \centering
            \includegraphics[width=\linewidth]{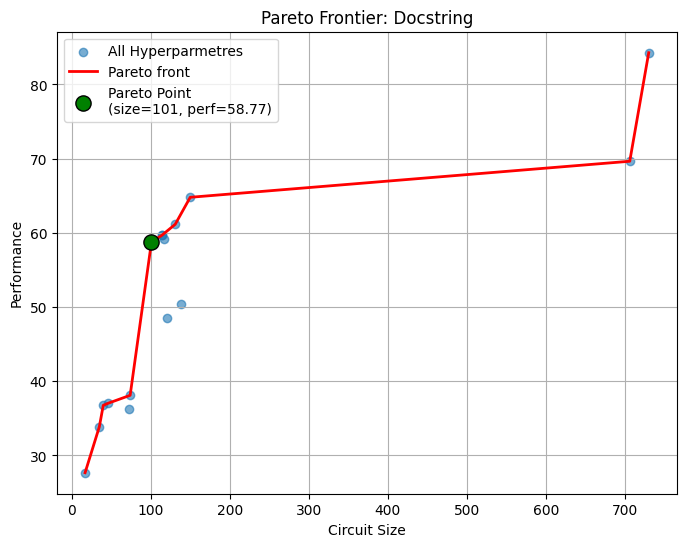}
            \caption{Docstring task}
        \end{subfigure} \\
        \par \medskip

        \small 
        \resizebox{\textwidth}{!}{%
        \begin{tabular}{l|l|cc|cc|cc|cc|cc}
        \toprule
        Maximum Value & Importance
        & \multicolumn{2}{c|}{\textbf{IOI}}
        & \multicolumn{2}{c|}{\textbf{GreaterThan}}
        & \multicolumn{2}{c|}{\textbf{GenderedPronouns}}
        & \multicolumn{2}{c|}{\textbf{Induction}}
        & \multicolumn{2}{c}{\textbf{Docstring}} \\
        & & \textbf{P} & size
        & \textbf{P} & size
        & \textbf{P} & size
        & \textbf{P} & size
        & \textbf{P} & size \\
        \midrule \midrule

        \multirow{4}{*}{0.01}
        & 1   & 73.25\% & 221 & -39.98\% & 10  & 10.89\% & 76  & 63.71\% & 4   & 59.11\% & 117 \\
        & 1.5 & 41.39\% & 125 & -39.98\% & 10  &  4.34\% & 45  & 63.71\% & 4   & \textbf{58.77\%} & \textbf{101} \\
        & 2   & 42.55\% & 97  & -39.98\% & 10  &  2.95\% & 31  & 63.71\% & 4   & 38.11\% & 74  \\
        & 2.5 & 38.07\% & 87  & -39.98\% & 10  &  3.12\% & 27  & 63.71\% & 4   & 36.26\% & 72  \\
        \hline

        \multirow{4}{*}{0.001}
        & 1   & 83.48\% & 721 & 87.72\% & 709 & 92.19\% & 392 & 85.97\% & 614 & 84.27\% & 730 \\
        & 1.5 & 85.62\% & 701 & 87.88\% & 689 & 90.45\% & 309 & 81.51\% & 470 & 64.79\% & 150 \\
        & 2   & 80.53\% & 652 & 87.05\% & 594 & 89.61\% & 247 & \textbf{67.90\%} & \textbf{10}  & 61.15\% & 131 \\
        & 2.5 & 74.99\% & 602 & 86.45\% & 248 & \textbf{68.69\%} & \textbf{125} & \textbf{67.90\%} & \textbf{10}  & 59.66\% & 114 \\
        \hline

        \multirow{4}{*}{0.02}
        & 1   & 37.03\% & 92  & -40.06\% & 8   &  2.21\% & 9   & 63.71\% & 4   & 37.03\% & 46  \\
        & 1.5 & 23.74\% & 49  & -40.06\% & 8   &  2.21\% & 9   & 63.71\% & 4   & 36.75\% & 40  \\
        & 2   & 24.23\% & 39  & -40.06\% & 8   &  2.21\% & 9   & 63.71\% & 4   & 27.68\% & 17  \\
        & 2.5 & 28.51\% & 38  & -40.69\% & 9   &  2.21\% & 9   & 64.57\% & 5   & 33.86\% & 35  \\
        \hline

        \multirow{4}{*}{0.002}
        & 1   & 76.08\% & 462 & 88.93\% & 625 & 90.75\% & 618 & 64.57\% & 5   & 69.65\% & 706 \\
        & 1.5 & 73.43\% & 326 & 85.18\% & 419 & 82.64\% & 560 & 63.71\% & 4   & 50.44\% & 138 \\
        & 2   & 85.38\% & 442 & \textbf{79.33\%} & \textbf{34}  & 63.61\% & 300 & 63.71\% & 4   & 48.57\% & 121 \\
        & 2.5 & \textbf{61.06\%} & \textbf{121} & 12.08\% & 29  & 33.00\% & 143 & 64.57\% & 5   & 59.66\% & 114 \\
        \bottomrule
        \end{tabular}%
        }
        \caption{PP on Qwen2.5-7B: Pareto points for each task are marked in bold.}
        \label{tab:Pareto_PP_Qwen2.5-7B-medium}
    \end{minipage}
}
\end{figure*}

In Figure~\ref{tab:Pareto_PP_GPT2-small}, \ref{tab:Pareto_PP_GPT2-large}, \ref{tab:Pareto_PP_Qwen2.5-0.5}, and \ref{tab:Pareto_PP_Qwen2.5-7B-medium}, we show the resulting circuits from Automatic PP and model performances across tasks and model families under different hyperparameter settings.

\subsection{Accelerated Path Patching}
\begin{figure*}[t]
\scalebox{0.85}{
    \centering
    \begin{minipage}{1\textwidth}
        \centering
        \textbf{Accelerated Path Patching: Hyperparameter testing for GPT-2 small}

        \begin{subfigure}{0.45\textwidth}
            \centering
            \includegraphics[width=\linewidth]{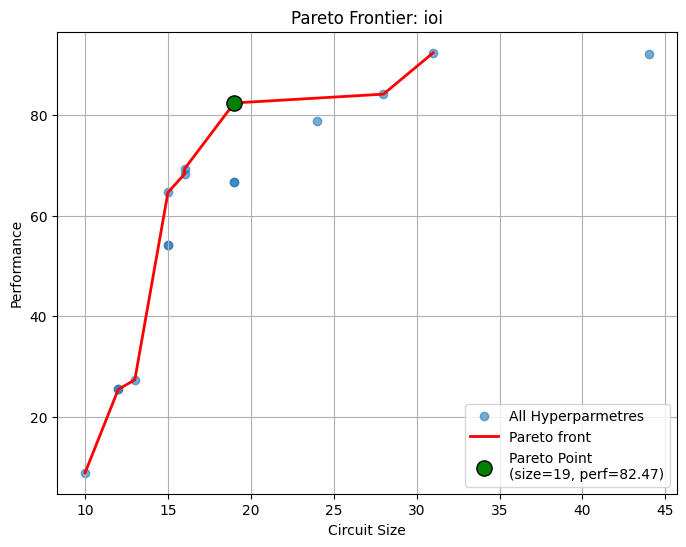}
            \caption{IOI task}
        \end{subfigure} \qquad
        \begin{subfigure}{0.45\textwidth}
            \centering
            \includegraphics[width=\linewidth]{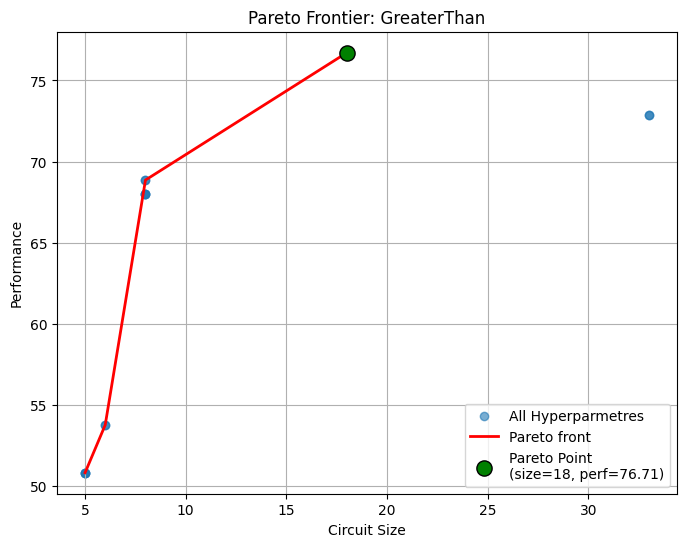}
            \caption{GreaterThan task}
        \end{subfigure} \\
        \par \medskip
        \begin{subfigure}{0.45\textwidth}
            \centering
            \includegraphics[width=\linewidth]{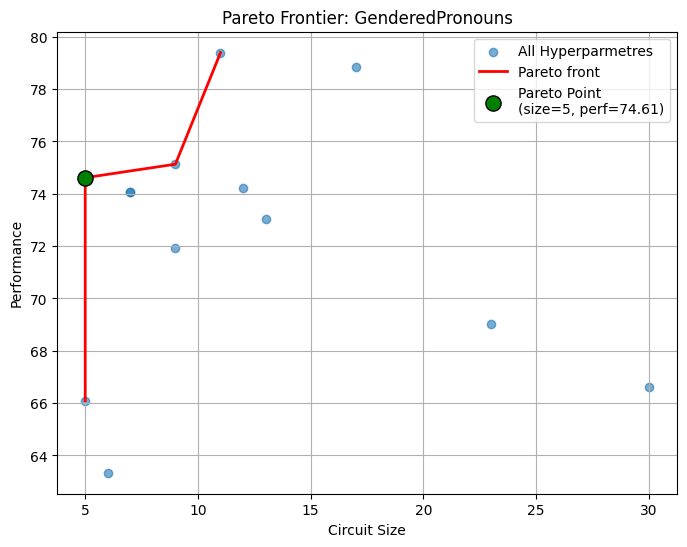}
            \caption{GenderedPronouns task}
        \end{subfigure} \qquad
        \begin{subfigure}{0.45\textwidth}
            \centering
            \includegraphics[width=\linewidth]{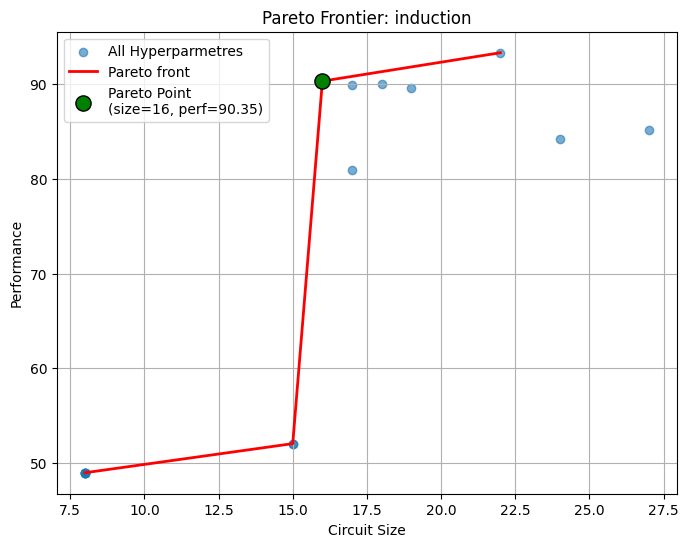}
            \caption{Induction task}
        \end{subfigure} \\
        \par \medskip
        \begin{subfigure}{0.45\textwidth}
            \centering
            \includegraphics[width=\linewidth]{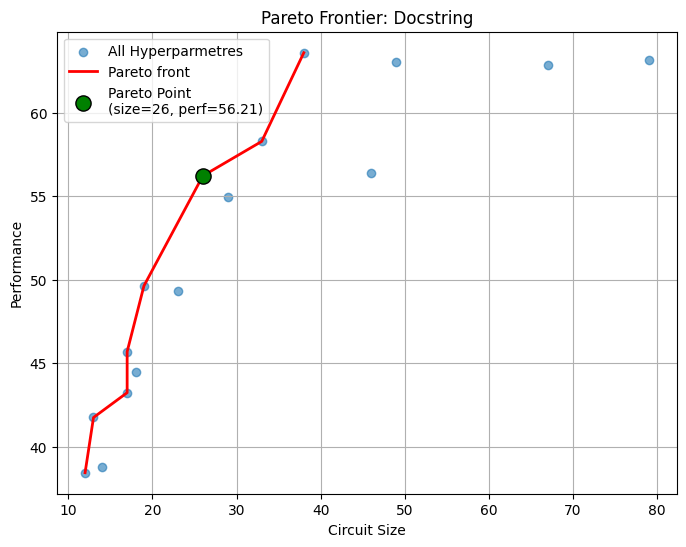}
            \caption{Docstring task}
        \end{subfigure} \\
        \par \medskip

        \small            
        \resizebox{\textwidth}{!}{%
        \begin{tabular}{l|l|cc|cc|cc|cc|cc}
        \toprule
        Maximum Value & Importance
        & \multicolumn{2}{c|}{\textbf{IOI}}
        & \multicolumn{2}{c|}{\textbf{GreaterThan}}
        & \multicolumn{2}{c|}{\textbf{GenderedPronouns}}
        & \multicolumn{2}{c|}{\textbf{Induction}}
        & \multicolumn{2}{c}{\textbf{Docstring}} \\
        & & \textbf{P} & size
        & \textbf{P} & size
        & \textbf{P} & size
        & \textbf{P} & size
        & \textbf{P} & size \\
        \midrule \midrule

        \multirow{4}{*}{0.01}
        & 1 & \textbf{82.47\%} & \textbf{19} & 68.86\% & 8 & 79.39\% & 11 & 52.03\% & 15 & \textbf{56.21\%} & \textbf{26} \\
        & 1.5 & 68.31\% & 16 & 68.86\% & 8 & 71.94\% & 9 & 52.03\% & 15 & 49.64\% & 19 \\
        & 2 & 69.35\% & 16 & 50.81\% & 5 & 63.34\% & 6 & 48.95\% & 8 & 38.79\% & 14 \\
        & 2.5 & 64.68\% & 15 & 50.81\% & 5 & 66.09\% & 5 & 48.95\% & 8 & 38.45\% & 12 \\
        \hline
        \multirow{4}{*}{0.001}
        & 1 & 92.14\% & 44 & 72.86\% & 33 & 66.61\% & 30 & 84.24\% & 24 & 63.03\% & 49 \\
        & 1.5 & 92.47\% & 31 & \textbf{76.71\%} & \textbf{18} & 69.02\% & 23 & 89.89\% & 17 & 58.27\% & 33 \\
        & 2 & 66.82\% & 19 & 68.03\% & 8 & 74.23\% & 12 & \textbf{90.35\%} & \textbf{16} & 49.33\% & 23 \\
        & 2.5 & 54.25\% & 15 & 68.03\% & 8 & 74.05\% & 7 & 89.64\% & 19 & 43.24\% & 17 \\
        \hline
        \multirow{4}{*}{0.02}
        & 1 & 27.37\% & 13 & 53.76\% & 6 & \textbf{74.61\%} & \textbf{5} & 48.95\% & 8 & 54.94\% & 29 \\
        & 1.5 & 25.45\% & 12 & 53.76\% & 6 & \textbf{74.61\%} & \textbf{5} & 48.95\% & 8 & 44.48\% & 18 \\
        & 2 & 25.45\% & 12 & 50.81\% & 5 & \textbf{74.61\%} & \textbf{5} & 48.95\% & 8 & 41.76\% & 13 \\
        & 2.5 & 8.87\% & 10 & 50.81\% & 5 & \textbf{74.61\%} & \textbf{5} & 48.95\% & 8 & 45.70\% & 17 \\
        \hline
        \multirow{4}{*}{0.002}
        & 1 & 84.22\% & 28 & 72.86\% & 33 & 78.83\% & 17 & 85.19\% & 27 & 63.15\% & 79 \\
        & 1.5 & 78.83\% & 24 & \textbf{76.71\%} & \textbf{18} & 73.02\% & 13 & \textcolor{Red}{93.35\%} & \textcolor{Red}{22} & 62.85\% & 67 \\
        & 2 & 66.82\% & 19 & 68.03\% & 8 & 75.13\% & 9 & 90.05\% & 18 & 56.36\% & 46 \\
        & 2.5 & 54.25\% & 15 & 68.03\% & 8 & 74.05\% & 7 & 80.94\% & 17 & 63.58\% & 38 \\
        \bottomrule
        \end{tabular}
        }
        \caption{APP on GPT-2 small: Pareto points for each task are marked in bold. In \textcolor{Red}{red} are the circuits discovered by PP.}
        \label{tab:Pareto_APP_gpt2-small}

    \end{minipage}
}
\end{figure*}

\begin{figure*}[t]
\centering
\scalebox{0.85}{
    \begin{minipage}{1\textwidth}
    \centering
    \textbf{Accelerated Path Patching: Hyperparameter testing for GPT-2 large}

    \begin{subfigure}{0.45\textwidth}
        \centering
        \includegraphics[width=\linewidth]{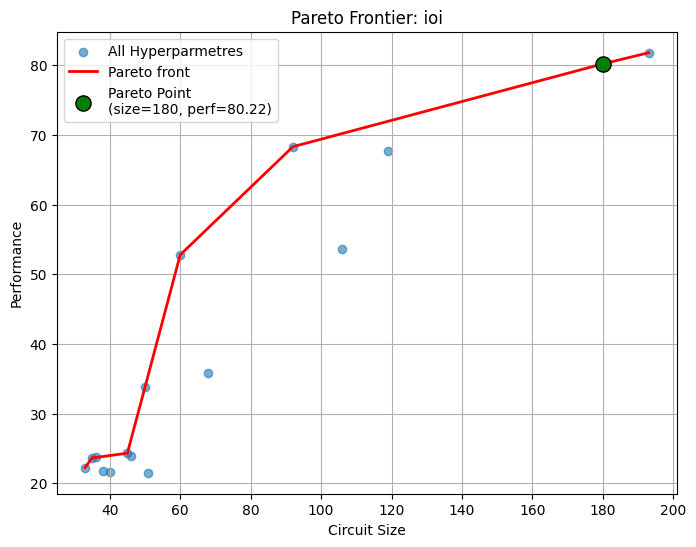}
        \caption{IOI task}
    \end{subfigure} \qquad
    \begin{subfigure}{0.45\textwidth}
        \centering
        \includegraphics[width=\linewidth]{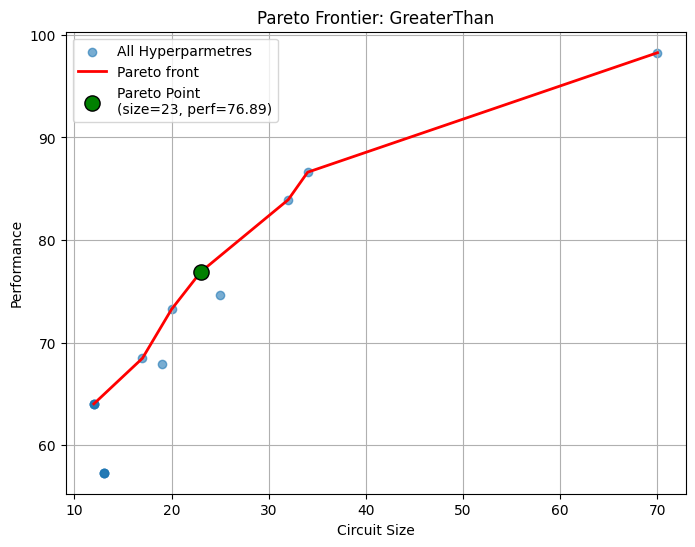}
        \caption{GreaterThan task}
    \end{subfigure} \\[0.5em]

    \begin{subfigure}{0.45\textwidth}
        \centering
        \includegraphics[width=\linewidth]{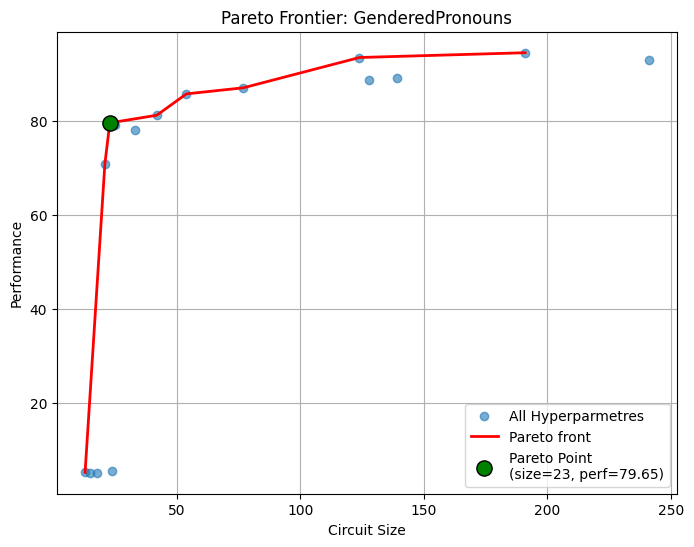}
        \caption{GenderedPronouns task}
    \end{subfigure} \qquad
    \begin{subfigure}{0.45\textwidth}
        \centering
        \includegraphics[width=\linewidth]{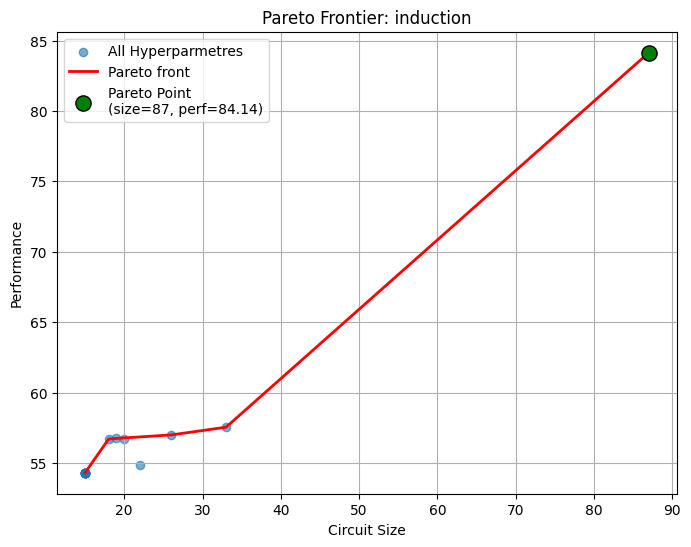}
        \caption{Induction task}
    \end{subfigure} \\[0.5em]

    \begin{subfigure}{0.45\textwidth}
        \centering
        \includegraphics[width=\linewidth]{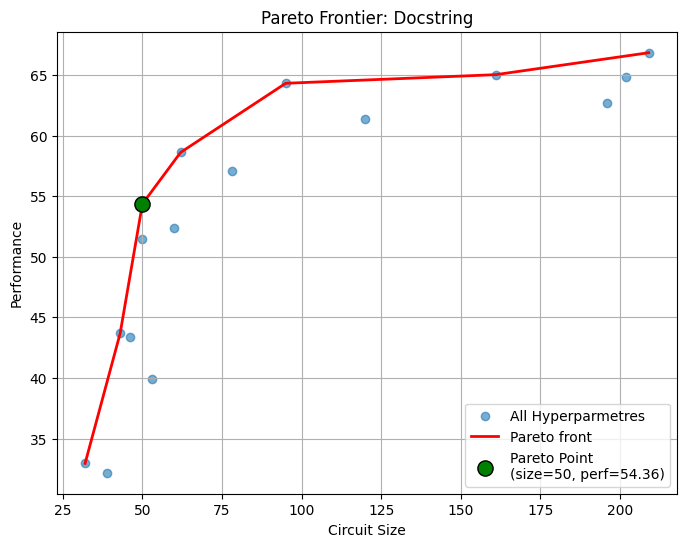}
        \caption{Docstring task}
    \end{subfigure} \\[1em]

    \small
    \begin{tabular}{l|l|cc|cc|cc|cc|cc}
        \toprule
        \multirow{2}{*}{Maximum Value} & \multirow{2}{*}{Importance}
        & \multicolumn{2}{c|}{\textbf{IOI}}
        & \multicolumn{2}{c|}{\textbf{GreaterThan}}
        & \multicolumn{2}{c|}{\textbf{GenderedPronouns}}
        & \multicolumn{2}{c|}{\textbf{Induction}}
        & \multicolumn{2}{c}{\textbf{Docstring}} \\
        & & \textbf{P} & size
        & \textbf{P} & size
        & \textbf{P} & size
        & \textbf{P} & size
        & \textbf{P} & size \\
        \midrule \midrule
        \multirow{4}{*}{0.01}
        & 1 & 21.53 & 51 & 57.33 & 13 & 78.22 & 33 & 54.30 & 15 & 58.61 & 62 \\
        & 1.5 & 24.34 & 45 & 57.33 & 13 & 79.10 & 25 & 54.30 & 15 & \textbf{54.36} & \textbf{50} \\
        & 2 & 21.81 & 38 & 64.04 & 12 & \textbf{79.65} & \textbf{23} & 54.30 & 15 & 43.37 & 46 \\
        & 2.5 & 23.68 & 35 & 64.04 & 12 & 70.90 & 21 & 54.30 & 15 & 43.67 & 43 \\
        \hline
        \multirow{4}{*}{0.001}
        & 1 & 81.82 & 193 & 98.25 & 70 & 92.94 & 241 & \textbf{84.14} & \textbf{87} & 66.83 & 209 \\
        & 1.5 & \textbf{80.22} & \textbf{180} & 86.61 & 34 & 94.57 & 191 & 57.53 & 33 & 64.80 & 202 \\
        & 2 & 67.70 & 119 & \textbf{76.89} & \textbf{23} & 93.55 & 124 & 56.98 & 26 & 62.72 & 196 \\
        & 2.5 & 52.80 & 60 & 67.94 & 19 & 88.67 & 128 & 54.85 & 2 & 51.48 & 50 \\
        \hline
        \multirow{4}{*}{0.02}
        & 1 & 23.94 & 46 & 57.33 & 13 & 5.52 & 24 & 54.30 & 15 & 52.41 & 60 \\
        & 1.5 & 21.66 & 40 & 57.33 & 13 & 5.28 & 18 & 54.30 & 15 & 39.94 & 53 \\
        & 2 & 22.26 & 33 & 64.04 & 12 & 5.25 & 15 & 54.30 & 15 & 32.18 & 39 \\
        & 2.5 & 23.72 & 36 & 64.04 & 12 & 5.32 & 13 & 54.30 & 15 & 32.95 & 32 \\
        \hline
        \multirow{4}{*}{0.002}
        & 1 & 68.33 & 92 & 83.93 & 32 & 89.28 & 139 & 56.69 & 20 & 65.02 & 161 \\
        & 1.5 & 53.68 & 106 & 74.68 & 25 & 87.11 & 77 & 56.73 & 19 & 64.31 & 95 \\
        & 2 & 35.83 & 68 & 73.26 & 20 & 85.82 & 54 & 56.68 & 18 & 57.07 & 78 \\
        & 2.5 & 33.80 & 50 & 68.48 & 17 & 81.28 & 42 & 54.30 & 15 & 61.35 & 120 \\
        \bottomrule
    \end{tabular}

    \caption{APP on GPT-2 large: Pareto points for each task are marked in bold.}
    \label{tab:Pareto_APP_GPT2-large}
    \end{minipage}
}
\end{figure*}

\begin{figure*}[t]
\scalebox{0.85}{
    \centering
    \begin{minipage}{1\textwidth}
        \centering
        \textbf{Accelerated Path Patching: Hyperparameter testing for Qwen2.5-0.5B}

        \begin{subfigure}{0.45\textwidth}
            \centering
            \includegraphics[width=\linewidth]{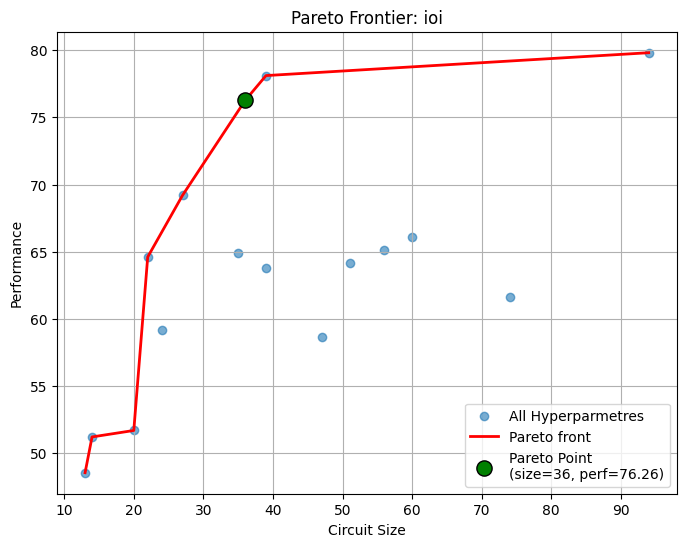}
            \caption{IOI task}
        \end{subfigure} \qquad
        \begin{subfigure}{0.45\textwidth}
            \centering
            \includegraphics[width=\linewidth]{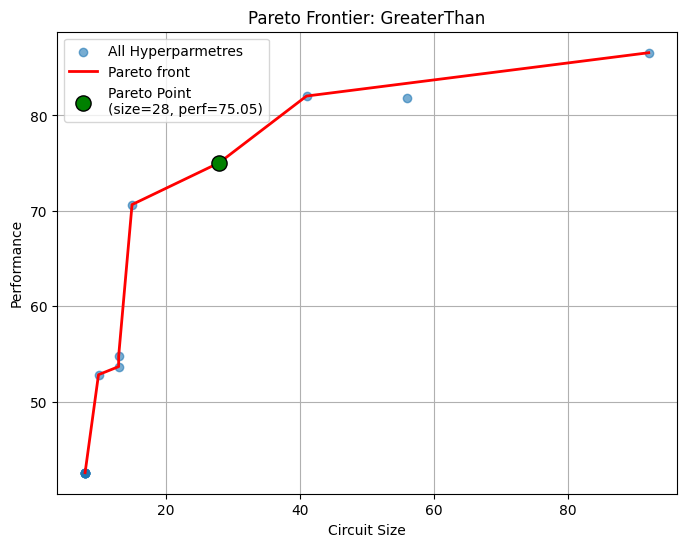}
            \caption{GreaterThan task}
        \end{subfigure} \\
        \par \medskip

        \begin{subfigure}{0.45\textwidth}
            \centering
            \includegraphics[width=\linewidth]{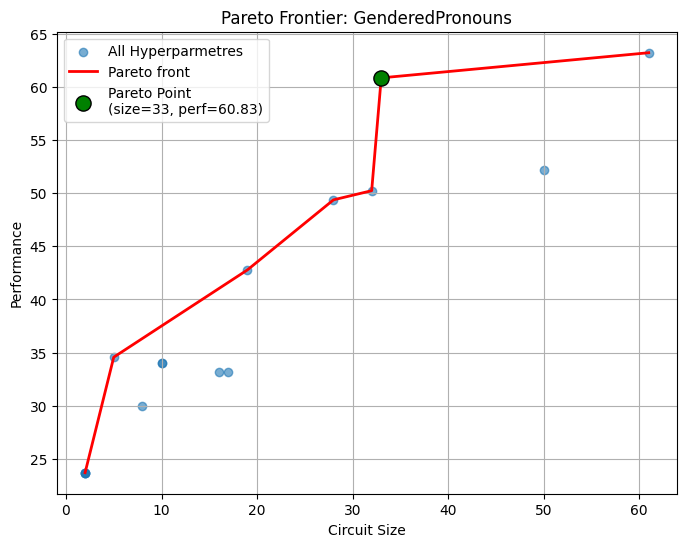}
            \caption{GenderedPronouns task}
        \end{subfigure} \qquad
        \begin{subfigure}{0.45\textwidth}
            \centering
            \includegraphics[width=\linewidth]{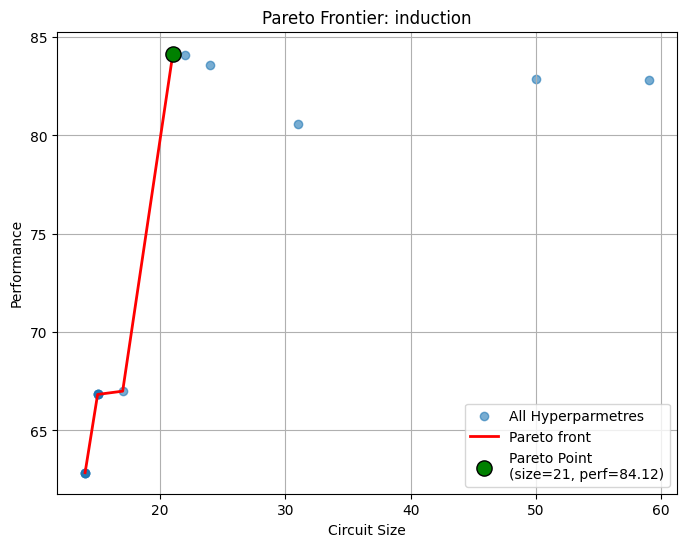}
            \caption{Induction task}
        \end{subfigure} \\
        \par \medskip

        \begin{subfigure}{0.45\textwidth}
            \centering
            \includegraphics[width=\linewidth]{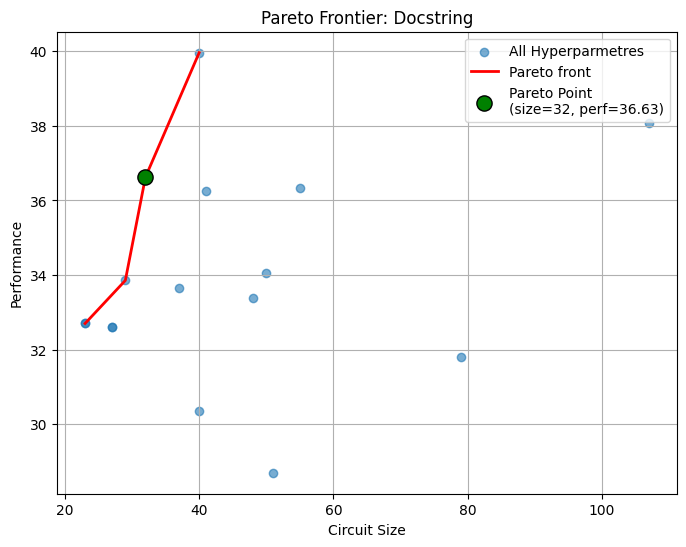}
            \caption{Docstring task}
        \end{subfigure}
        \\
        \par \medskip
        \par \medskip

        \small
        \resizebox{\textwidth}{!}{%
        \begin{tabular}{l|l|cc|cc|cc|cc|cc}
        \toprule
        \multirow{2}{*}{Maximum Value} & \multirow{2}{*}{Importance}
        & \multicolumn{2}{c|}{\textbf{IOI}}
        & \multicolumn{2}{c|}{\textbf{GreaterThan}}
        & \multicolumn{2}{c|}{\textbf{GenderedPronouns}}
        & \multicolumn{2}{c|}{\textbf{Induction}}
        & \multicolumn{2}{c}{\textbf{Docstring}} \\

        & & \textbf{P} & size
        & \textbf{P} & size
        & \textbf{P} & size
        & \textbf{P} & size
        & \textbf{P} & size \\
        \midrule \midrule

        \multirow{4}{*}{0.01}
        & 1   & 78.11 & 39 & 42.55 & 8 & 49.36 & 28 & 66.83 & 15 & 34.06 & 50 \\
        & 1.5 & 69.19 & 27 & 42.55 & 8 & 42.78 & 19 & 66.83 & 15 & 36.24 & 41 \\
        & 2   & 64.61 & 22 & 42.55 & 8 & 29.98 & 8 & 66.83 & 15 & \textbf{36.63} & \textbf{32} \\
        & 2.5 & 51.22 & 14 & 42.55 & 8 & 34.57 & 5 & 66.83 & 15 & 33.86 & 29 \\
        \hline

        \multirow{4}{*}{0.001}
        & 1   & 61.61 & 74 & 86.57 & 92 & 63.19 & 61 & 82.84 & 59 & 38.09 & 107 \\
        & 1.5 & 58.65 & 47 & 81.79 & 56 & \textbf{60.83} & \textbf{33} & 82.88 & 50 & 36.34 & 55 \\
        & 2   & 62.13 & 43 & \textbf{75.05} & \textbf{28} & 33.16 & 17 & 84.08 & 22 & 33.39 & 48 \\
        & 2.5 & 63.79 & 39 & 53.67 & 13 & 34.06 & 10 & \textbf{84.12} & \textbf{21} & 33.65 & 37 \\
        \hline

        \multirow{4}{*}{0.02}
        & 1   & \textbf{76.26} & \textbf{36} & 42.55 & 8 & 23.71 & 2 & 62.85 & 14 & 32.60 & 27 \\
        & 1.5 & 59.17 & 24 & 42.55 & 8 & 23.71 & 2 & 62.85 & 14 & 32.60 & 27 \\
        & 2   & 51.70 & 20 & 42.55 & 8 & 23.71 & 2 & 62.85 & 14 & 32.70 & 23 \\
        & 2.5 & 48.56 & 13 & 42.55 & 8 & 23.71 & 2 & 62.85 & 14 & 32.70 & 23 \\
        \hline

        \multirow{4}{*}{0.002}
        & 1   & 79.81 & 94 & 82.03 & 41 & 52.17 & 50 & 80.60 & 31 & 31.80 & 79 \\
        & 1.5 & 66.09 & 60 & 70.66 & 15 & 50.22 & 32 & 83.58 & 24 & 28.70 & 51 \\
        & 2   & 64.13 & 51 & 54.76 & 13 & 33.19 & 16 & \textbf{84.12} & \textbf{21} & 30.36 & 40 \\
        & 2.5 & 64.94 & 35 & 52.83 & 10 & 34.06 & 10 & 67.00 & 17 & 39.96 & 40 \\
        \bottomrule
        \end{tabular}
        }

        \caption{APP on Qwen2.5-0.5B: Pareto points for each task are marked in bold.}
        \label{tab:Pareto_APP_qwen2.5-0.5B}
    \end{minipage}
    }
\end{figure*}

\begin{figure*}[t]
\centering
\textbf{Accelerated Path Patching: Hyperparameter testing for Qwen2.5-7B}

\begin{subfigure}{0.26\textwidth}
    \centering
    \includegraphics[width=\linewidth]{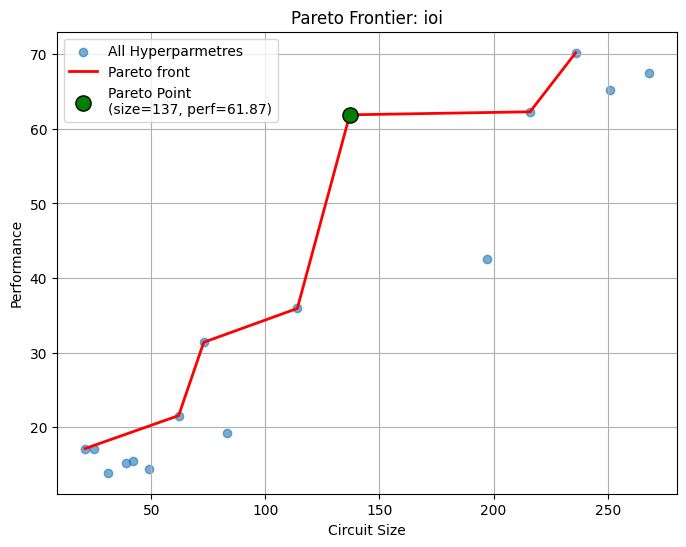}
    \caption{IOI task}
\end{subfigure} 
\begin{subfigure}{0.26\textwidth}
    \centering
    \includegraphics[width=\linewidth]{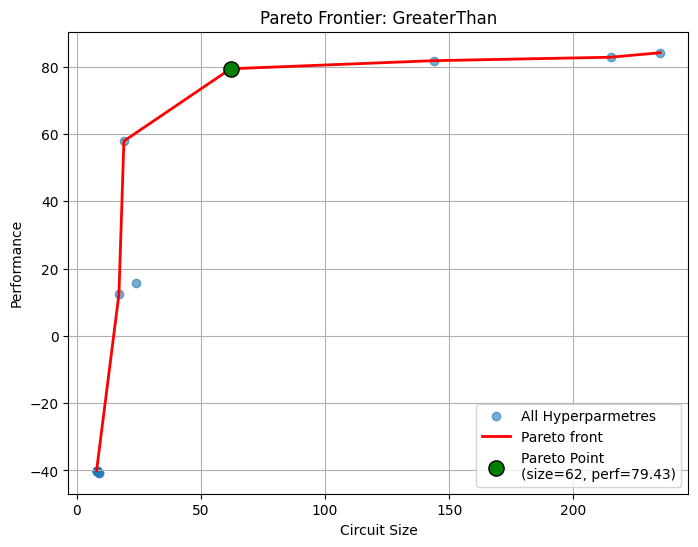}
    \caption{GreaterThan task}
\end{subfigure}
\begin{subfigure}{0.26\textwidth}
    \centering
    \includegraphics[width=\linewidth]{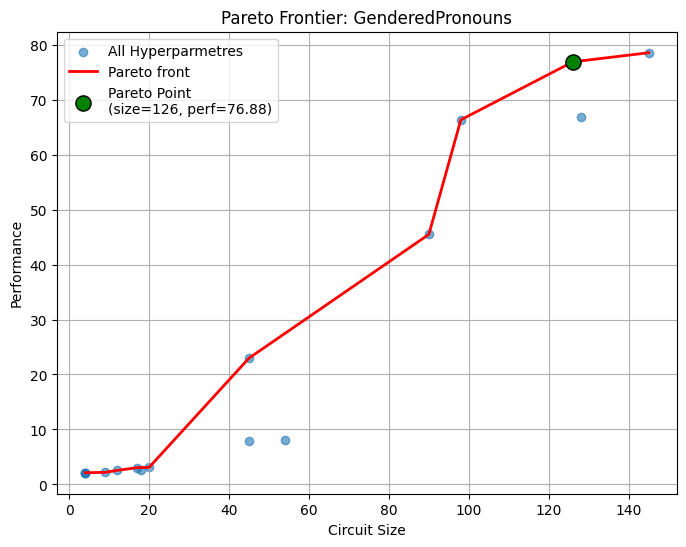}
    \caption{GenderedPronouns task}
\end{subfigure} 
\begin{subfigure}{0.26\textwidth}
    \centering
    \includegraphics[width=\linewidth]{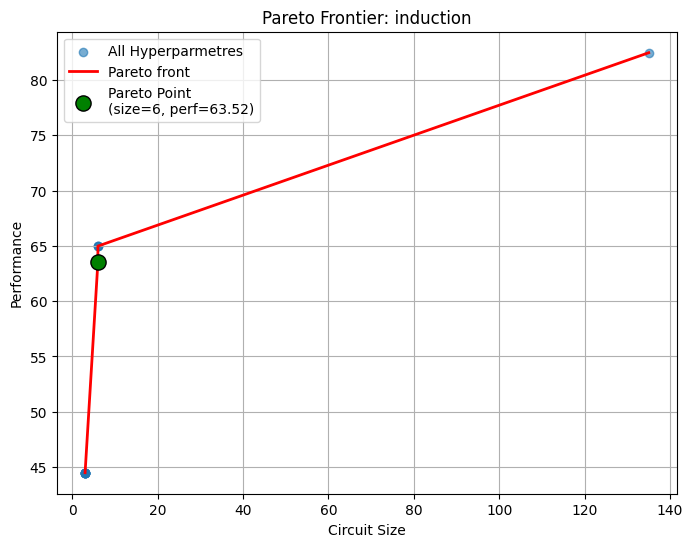}
    \caption{Induction task}
\end{subfigure}
\begin{subfigure}{0.26\textwidth}
    \centering
    \includegraphics[width=\linewidth]{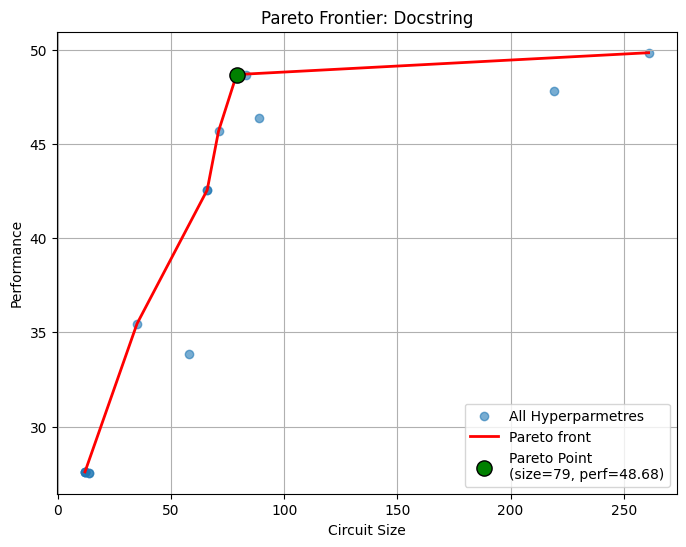}
    \caption{Docstring task}
\end{subfigure}

\small
\resizebox{\textwidth}{!}{%
\begin{tabular}{l|l|cc|cc|cc|cc|cc}
\toprule
\multirow{2}{*}{Maximum Value} & \multirow{2}{*}{Importance}
& \multicolumn{2}{c|}{\textbf{IOI}}
& \multicolumn{2}{c|}{\textbf{GreaterThan}}
& \multicolumn{2}{c|}{\textbf{GenderedPronouns}}
& \multicolumn{2}{c|}{\textbf{Induction}}
& \multicolumn{2}{c}{\textbf{Docstring}} \\
& & \textbf{P} & size & \textbf{P} & size & \textbf{P} & size & \textbf{P} & size & \textbf{P} & size \\
\midrule \midrule

\multirow{4}{*}{0.01}
& 1 & 19.27 & 83 & -40.06 & 8 & 4.20 & 30 & \textbf{64.98} & \textbf{6} & 27.55 & 14 \\
& 1.5 & 21.58 & 62 & -40.06 & 8 & 3.01 & 21 & \textbf{64.98} & \textbf{6} & 27.59 & 12 \\
& 2 & 14.49 & 49 & -40.06 & 8 & 3.08 & 20 & \textbf{64.98} & \textbf{6} & 27.59 & 12 \\
& 2.5 & 15.44 & 42 & -40.06 & 8 & 3.04 & 17 & 44.49 & 3 & 27.59 & 12 \\
\hline

\multirow{4}{*}{0.001}
& 1 & 67.48 & 268 & 84.20 & 235 & 78.54 & 145 & 90.07 & 190 & 49.83 & 261 \\
& 1.5 & 65.16 & 251 & 82.86 & 215 & 66.83 & 128 & 64.30 & 9 & 48.65 & 83 \\
& 2 & 62.28 & 216 & 81.84 & 144 & 45.49 & 90 & \textbf{64.98} & \textbf{6} & 42.55 & 66 \\
& 2.5 & 35.93 & 114 & 57.92 & 19 & 8.00 & 54 & \textbf{64.98} & \textbf{6} & 33.83 & 58 \\
\hline

\multirow{4}{*}{0.02}
& 1 & 15.18 & 39 & -40.06 & 8 & 2.19 & 9 & \textbf{64.98} & \textbf{6} & 27.55 & 14 \\
& 1.5 & 17.16 & 25 & -40.06 & 8 & 2.19 & 9 & \textbf{64.98} & \textbf{6} & 27.59 & 12 \\
& 2 & 17.16 & 21 & -40.06 & 8 & 2.19 & 9 & \textbf{64.98} & \textbf{6} & 27.59 & 12 \\
& 2.5 & 13.92 & 17 & -40.09 & 8 & 2.19 & 9 & 44.49 & 3 & 27.59 & 12 \\
\hline

\multirow{4}{*}{0.002}
& 1 & 42.61 & 197 & \textbf{79.43} & \textbf{62} & \textbf{76.88} & \textbf{126} & \textbf{64.98} & \textbf{6} & 47.79 & 219 \\
\rowcolor{gray!15}
& 1.5 & 70.19 & 236 & 15.84 & 24 & 66.34 & 98 & \textbf{64.98} & \textbf{6} & \textbf{48.68} & \textbf{79} \\
& 2 & \textbf{61.87} & \textbf{137} & 12.42 & 17 & 22.97 & 45 & \textbf{64.98} & \textbf{6} & 42.55 & 66 \\
& 2.5 & 31.40 & 73 & -40.69 & 9 & 7.94 & 45 & 44.49 & 3 & 33.83 & 58 \\
\bottomrule
\end{tabular}
} \caption{APP on Qwen2.5-7B: Pareto points for each task are marked in \textbf{bold}. The highlighted row corresponds to the Path Patching configuration.}
\label{tab:Pareto_APP_qwen2.5-7B}
\end{figure*}

In Figure~\ref{tab:Pareto_APP_gpt2-small}, \ref{tab:Pareto_APP_GPT2-large}, \ref{tab:Pareto_APP_qwen2.5-0.5B}, and \ref{tab:Pareto_APP_qwen2.5-7B}, we show the resulting circuits from APP and model performances across tasks and model families under different hyperparameter settings.

\newpage 

\end{document}